%% file: main.tex
\documentclass[pdflatex, sn-apa, iicol]{sn-jnl} 

\usepackage{graphicx}%
\usepackage{multirow}%
\usepackage{amsmath,amssymb,amsfonts}%
\usepackage{amsthm}%
\usepackage{mathrsfs}%
\usepackage[title]{appendix}%
\usepackage{xcolor}%
\usepackage{textcomp}%
\usepackage{manyfoot}%
\usepackage{booktabs}%
\usepackage{algorithm}%
\usepackage{algorithmicx}%
\usepackage{algpseudocode}%
\usepackage{listings}%

\usepackage{microtype}
\usepackage{subcaption}
\graphicspath{ {images/} }
\usepackage{booktabs,tabularx}
\usepackage{float}
\usepackage{caption}
\usepackage{makecell}
\usepackage{array}
\usepackage{comment}
\usepackage{verbatim}

\usepackage{paralist} 
\usepackage{enumitem} 
\usepackage{siunitx} 
\usepackage{tikz,bm} 
\usepackage{circuitikz}
\usepackage{tikz}
\usetikzlibrary{decorations.pathmorphing,shapes,calc,shapes.geometric, decorations.pathmorphing, shapes.arrows, fadings, shadows, arrows.meta, positioning, decorations.pathreplacing, calc, fit, backgrounds} 
\usepackage{mathtools}
\usepackage[T1]{fontenc}
\usepackage[outline]{contour}
\usepackage{upgreek}
\usepackage{empheq}
\usepackage{caption}
\usepackage{siunitx}

\usepackage{xr}
\usepackage{cleveref}

\usepackage{hyperref}
\usepackage{wrapfig}
\usepackage{subcaption}
\usepackage{subfiles}

\usepackage [english]{babel}

\newcommand{\floor}[1]{\left \lfloor #1 \right \rfloor}
\newcommand{\ceil}[1]{\left \lceil #1 \right \rceil}

\theoremstyle{thmstyleone}%
\theoremstyle{thmstyletwo}%

\theoremstyle{thmstylethree}%

\begin{document}

\title[Article Title]{Noise2Params: Unification and Parameter Determination from Noise via a Probabilistic Event Camera Model}


\author[1]{\fnm{Owen} \sur{Root}}\email{owen.root14@login.cuny.edu}

\author[1]{\fnm{Julinda} \sur{Mujo}}

\author*[1]{\fnm{Min} \sur{Xu}}\email{min.xu04@login.cuny.edu}

\affil[1]{\orgdiv{Department of Physics and Astronomy}, \orgname{CUNY Hunter College}, \orgaddress{\street{695 Park Ave}, \city{New York}, \postcode{10065}, \state{NY}, \country{USA}}}




\abstract{Accurate, unified models for event cameras (ECs) remain elusive, hampering calibration and algorithm design. We develop a foundational probabilistic model for EC event detection, grounded in photon statistics, that unifies the description of static scene noise events and step response curves (S-curves) within a single analytical framework. Three formulations of the probability distributions are derived, spanning all intensity regimes: exact Poisson, saddle-point, and Gaussian. The model reveals the underlying connection between these otherwise disparate EC behaviors and clarifies the interpretation of S-curves, which we show is more nuanced than selecting a fixed probability threshold. Based on this model, we propose Noise2Params, a method for determining camera-specific values of the log-contrast threshold $B$, the lux-to-photon conversion factor $\alpha$, and the leakage term $\theta$ (found to be intensity dependent), via error minimization against observed noise-event distributions. Noise2Params requires only recordings of static, uniform scenes, offering an experimentally accessible alternative to approaches that demand specialized dynamic light sources. We further support the validity the model by training convolutional neural networks (CNNs) on synthetic noise images generated from our distributions and evaluating their ability to reconstruct static scenes from experimental data. We further demonstrate the utility of our model by showing that CNNs incorporating synthetic data outperform those trained solely on experimental data. Our framework provides a quantitative foundation for EC calibration, noise-aware algorithm design, and applications in photon-limited regimes.}




\maketitle

\section{Introduction}
Event cameras (ECs) are an emerging digital vision technology, with advantages over traditional camera technology such as asynchronous recording, high-dynamic range, low energy usage, and fast temporal response, making them ideal for computer vision \citep{gallego2020event}. These advantages are based upon the operation of ECs wherein each pixel responds independently to changes in light intensity, as opposed to frame-based traditional cameras (TCs), where all pixels record light intensity at the same discrete time steps. 
\par
Events that result from increases in intensity are referred to as positive events and ones resulting from decreases are referred to as negative events. A recording from an EC is a stream of these events. Given that frame-based imaging approaches lend themselves to greater accessibility for humans, a significant portion of the research on ECs has involved developments of ``frame reconstruction'' methods, wherein frames similar to those created by TCs are reconstructed from event-based recordings \citep{BrandliChristian2014Rhvd, wang2020eventcameracalibrationperpixel, cao2025noise2image}. Regardless of application, all such efforts depend on accurate characterization of EC behavior.
\par
While ECs are commercially available and are being studied in a wide variety of areas \citep{gallego2020event}, a quantitative model for EC event generation is still lacking. This has led to challenges and differing (often conflicting) approaches in how to determine physical parameters of ECs, which is of foundational importance for precise scientific applications. A common approach to parameter estimation is based on the step response curve (S-curve), which is a pixel response probability curve at varying log-contrast values, generated by exposing EC pixels to periodic stimuli with known amplitudes and baselines. However, interpretations of S-curves vary, with some authors employing a 50\% probability threshold \citep{posch2011sensitivity, finateu2020event, niwa2023sharedpixel, guo2023hybrid}, and others arguing for the use of a 100\% threshold \citep{mcreynolds2024stepresponse, mcreynolds2022experimental}. Another challenge that ECs face is the presence of noise events (events recorded when no dynamics occurred), a thorough understanding of which is crucial for high-resolution applications. This EC noise is less straightforward to remedy than the analogous noise in TCs, due to the differences in their operational principles.
\par
In this work we present a foundational, quantitative model for the probability of event detections, based on the discrete nature of photon arrivals. \footnote{A preliminary conference version of this work appeared in the proceedings of SPIE Photonics West \citep{Root2026SPIE}. The present manuscript is a substantially expanded journal article, with additional theoretical derivations, experimental characterization, parameter-estimation analysis, synthetic-data generation, support via deep-learning reconstruction, and supplementary material beyond the conference version.} Our model is valid when observing static scenes, where all events are noise events, as well as when there are actual dynamics, such as in S-curve generation, which offers clarity on the interpretation of S-curves. Furthermore, we propose Noise2Params, a method for determination of camera-specific parameter values, enabled by our analytic model, via error-minimization of the model against observed static scene noise-event distributions. This allows for parameter determination using highly accessible measurements (the noise-event distributions), which can be supplemented with additional observations, such as S-curves, if available. 
\par
The rest of this work is structured as follows. In Section~\ref{section: EC principles}, we discuss the working principles of ECs and define the key parameters. In Section~\ref{section: review}, we review related work on EC parameter estimation, including the S-curve method and noise event models. In Section~\ref{section: theory}, we present three formulations of an analytical probability model for event detection, while in Section~\ref{section: experiments} we present experimental observations of noise events and compile reported S-curves. In Section~\ref{section: param determ}, we use these data to determine the values of EC parameters via alignment with our model, a process we call Noise2Params. In Section~\ref{sec: noise_event_images}, we apply the fitted model to generate synthetic noise images and validate it through deep-learning-based image reconstruction. Finally, in Section~\ref{section: discussion} we discuss the physical interpretation of the results, clarify the interpretation of S-curves, provide practical guidance on the use of Noise2Params, and outline applications of our framework, before concluding in Section~\ref{section: conclusion}.
\par
A supplementary information document (Online Resource 1) accompanies this paper. Sections of the supplement are referred to as Sec.~\ref{s-section: derivation of distributions} through Sec.~\ref{s-sec:recon_from_noise_images} in the main text.


\section{Event Camera Principles} \label{section: EC principles}
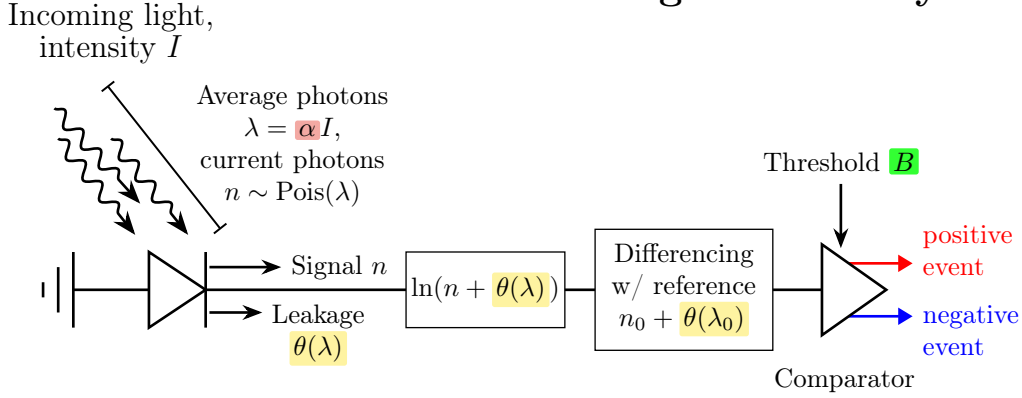
\begin{figure*} 
    \centering
    \definecolor{redHL}{HTML}{f0948d}
    \definecolor{yelHL}{HTML}{fef08a} 
    \definecolor{grnHL}{HTML}{bbf7d0} 
    
    \newcommand{\mathhl}[2]{%
      \tikz[baseline=(X.base)]{
        \node[
          fill=#1,
          fill opacity=0.8,
          text opacity=1,
          rounded corners=1pt,
          inner xsep=1.5pt,
          inner ysep=1.5pt,
          anchor=base
        ] (X) {$#2$};
      }%
    }

    
    \begin{tikzpicture}
    \draw[line width=1pt] (0,0) -- (10,0);
    \draw[line width=1pt] (0,-.5)--(0,0.5);
    \draw[line width=1pt] (-0.2,-.3)--(-0.2,0.3);
    \draw[line width=1pt] (-0.4,-.1)--(-0.4,0.1);

    \draw[line width=1pt, fill=white] (1,-0.5)--(1,0.5)--(1.75,0)--cycle;
    \draw[line width=1pt](1.75,-0.5)--(1.75,0.5);

    \draw[line width=1pt, -{Stealth[scale=1.2]}, shorten >=-10pt, decorate, decoration = {snake}] (-0.2,2)--(0.6,1);
    \draw[line width=1pt, -{Stealth[scale=1.2]}, shorten >=-10pt, decorate, decoration = {snake}] (0.5,2)--(1.3,1);
    \draw[line width=1pt, -{Stealth[scale=1.2]}, shorten >=-10pt, decorate, decoration = {snake}] (-0.25,2.4)--(0.65,1.4);

    \draw[line width=0.75pt, |-|] (0.45,2.7)--(1.95,.8);
    \node[align=center, rotate=0] at (0.5,3.4) {\large{Incoming light}, \\ \large{intensity $I$}};
    \node[align=center, rotate=0] at (2.9,1.9) {Average photons \\
    $\lambda=\mathhl{redHL}{\alpha} I$, \\
    current photons \\
    $n \sim \text{Pois}(\lambda)$};

    \draw[line width=1pt, -{Stealth}] (1.8,0.3)--(2.75,0.3) node[pos=1, right] {Signal $n$};
    \draw[line width=1pt, -{Stealth}] (1.8,-.3)--(2.5,-0.3) node[pos=1, right, align=center, yshift=-0.3 cm] {Leakage \\ $\mathhl{yelHL}{\theta(\lambda)}$};
    
    \draw[line width=0.5pt, fill=white] (4.4,-0.5)--(4.4,0.5)--(6.5,0.5)--(6.5,-0.5)--cycle;
    \node[align=center] at (5.45,0) {$\ln(n+\mathhl{yelHL}{\theta(\lambda)})$};


    \draw[line width=0.5pt, fill=white] (6.9,-0.8)--(6.9,0.8)--(9.25,0.8)--(9.25,-0.8)--cycle;
    \node[align=center] at (8.075,0) {Differencing \\ w/ reference \\ $n_0+\mathhl{yelHL}{\theta(\lambda_0)}$};

    \draw[line width=1pt, color=red, -{Triangle}] (10.1,0.35)--(11.1,0.35) node[pos=1,right, align=left, yshift=+0.18cm] {positive \\ event};
    \draw[line width=1pt, color=blue, -{Triangle}] (10.1,-0.35)--(11.1,-0.35) node[pos=1,right, align=left, yshift=-0.18cm] {negative \\  event};

    \draw[line width=1pt, fill=white] (9.9,-0.6)--(9.9,0.6)--(10.75,0)--cycle;
    \draw[line width=1pt, -{Stealth}] (10.15, 1.4) -- (10.15, 0.55) node[pos=0, above] {Threshold $\mathhl{green}{B}$};
    \node[align=center] at (10.2, -1.2) {Comparator};

    \node[align=center] at (7, 4) {\Large{\textbf{Event Detection Signal Pathway}}};
    
    \end{tikzpicture}

    \caption{Schematic of the event camera detection signal pathway, illustrating the origin of the key parameters: the photon conversion factor $\alpha$, the leakage current $\theta$, and the log-contrast threshold $B$. Incoming light of intensity $I$ produces photon counts $n$ drawn from a Poisson process, which constitute the underlying signal. The circuit-style representation is used as a functional analogy to the sensor's internal processing stages and should not be interpreted as literal circuitry.}
    \label{fig:origin_of_parameters}
\end{figure*}

Here we discuss the working principle of ECs and the key parameters involved, which are shown schematically in Fig.~\ref{fig:origin_of_parameters}. As mentioned previously, the pixels of an EC sensor operate independently and asynchronously. When the intensity incident on a pixel increases or decreases, the pixel records a positive or negative event, respectively, if the log contrast ratio between the current intensity and a reference intensity exceeds a given threshold, which is termed the log contrast threshold or the sensitivity parameter. Each event is recorded as a four-vector, $(x, y, t, p)$, where $x, y$ are the pixel coordinates, $t$ is the timestamp, and $p$ is the polarity of the event. Typical conventions for polarity are $p=\pm1$ and also $p=1$ or $0$ for positive or negative events, respectively.
\par
The threshold value may be different or the same for positive and negative events. Furthermore, while an ideal event camera would have pixels with precisely the same threshold, real event cameras typically exhibit per-pixel threshold variations, among other potential differences in pixel behavior \citep{posch2011sensitivity,IMX636sample}.  
\par
For an event to be detected, the log contrast between the current intensity $I$ and a reference intensity $I_0$ must exceed the log contrast threshold, that is, 
\begin{equation} \label{eq:1}
    \ln \left(\frac{I}{I_0} \right) > B_+, 
\end{equation}
for a positive event, or 
\begin{equation} \label{eq:2}
    \ln \left(\frac{I}{I_0}\right) < B_-,
\end{equation}
for a negative event, where $B_\pm$ is the log contrast threshold parameter.
\par
When $B_-=-B_+$, we simply denote it $B$, with the implicit understanding that it is a positive value for positive events, and a negative value for negative events, distinguishing between them with $\pm B$ where necessary. Modern commercial ECs typically have $B$ between 0 and 1 \citep{IMX636sample, mcreynolds2024stepresponse, 2024amos.conf...50O, Prophesee_EVK4}.
\par
Given the discrete nature of light, and assuming sufficient spectral uniformity, the intensities are determined by photon counts, $n$ and $n_0$ for $I$ and $I_0$, respectively. We use $\alpha$ to denote the conversion factor between intensity in lux and photon counts. After an event is detected, a pixel is ``blind'' for a period of time $R$, during which no new events can be detected. This is referred to as the refractory time, or the dead time, of the pixels. After $R$ passes, the pixel is reset and a new $I_0$ is chosen based on the intensity then observed by the pixel. The detection process then starts over again. 
\par
As there has been conflicting information in the literature, we note some clarifications regarding the working principle of ECs, at least specifically for the model we have employed, the EVK4 HD by Prophesee. Some works have implied that a larger intensity change, and hence a larger log contrast ratio, results in more event detections \citep{cabriel2023event}. However, regardless of the size of the log contrast ratio, as long as the threshold is met, only a single event is recorded by a pixel; an intensity change that results in a log contrast ratio further exceeding the threshold does not result in additional events \citep{Prophesee_LargeIntensityResponse}.
\footnote{Confusion about this may have arisen from the fact that in practice a larger intensity change may have a longer transition time between levels (such as a light bulb taking longer to reach a high brightness level when turned on compared to a lower brightness level). This would appear to trigger more events, as the longer transition time allows more temporal opportunities for event triggers. This phenomenon is unrelated to the threshold $B$ itself.}
\par
After a detection, the refractory time must then pass before further events can be detected by that pixel (this fact is useful in the detection of incorrectly operating pixels, see Sec.~\ref{s-section: hot pixels}). Some works have stated that, following an event detection, the reference level $I_0$ is updated to the value of the intensity that triggered the event \citep{wang2020eventcameracalibrationperpixel, cabriel2023event}. This is not accurate for our EC. The new reference intensity is chosen to be the intensity that the pixel first observes after the refractory period passes \citep{mcreynolds2024stepresponse} (Prophesee Support, personal communication, July 29, 2025).
\par
With the basic detection mechanism established, we now refine the event condition to account for non-ideal circuit behavior. In terms of the circuitry of an EC sensor, $I$ and $I_0$ correspond to the currents emitted by the pixel photoreceptor. However, there may also be leakage current present in the circuitry. This can be attributed to various causes, such as dark current leakage from the photoreceptor in low-light conditions, but we can, in general, account for it by including an additional term in the event condition. The positive event condition is then 
\begin{equation} \label{eq:3}
    \ln \left(\frac{I+\theta}{I_0+\theta}\right)>B,
\end{equation}
and the negative event condition is
\begin{equation} \label{eq:4}
    \ln \left(\frac{I+\theta}{I_0+\theta}\right)<-B,
\end{equation}
where $\theta$ accounts for this current leakage. This has led to $\theta$ being referred to as the leakage current or dark current, as well as the ``photoreceptor bias'' \citep{2024amos.conf...50O, mcreynolds2024stepresponse, cao2025noise2image}.

\par
There are additional elements of event camera circuitry that impact event detection behavior, such as high- and low-pass filtration or circuit thermal noise, which we have not explicitly accounted for here. We assume that any impact by such elements can, at least at a first-order level, be encoded into the $\theta$ term. However, we still refer to $\theta$ as the leakage term. To distinguish between the event condition with or without $\theta$, eqs.~(\ref{eq:1}) and~(\ref{eq:2}) shall be referred to as the apparent event conditions, with the apparent sensitivity parameter $B_{\text{app}}$, while~(\ref{eq:3}) and~(\ref{eq:4}) shall be referred to as the fundamental event conditions, with the fundamental sensitivity parameter $B$.
\par
Lastly, we note that commercial ECs typically have various settings that the user can modify to alter the event detection behavior. These can include making the camera more or less sensitive (effectively changing $B$), changing the refractory time, or altering circuit-level effects such as high- or low-pass signal filtering, among others. In keeping with the manufacturer's terminology for our camera, we refer to these settings as bias settings. 
\par
While $R$ can be determined straightforwardly, for example, by observing some smooth dynamics and finding the shortest time between recorded events, determining $B$, $\alpha$, and $\theta$ presents a greater challenge. These are the physical parameters of the camera that we wish to investigate. 

\par

\section{Related Work} \label{section: review}
While the determination of these physical parameters is critical for scientific considerations of ECs, no standardized process for this has emerged. Here we review previous related work, such as parameter estimation, S-curves, and noise event models. 
\par
\subsection{Various Parameter Estimation Methods}
Brandli \textit{et al.} presented a high-speed decompression algorithm for frame reconstruction when using a combined TC and EC, which, while not their end goal itself, involved estimating $B$ \citep{BrandliChristian2014Rhvd}. Their algorithm operated by converting each TC frame to log intensity as a baseline, then integrating polarity-dependent event ``steps'' in log space between frames. Those event steps were determined by attributing the log-space mismatch between the integrated reconstruction and the next TC frame to the counts of positive vs negative events, which both calibrates and regularizes the decoding. Conceptually, the TC frames acted as anchors, which helped initialization and periodically reset drift, while the EC contributed high-temporal-resolution updates. The drawbacks of this approach were that it required the use of a combined TC and EC (and so would not work with a standalone EC) and the observation of dynamic scenes. 
\par
Wang \textit{et al.} further developed this method for frame reconstruction to account for any per-pixel deviations from $B$ resulting from imperfect pixel resets, as well as to handle slowly varying $B$, though it still suffers the same drawbacks \citep{wang2020eventcameracalibrationperpixel}. Additionally, they presented an event-only method via linear regression of the cumulative polarity sum (using the $p=\pm1$ convention) vs.\ cumulative number of events, at a given pixel. They then utilized the RMSE of the regression to compute an estimate of the threshold. However, this only determines the relative magnitude of $\pm B$ and does not determine the precise value of $B$.
\par
Cabriel \textit{et al.} proposed the following approach to determine $B$ \citep{cabriel2023event}. First, the EC is stimulated by square wave light pulses from a laser source, modulated from known low intensity levels $I_0$ to varying high levels $I$. The average number of events per pixel per period, $N$, is then plotted against the relative modulation amplitude (high level intensity divided by the lower level), with the relative modulation amplitude on a log scale. A linear fit (on semi-log scale) was applied and $B$ was taken to be the inverse of the slope of the fit. 
\par
This was based on their notion that the average counts should have a linear relation with the relative modulation amplitude (the size of the intensity change), $N\approx1/B\ln(I/I_0)$. However, this is based on the expectation that a larger intensity change results in more event detections, which, as noted previously, is not the case. Even overlooking the aforementioned misconception, this approach requires a dynamic light source with intensity changes at microsecond precision. 
\par
Oliver and Savransky approached the parameter estimation of $B$ by developing an EC simulation capable of robustly simulating noise events and then tuning parameters until the simulated noise event rates aligned with observed event rates \citep{2024amos.conf...50O}. It is, however, not applicable to experimental S-curves for log contrasts measured at different light levels. 
\par
In addition to the discussed drawbacks of all the aforementioned methods, the primary focus is on the parameter $B$, and not on the others that we are interested in. Some authors, such as Oliver and Savransky, as well as McReynolds \textit{et al.}, do give some consideration to $\theta$, but only in the context of it being the dark current, and so do not capture the full scope of its nature \citep{2024amos.conf...50O, mcreynolds2024stepresponse}. McReynolds \textit{et al.} do touch on the conversion factor $\alpha$ briefly, but they do not elaborate further \citep{mcreynolds2024stepresponse}. 
\par
We note that the work of Cao \textit{et al.} involved parameter determination of $B$ and $\theta$ in their machine-learning-based frame reconstruction algorithm \citep{cao2025noise2image}. Like Brandli \textit{et al.} and Wang \textit{et al.}, the parameter determination was not the focus of their investigation. However, due to the significance of Cao \textit{et al.}'s consideration of noise events, we save discussion of their work for Section~\ref{subsection: related noise works}.
 
\subsection{Step Response Curves} \label{subsection: S-curves}
Step response curves, or S-curves, originally proposed by Posch and Matolin \citep{posch2011sensitivity}, have become a common approach to estimate parameters of event cameras, particularly $B$ \citep{finateu2020event, niwa2023sharedpixel, guo2023hybrid}. Stimuli are applied to the camera, typically in the form of square waves, with controlled lower and higher light levels ($I_0$ and $I$). The lower light level is kept fixed, while the higher light level is steadily increased until the logarithmic contrast range, $\ln(I/I_0)$, from 0 to 1 is covered. This procedure is then repeated for a range of lower light levels. We use the term Heaviside light to describe this sort of intensity transition, given that, at least in the ideal case, the intensity follows a Heaviside distribution from the lower level $I_0$ to a higher level $I$. We refer to the corresponding events as \textit{Heaviside events}. 
\par
The idea is that, at the edge of the square wave pulse (the rising edge for positive events and the falling edge for negative events), an ideal EC will only record an event once the log contrast surpasses $B$. Consider now the event probability, which here means the number of events within the time window of interest, per stimulus, per pixel. A plot of this event probability vs log contrast would result in a perfect step curve occurring at the log contrast threshold $B$. This is the S-curve itself. 
\par
We note that a critical requirement for this is that the refractory time of the EC pixels is sufficiently large relative to the transition time of the square wave pulse. If the transition time of the edge is longer than the refractory time, multiple events may be generated by the same pulse.
\par
However, real event cameras are non-ideal systems and, as such, S-curves deviate from the ideal case. Two distinct forms of S-curve distortion may occur: bending, where the step curve is ``bent'' or rotated (resulting in the characteristic S shape); or shifting, where the step is moved to the right or the left of the true $B$ value. The two forms of deviations may occur simultaneously. These distortions are broadly attributed to noise \citep{posch2011sensitivity, mcreynolds2024stepresponse}. 
\par
In light of these distortions, the traditional view is that the 50\% probability intercept is what should be regarded as $B$ \citep{posch2011sensitivity, finateu2020event, niwa2023sharedpixel, guo2023hybrid}. However, McReynolds \textit{et al.} \citep{mcreynolds2024stepresponse, mcreynolds2022experimental} have argued that it is where the S-curve approaches 100\% probability that should be used for $B$ instead. We show that our model can be used to derive S-curves theoretically (in Section~\ref{subsection: heaviside case}), which sheds light on the proper interpretation of S-curves as being more nuanced than picking the appropriate probability threshold (discussed in Section~\ref{subsubsection: S-curve interpretation}).
 
\subsection{Noise Events} \label{subsection: related noise works}
Many previously mentioned works have investigated noise events as part of their other aims \citep{mcreynolds2024stepresponse, cabriel2023event, 2024amos.conf...50O} or as a means of validating EC sensors \citep{posch2011sensitivity, finateu2020event, niwa2023sharedpixel, guo2023hybrid, IMX636sample}. Typically, observations of noise events have been given as noise event rates, events per pixel per second, for example, vs illuminance.
\par
Often, the reported observations of noise events are recorded over a region of interest (ROI) of pixels, instead of the entire sensor array. For example, McReynolds \textit{et al.} used a centered $100\times100$ ROI. While this approach is valid for drawing conclusions about the full sensor array if the sensor array behaves uniformly, there is no guarantee that it does. Thus, using a small ROI without validation of sufficiently uniform pixel behaviors may obfuscate underlying patterns and lead to skewed conclusions.
\par
While these works have provided a wealth of reference data regarding noise events, few considerations of noise events in the sense of noise probability distributions (and any analytic representation thereof) have been made. Cao \textit{et al.}, who were a source of inspiration for our work, are the closest prior work to an analytic probability-distribution treatment \citep{cao2025noise2image}. This was done by considering the events as being triggered by Poisson-distributed photon arrivals, as part of their Noise2Image method, which works to recreate frame-based images from accumulated noise events of a static scene.  
\par
While their work is influential, it has certain limitations that our approach addresses. In the construction of their probability distributions, they utilized a Gaussian approximation of the photon arrivals, which restricted its validity to only the mid to high intensities. So, while they did make a consideration of $\theta$, it was primarily motivated by correcting the otherwise erroneous behavior of the Gaussian-based distribution at low intensity levels. 
\par
Furthermore, all of the works mentioned in this section omit discussion of certain details, such as quantum efficiency, spectral considerations, and unaccounted-for experimental interferences (like flickering light sources), that may have significant impacts on noise event rates. It is known that the quantum efficiency of ECs can vary across the visible spectrum by as much as 50\% \citep{framos2023fsm, ids2025ue}, and therefore a discussion of the spectrum of light observed when recording noise events is critical for their correct interpretation. Light sources may flicker at rates imperceptible to humans, but, due to their high dynamic range, easily detectable by ECs, and can heavily skew event rates. Even factors such as the aperture and focus of the EC lens are essential for the reproduction of results, but are often omitted. 
 
\subsection{Outlier Pixels}
As noted, in a realistic, non-ideal EC, there will be variations in per-pixel event detection behaviors. At the extremes of this phenomenon are pixels whose behavior is so different from the ideal case as to represent extreme outliers in the data. This can range from pixels that report excessively many events, to pixels that report no events, to pixels that report non-physical observations (such as multiple events per timestep). Depending on the situation, it is often desirable to identify and remove these extreme outliers, as they can drastically skew event counts, and thus render erroneous any conclusions drawn from the resulting data.
\par
While such outliers are noted by some authors, such as Wang \textit{et al.} (who developed classifications for them), and by EC manufacturers \citep{prophesee2025mask}, many do not discuss them. Likewise, there appears to be no standard approach for the identification and handling of these outlier pixels. Differing approaches, or the lack of handling outlier pixels at all, may lead to differing results and misleading interpretations.

\section{Event Detection Probability Model} \label{section: theory}
 
Here we present the results of several different approaches to the construction of event probability distributions. We begin with the general case in which $I\ne I_0$, which can account for intensity changes. 
\subsection{Impact of Discrete Nature of Photons on Event Conditions}
Under sufficient spectral uniformity, the average photon count incident on a pixel, $\lambda$, is linearly proportional to the intensity (in lux) of the given light level, that is $\lambda=\alpha I$, where $\alpha$ is the conversion factor between lux and photon counts. This similarly holds for the reference intensity, $\lambda_0=\alpha I_0$.
\par
Given then that photon arrival is a Poisson process, the photon counts $n$ and $n_0$ are Poisson-distributed random variables, with the current and reference average photon counts as the Poisson parameters,
\begin{equation*}
    n\sim \text{Pois}(\lambda), \qquad n_0\sim\text{Pois}(\lambda_0)
\end{equation*}
Hence, the probability of a given number of photons arriving at the current and reference steps are $P(n)=\lambda^ne^{-\lambda}/n!$ and $P(n_0)=\lambda^{n_0}e^{-\lambda_0}/n_0!$. In terms of the apparent event condition, for positive events, we have now
\begin{equation}
    \ln\left(\frac{n}{n_0}\right) > B_{\text{app}},
\end{equation}
while the fundamental event condition is 
\begin{equation}
    \ln\left(\frac{n+\theta}{n_0+\theta}\right) > B,
\end{equation}
and similarly for negative events. We can then rearrange this into the random detection variables, $Z_\pm$, for positive and negative events respectively:
\begin{equation} \label{eq:Z_pm}
    Z_\pm=n-e^{\pm B} n_0+\theta-\theta e^{\pm B}.
\end{equation}
The probability of a positive event is the probability $P_+\equiv P(Z_+>0)$ and the probability of a negative event is the probability $P_-\equiv P(Z_-< 0)$. 
\par
 
\subsubsection{Intensity-Dependent Leakage}
We now make a critical assumption, that $\theta$ is not a constant but a function of the intensity level (and so, average photon counts), that is $\theta \rightarrow\theta(\lambda)$. For now, we need not establish its exact form, which is discussed in Section~\ref{section: param determ}. The fundamental event condition for positive events is then
\begin{equation}
    \ln\left(\frac{n+\theta(\lambda)}{n_0+\theta(\lambda_0)}\right)>B,
\end{equation}
and similarly for negative events. The detection variables then become 
\begin{equation}\label{eq:Z_pm lambda dependent}
    Z_\pm=n-e^{\pm B}\left(n_0+\theta(\lambda_0)\right)+\theta(\lambda).
\end{equation}
We note that we also investigated the case in which $\theta$ was a function of the photon counts $n,n_0$ directly, instead of the average photon counts $\lambda$. However, this yielded unsatisfactory results and so we adopt $\theta(\lambda),\,\theta(\lambda_0)$.
\par
From considerations of $Z_\pm$, there are different possible paths to establishing the distributions $P_\pm(\lambda,\lambda_0)$. We present the results of three approaches to establish the distributions, the derivations of which are shown in Section~\ref{s-section: derivation of distributions}. 
 
\subsubsection*{Exact Poisson Distribution}
The most straightforward approach is to treat $P_\pm(\lambda,\lambda_0)$ exactly by considering the fundamental Poisson nature underlying $Z_\pm$ (through $n, n_0$), which results in 
 
\begin{equation} \label{eq: Poisson dist pos}
    P_+^{\text{Pois}}=\!\sum_{n_0=0}^\infty \sum_{\substack{n= \floor{e^B\left(n_0+\theta(\lambda_0)\right)-\theta(\lambda)}+1}}^\infty \!\!\! \frac{e^{-2 \lambda}\lambda^{n+n_0}}{n!n_0!},
\end{equation}
and 
\begin{equation} \label{eq: Poisson dist neg}
    P_-^{\text{Pois}}=\!\sum_{n_0=0}^\infty \frac{e^{-\lambda}\lambda^{n_0}}{n_0!} \sum_{n=0}^{\ceil{e^B(n_0+\theta(\lambda_0))-\theta(\lambda)}-1} \!\frac{e^{-\lambda}\lambda^{n}}{n!}.
\end{equation}
 
An interesting aspect of these results, which is also shown in Section~\ref{s-section: derivation of distributions}, is that, despite their rather different forms, (\ref{eq: Poisson dist pos}) and (\ref{eq: Poisson dist neg}) are equivalent. That is, positive and negative noise events are equally likely, and so in principle, one can use $P_\pm(\lambda,\lambda_0)$ interchangeably. 
\par
 
\subsubsection*{Gaussian Approximation}
 
While (\ref{eq: Poisson dist pos}) and (\ref{eq: Poisson dist neg}) are exact, they are unwieldy and may be intensive to actually use. However, if one is only interested in the mid to high intensity regime, $\lambda,\lambda_0 \gtrsim 10$, $n$ and $n_0$ can be approximated as Gaussian-distributed random variables. This leads to the approximate distributions 
 
\begin{equation}
    P_\pm^{\text{Gaus}} \approx \frac{1}{2}\pm\frac{1}{2}\text{erf}\left(\frac{\lambda-e^{\pm B}(\lambda_0+\theta(\lambda_0))+\theta(\lambda)}{\sqrt{2(\lambda+e^{\pm 2B}\lambda_0)}}\right),
\end{equation}
    
 
where $\text{erf}$ denotes the error function. This was shown by Cao \textit{et al.} \citep{cao2025noise2image}, though they only considered the static $I=I_0\rightarrow\lambda=\lambda_0$ case.
\par
 
\subsubsection*{Saddle-Point Approximation}
\label{subsubsec: saddle_point_approximation}
 
We can also reach a more refined approximate solution, without having to restrict ourselves to a certain intensity regime, via the saddle-point approximation method \citep{daniels1954saddlepoint}, which results in 
 
\begin{equation}  \label{eq:saddle dist 0}
    P_\pm^{\text{saddle}}= \pm \frac{1}{ s_\pm\sqrt{2\pi\kappa_\pm''(s_\pm)}}e^{\kappa_\pm(s_\pm)},
\end{equation}
 
where $\kappa_\pm(t)$ is the cumulant generating function (CGF) of $Z_\pm$,
 

\begin{multline}
    \kappa_\pm(t)\;=\;\ln\left(\text{E}\left(e^{tZ_\pm}\right)\right) \\ =\lambda\left(e^t-1\right)+\lambda\left(e^{-te^{\pm B}}\!\!-1\right)\\+t\theta(\lambda)-te^{\pm B}\theta(\lambda_0),
\end{multline}
 
and $s_\pm$, the saddle point, must be found numerically by solving the saddle-point equation, $\kappa_\pm'(s_\pm)=0$.

\subsection{Heaviside and Static Scene Cases} \label{subsection: heaviside case}
The previously stated probability distributions are in their most general forms, and so naturally cover the case of idealized Heaviside step intensity changes. Despite any real changes in intensity having a finite transition time from one level to the other, if we assume the transition time is short enough relative to the refractory time as to not be impactful, we can neglect considerations of the transition time. Therefore, the probability distributions shown are sufficient to describe the case of S-curves.
\par
The case of static scene noise is a special case of these general distributions. The static scene nature implies $I=I_0 \Rightarrow\lambda=\lambda_0$, that is, there is no signal intensity change. It follows that in the static scene case, $n$ and $n_0$ are both governed by $\lambda$ and the leakage terms then all become $\theta(\lambda)$. Thus the probability distribution expressions simplify slightly. 
\par
\begin{figure*}
    \centering
    \includegraphics[width=0.7\linewidth]{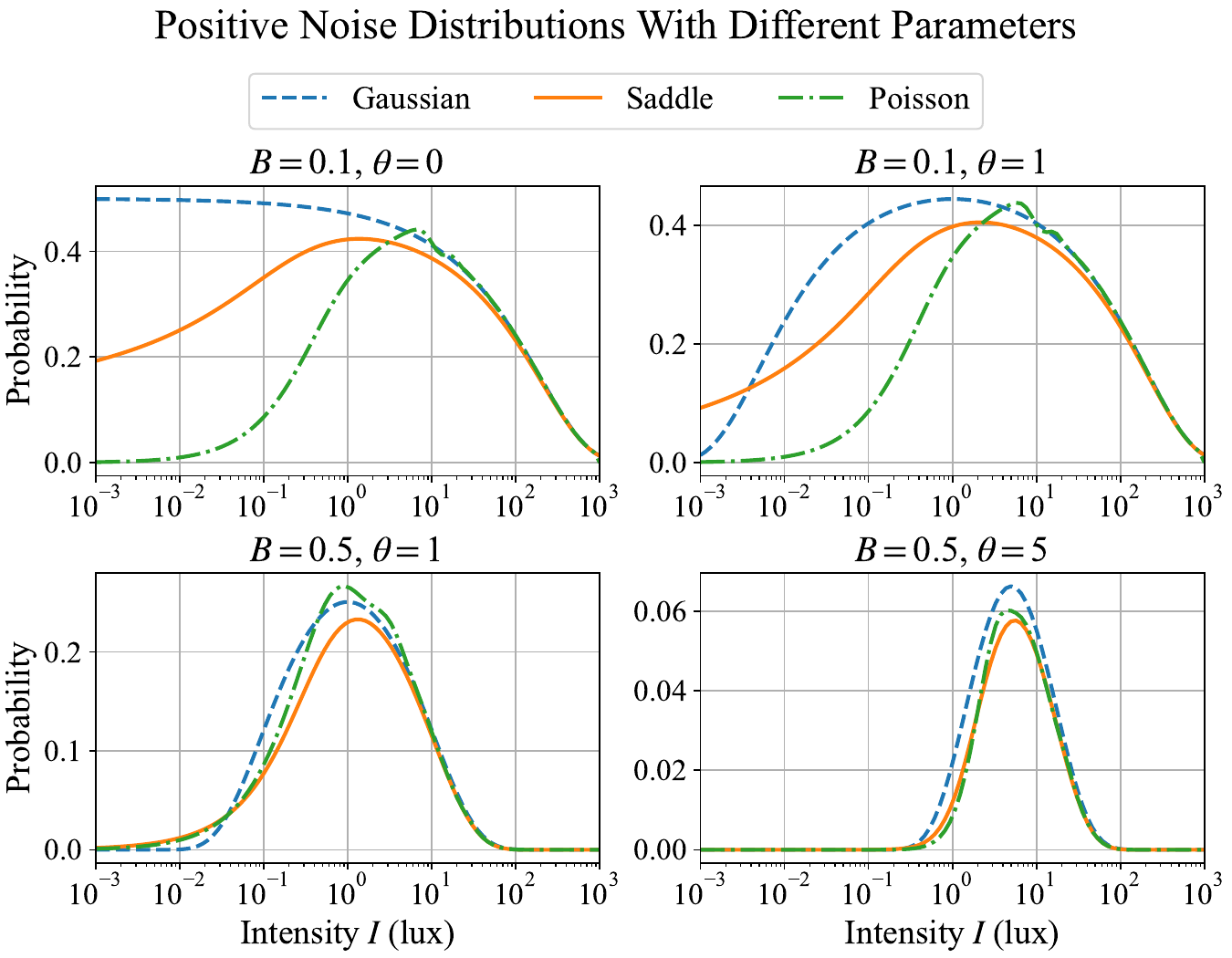}
    \caption{The different positive noise event probability distributions for the static scene case, with $\alpha=1$ and different values of $B$ and $\theta$, with $\theta$ is taken to be a constant for illustrative purposes. Comparing the magnitudes of the distribution peaks between the top and bottom rows, it can be seen that increasing $B$ lowers the event probability while preserving the position of the peak. Increasing $\theta$ shifts the position of the peak and its magnitude. Notice that in the $\theta=0$ case, only the saddle-point approximation and the Poisson distribution result in the physically expected case of going to 0 as intensity goes to 0.}
    \label{fig:different noise dists}
\end{figure*}
\begin{figure*}[h]
    \centering
    \includegraphics[width=0.9\linewidth]{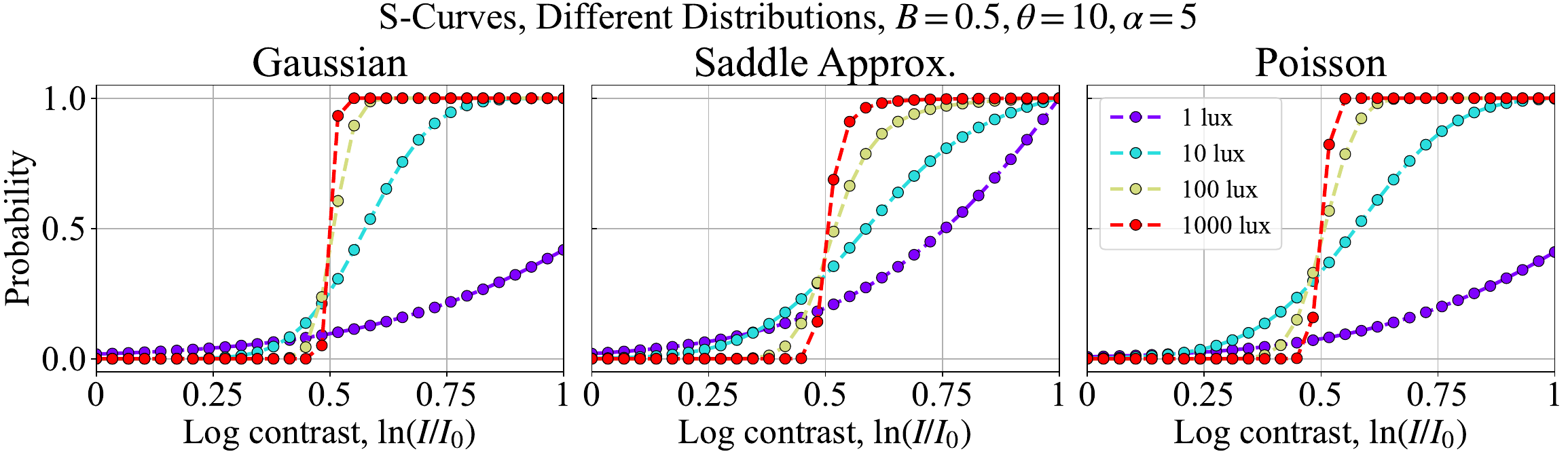}
    \caption{Example computed Heaviside event probability distributions, or computed S-curves, for each of the different approaches, for example parameter values of $B=0.5, \theta=10$ and $\alpha=5$. Again, $\theta$ is taken to be a constant for illustrative purposes.}
    \label{fig:S-curves diff dist}
\end{figure*}

Examples of the distributions for the static scene case can be seen in Fig.~\ref{fig:different noise dists}, while examples of the distributions for the Heaviside step intensity S-curve case can be found in Fig.~\ref{fig:S-curves diff dist}.
 
\subsection{Distribution Properties, Strengths, and Weaknesses}
The three distribution approaches have distinct trade-offs between accuracy and computational cost; we summarize their key properties here.
 
The exact Poisson distributions are valid across all intensity regimes, which is critical for describing EC behavior at the low-light limit. However, they require numerical truncation to evaluate, with the appropriate truncation index depending on $B$, $\theta(\lambda)$, $\lambda$, and the desired precision. Additionally, because $\theta$ depends on $\lambda$, the floor and ceiling functions in the summation bounds can produce discontinuities in the distribution; we revisit this phenomenon in Section~\ref{section: param determ}.
 
The Gaussian distributions are computationally inexpensive but are only valid for $\lambda \gtrsim 10$, limiting their applicability at low light levels. For sufficiently large $\lambda$, they agree well with the exact Poisson distributions. Notably, the Gaussian distributions require non-zero $\theta$ in order to vanish as $\lambda \to 0$ (see Fig.~\ref{fig:different noise dists}), which is the physically expected behavior when no photons arrive. Whereas the saddle-point and Poisson distributions vanish at $\lambda = 0$ regardless of $\theta$ This illustrates that accurate inference of $\theta$ \textit{requires} the saddle-point or Poisson distributions.
 
The saddle-point distributions offer a practical middle ground: they remain valid across all intensity regimes while requiring only a single numerical root-finding operation (to determine $s_\pm$). Their peak locations also agree better with the exact Poisson distributions than do the Gaussian approximations.
 
Because the Gaussian and saddle-point distributions are approximate, normalization may need to be enforced. For Heaviside events this is straightforward, since the maximum probability is 1. For noise events, normalization requires $P_+ + P_- + P_0 = 1$ at each $\lambda$, where $P_0$ is the null-event probability; however, comparisons with the exact Poisson distributions indicate that such enforcement is seldom necessary.
 
Finally, we note the distinct roles of the parameters for static scene probability distributions. Within typical ranges, $B$ primarily affects the distribution magnitude while leaving the peak location unchanged (compare the upper-right and lower-left panels of Fig.~\ref{fig:different noise dists}). In contrast, $\theta$ more readily shifts the peak horizontally (compare the lower two panels). Both parameters influence the distribution width. We discuss the parameters' impact on the nature of the S-curve in greater detail in Section~\ref{subsubsection: S-curve interpretation}.

\section{Experimental Observations} \label{section: experiments}
 
\subsection{Setup and Data Processing} \label{subsection: setup}
\begin{figure}
    \centering
    \definecolor{screenblue}{RGB}{100,150,220}
    \definecolor{lightcone}{RGB}{255,245,200}
    \definecolor{coneedge}{RGB}{220,200,80}
    \definecolor{camerabody}{RGB}{50,50,50}
    \definecolor{cameralens}{RGB}{70,70,70}
    \definecolor{surfacegray}{RGB}{210,210,210}
 
    \pgfdeclarehorizontalshading{screenglow}{100bp}{%
        color(0bp)=(white);
        color(60bp)=(screenblue!18);
        color(85bp)=(screenblue!40);
        color(100bp)=(screenblue!55)
    }
 
    \begin{tikzpicture}[
        transform shape,
        every node/.style={font=\small},
        label/.style={font=\small, align=center},
        sublabel/.style={font=\normalsize\bfseries, anchor=north west},
        arrow/.style={thick, ->, >=Stealth},
        fov/.style={thick, black!60, dashed},
    ]
 
    \pgfmathsetmacro{\scaleA}{1}
    \pgfmathsetmacro{\scaleBC}{0.5}
    \pgfmathsetmacro{\vgap}{1}          
    \pgfmathsetmacro{\hgap}{0}
 
    \pgfmathsetmacro{\localWbc}{6.5}
    \pgfmathsetmacro{\localHbc}{7.2}
 
    \pgfmathsetmacro{\panelBCw}{\localWbc * \scaleBC}
    \pgfmathsetmacro{\panelBCh}{\localHbc * \scaleBC-0.75}
    \pgfmathsetmacro{\bottomW}{2*\panelBCw + \hgap}
 
    \pgfmathsetmacro{\shiftBx}{0}
    \pgfmathsetmacro{\shiftBy}{0}
    \pgfmathsetmacro{\shiftCx}{\panelBCw + \hgap}
    \pgfmathsetmacro{\shiftCy}{0}
 
    \pgfmathsetmacro{\panelAcontentW}{5.0 * \scaleA}
    \pgfmathsetmacro{\shiftAx}{(\bottomW - \panelAcontentW)/2}
    \pgfmathsetmacro{\shiftAy}{\panelBCh + \vgap}
 
    \newcommand{\eventcamera}{%
        \fill[black, rounded corners=0.5pt] (0,1) rectangle (1,2);
        \fill[black, rounded corners=0.5pt] (0.85,1.12) rectangle (1.55,1.88);
        \draw[fov] (1.83,1.78) -- (3.1,2.25);
        \draw[fov] (1.83,1.22) -- (3.1,0.75);
        \node[label, anchor=north] at (1.5,-0.15) {Event camera};
    }

    \newcommand{\eventcameraTop}{%
        \fill[black, rounded corners=0.5pt] (0,1) rectangle (1,2);
        \fill[black, rounded corners=0.5pt] (0.85,1.12) rectangle (1.55,1.88);
        \draw[fov] (1.83,1.78) -- (3.1,2.25);
        \draw[fov] (1.83,1.22) -- (3.1,0.75);
        \node[label, anchor=north] at (0.6,0.2) {Event camera};
    }
 
    \begin{scope}[
        scale=\scaleA,
        shift={(\shiftAx/\scaleA, \shiftAy/\scaleA)},
        every node/.style={font=\scriptsize},
        label/.style={font=\small, align=center},
        sublabel/.style={font=\small\bfseries, anchor=north west},
    ]
        \node[sublabel] at (-0.75,3.5) {(a)};
 
        \begin{scope}
            \clip (2.3,-0.5) -- (3.55,-0.3) -- (3.55,3.3) -- (2.3,3.5) -- cycle;
            \shade[shading=screenglow, shading angle=0]
                (2.3,-0.5) rectangle (3.55,3.5);
        \end{scope}
        \fill[black!70, rounded corners=0.5pt] (3.95,-0.35) rectangle (4.2,3.35);
        \fill[black!55] (3.75,-0.35) rectangle (3.98,3.35);
        \fill[screenblue!60, draw=black!50, rounded corners=0.3pt]
            (3.55,-0.35) rectangle (3.78,3.35);
        \fill[screenblue!90] (3.55,-0.3) rectangle (3.60,3.3);
 
        \node[label, anchor=south] at (4.875,1) {Monitor\\[-2pt]screen};
        \eventcameraTop
    \end{scope}
 
    \pgfmathsetmacro{\delinY}{\panelBCh}
    \draw[black!25, dashed]
        (-0.3, \delinY) -- ({\bottomW + 0.3}, \delinY);
 
    \begin{scope}[
        scale=\scaleBC,
        shift={(\shiftBx/\scaleBC, \shiftBy/\scaleBC)},
        every node/.style={font=\LARGE},
        label/.style={font=\LARGE, align=center},
        sublabel/.style={font=\huge\bfseries, anchor=north west},
    ]
        \node[sublabel] at (-1.5,5.5) {(b)};
 
        \fill[lightcone]
            (1.52,4.151) -- (3.5,3) -- (3.5,0) -- (1.03,3.57) -- cycle;
        \draw[coneedge] (1.52,4.151) -- (3.5,3);
        \draw[coneedge] (1.03,3.57)  -- (3.5,0);
 
        \begin{scope}[rotate around={-40:(0.5,4.525)}]
            \fill[camerabody!60, draw=black!70, thick, rounded corners=1pt]
                (-0.25,4.25) rectangle (1.25,4.8);
            \fill[camerabody!60, draw=black!70, thick]
                (1.25,4.8) -- (1.5,4.9) -- (1.5,4.15) -- (1.25,4.25) -- cycle;
            \draw[black!50, fill=white, thick] (2,3.75) rectangle (2.15,5.25);
        \end{scope}
 
        \eventcamera
 
        \fill[surfacegray!40, draw=black!70, thick] (3.5,0) rectangle (3.65,3);
 
        \node[label, anchor=west] at (3.85,1) {White\\[-2pt]surface};
        \node[label] (ctrllight) at (5,5) {Control\\[-2pt]light};
        \draw[arrow] (ctrllight.west) -- (1.0,5.1);
        \node[label] (diffuser) at (5.3,3.4) {Diffuser};
        \draw[arrow] (diffuser.west) -- (2.5,4);
    \end{scope}
 
    \pgfmathsetmacro{\delinX}{\panelBCw + \hgap/2}
    \draw[black!25, dashed]
        (\delinX, -0.3) -- (\delinX, \panelBCh);
 
    \begin{scope}[
        scale=\scaleBC,
        shift={(\shiftCx/\scaleBC+0.5, \shiftCy/\scaleBC)},
        every node/.style={font=\LARGE},
        label/.style={font=\LARGE, align=center},
        sublabel/.style={font=\huge\bfseries, anchor=north west},
    ]
        \node[sublabel] at (-0.1,5.5) {(c)};
 
        \fill[lightcone]
            (2,3) -- (4.95,1.9) -- (4.95,1.1) -- (2,0) -- cycle;
        \draw[coneedge] (2,3)   -- (4.95,1.9);
        \draw[coneedge] (2,0)   -- (4.95,1.1);
 
        \coordinate (flashlight) at (6,1.5);
        \begin{scope}[shift={(flashlight)}, rotate around={180:(0,0)}]
            \fill[camerabody!60, draw=black!70, thick, rounded corners=1pt]
                (-0.75,-0.275) rectangle (0.75,0.275);
            \fill[camerabody!60, draw=black!70, thick]
                (0.75,0.275) -- (1.0,0.375) -- (1.0,-0.375) -- (0.75,-0.275);
        \end{scope}
 
        \eventcamera
 
        \draw[black!50, fill=white, thick] (3.5,0.4) rectangle (3.65,2.6);
 
        \node[label] (ctrlC) at (5.5,4) {Control\\[-2pt]light};
        \draw[arrow] (ctrlC.south) -- (6.2,2.0);
        \node[label] (Diffuser) at (2.2,3.7) {Diffuser};
        \draw[arrow] (Diffuser.east) -- (3.62,2.7);
    \end{scope}
 
    \end{tikzpicture}
 
    \caption{Schematic diagrams of the top-down view of the three experimental setups employed:
    (a)~the EC viewing a monitor screen,
    (b)~the EC viewing a uniform white surface evenly lit by the control light source through a diffuser, and
    (c)~the control light source viewed directly through a diffuser by the EC. In all three cases the room was otherwise dark. The majority of the useful data employed was captured using setup~(a); the other two, (b) and (c), were used to calibrate (a) and provide observations of higher intensities than what the monitor could achieve.}
    \label{fig:experimental_set_up}
\end{figure}
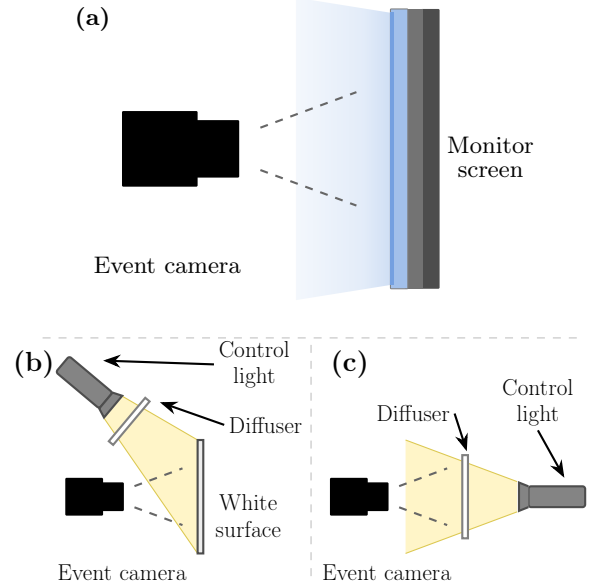

Figure~\ref{fig:experimental_set_up} shows the experimental setup, as described in the following paragraphs. Static scenes were captured by an EVK4 HD camera (Prophesee, US) with the aperture at the most open setting and the focus at the most near setting. The static scenes were primarily constructed by filling the EC's field of view entirely with a commercial back-lit LED screen (average wavelength of $\approx550$ nm when displaying white), at varying display brightness settings, with different uniform greyscale images displayed, so as to record noise events at varying intensities. The room was otherwise kept dark.
\par
To ensure no flickering or other unexpected effects were present, an incandescent bulb on DC battery power (Maglite 6-Cell D Xenon Flashlight, peak wavelength of $\approx666$ nm based on a color temperature of $\approx3000$K) was used as a control. Static scenes from the control source were generated via illumination of a white surface, through a diffuser to produce even illuminance, which was then viewed by the EC, or by direct viewing of the diffuser. Both screen and control light sources gave comparable noise event probabilities at similar intensities (Section~\ref{s-section: light sources}). 
\par
The EC has a $1280\times720$ pixel array with a pixel size of 4.86 $\mu \text{m}^2$ and operates with a timestamp resolution of 1 $\mu$s. A centered region of interest (ROI) of $640\times360$ was used to avoid potential distortions due to the lens at the edge of the frame. Recordings of varying length (10--15 s) were recorded using Prophesee Metavision Studio software \citep{prophesee2025metavisionstudio} or a custom Python code based on Prophesee's open-source library OpenEB \citep{prophesee2025openeb}. Default bias settings were used unless otherwise indicated. The default value of all bias settings is 0 for our camera. Outlier pixels were identified and removed via the processes discussed in Section~\ref{s-section: hot pixels}. The on-chip illuminance, in lux, for a given static scene was self-reported by the camera's built-in illuminometer. The refractory period $R$ was also self-reported by the camera, with a value of $R=79 \;\mu\text{s}$ at default bias settings. 
 
\subsection{Measured Noise Distributions}
To determine the estimated event probabilities $\hat{P}_\pm$ (probability of an event, at one pixel, in one timestep) from the raw event counts of a recording, we use 
\begin{equation}\label{eq: prob estimate}
    \hat{P}_\pm=\frac{N_\pm}{TM-RN_\text{tot}},
\end{equation}
where $N_\pm$ is the total number of positive or negative events, $T$ is the total length of the recording (in $\mu$s, the size of one timestamp), $M$ is the number of pixels in the array (total pixel count minus any outliers that have been removed), and $N_\text{tot}$ is the total number of events. The $-RN_\text{tot}$ term accounts for the fact that event detections cause pixels to be blind for the refractory period, thus reducing the total allowed opportunities for detections.
\par
\begin{figure}
  \centering
  \fontsize{13pt}{9pt}\selectfont{Observed Noise Event Probabilities, Temporal Variance}
  \includegraphics[width=0.5\textwidth]{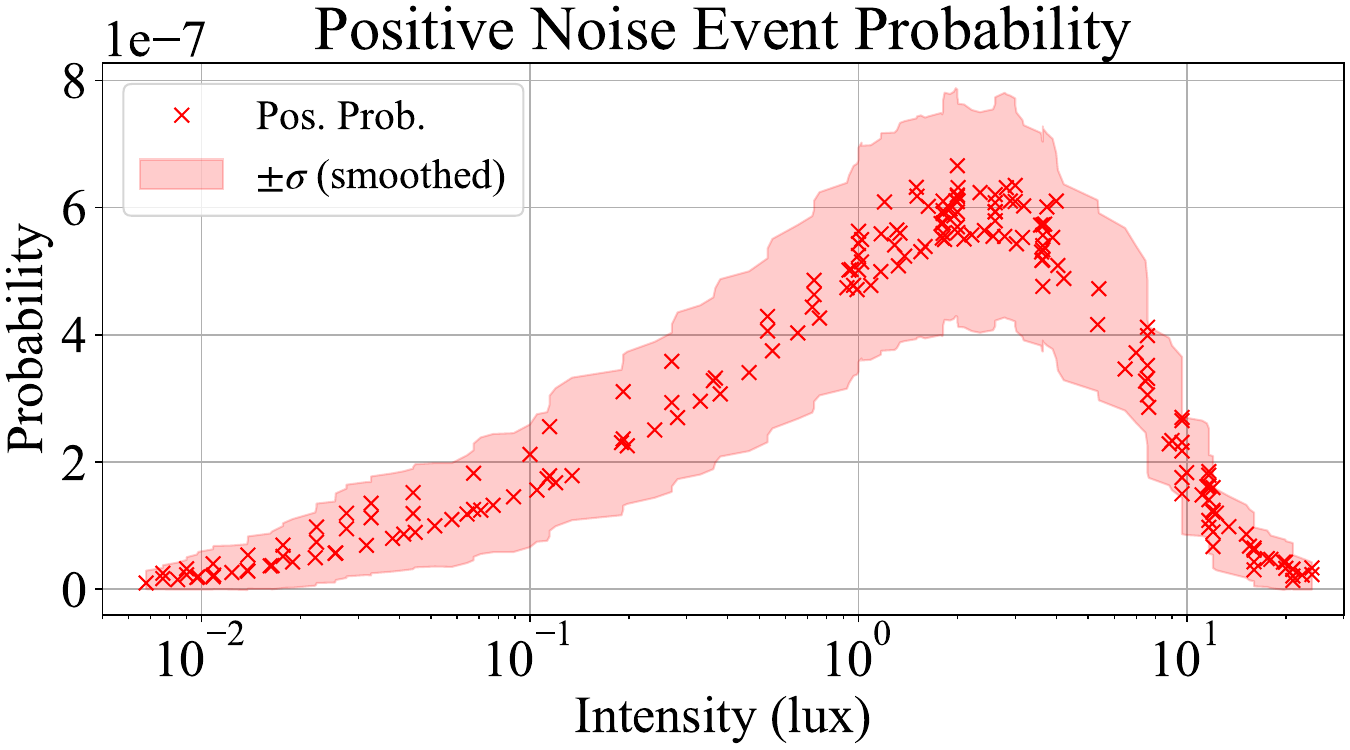}
  \includegraphics[width=0.5\textwidth]{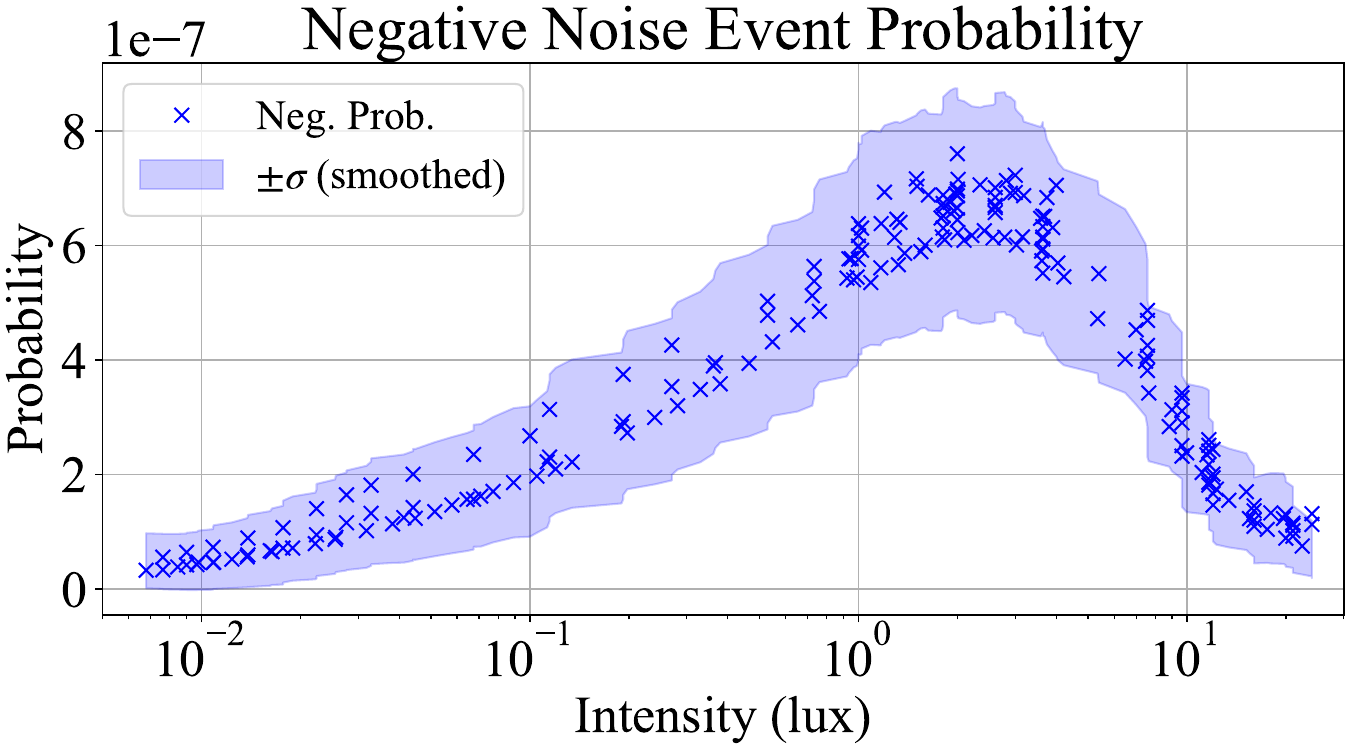}
  \caption{Observed $\hat{P}_\pm$ with temporal variances. While both have the same shape, the magnitude of $\hat{P}_-$ is slightly larger than $\hat{P}_+$. The uncertainty bands were smoothed using a Savitzky-Golay filter (window size 25, order 2) for visual clarity.}
  \label{fig:_noise_prob_default}
\end{figure}
\begin{figure}
  \centering
  \fontsize{13pt}{9pt}\selectfont{Observed Noise Event Probabilities, Spatial Variance}
   \includegraphics[width=0.5\textwidth]{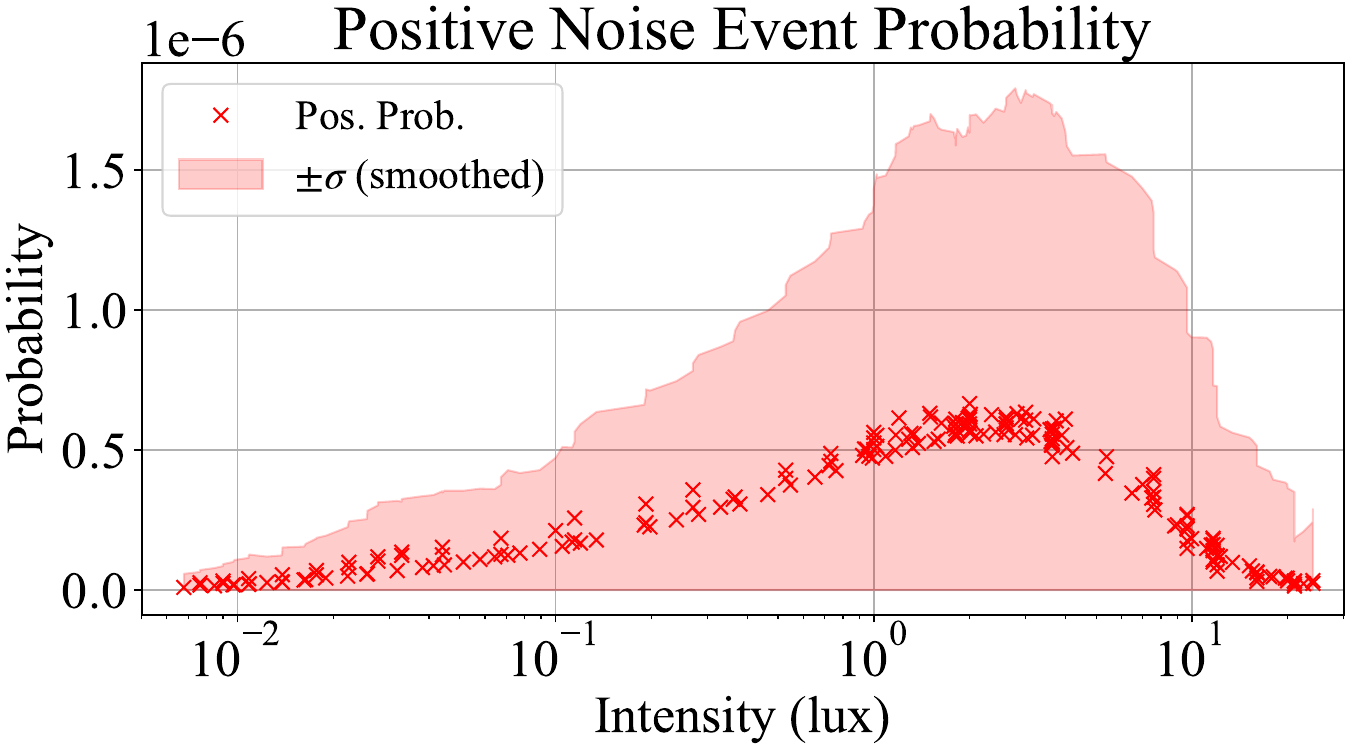}
   \includegraphics[width=0.5\textwidth]{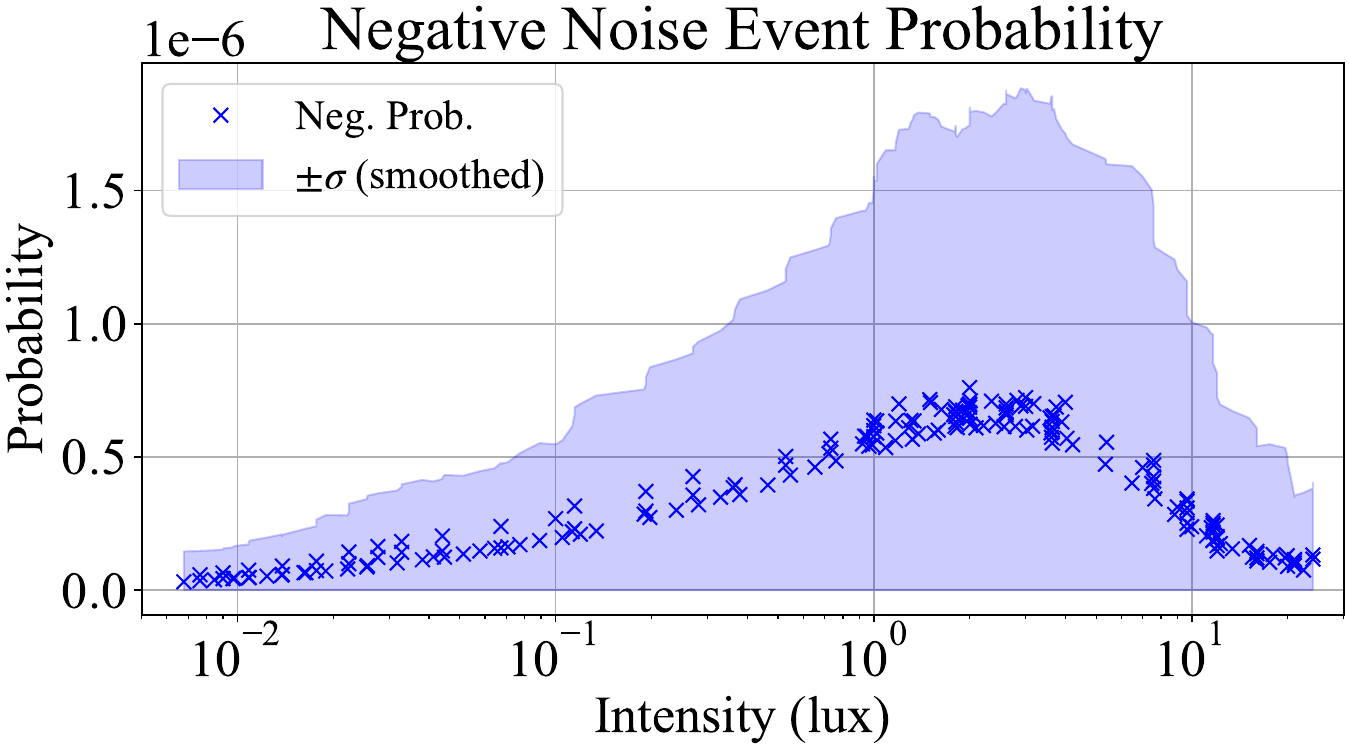}
 
  \caption{Observed $\hat{P}_\pm$ with spatial variances. The uncertainty bands here were smoothed by the same means as the temporal case above.}
  \label{fig:_noise_prob_default_spatial_var}
\end{figure}

The observed noise probability distributions, for default bias settings, with temporal variations are shown in Fig.~\ref{fig:_noise_prob_default} and with spatial variations in Fig.~\ref{fig:_noise_prob_default_spatial_var}. The non-default bias setting cases are discussed in Section~\ref{s-section: non-default settings}. For temporal variations, the standard deviations of $\hat{P}_\pm$ were found by dividing a recording into bins of 100 $\mu$s (discarding remainders), and finding the standard deviations of the probabilities computed for each bin. A bin size of 100 was chosen based on an elbow method analysis of the relative standard deviations vs bin size, as seen in Fig.~\ref{s-fig:elbow method}. This behavior was observed to hold across all intensities and bias settings. For spatial variations, the probability was estimated per pixel and the standard deviation was taken across all the pixels in the ROI. The magnitude of the spatial variances is greater than the temporal variances. That is, the pixels vary greatly in sensitivity across the array, but their behavior is consistent over time. 
\par
 
\subsection{Reported S-Curves}
\begin{figure}
    \centering
    \includegraphics[width=1\linewidth]{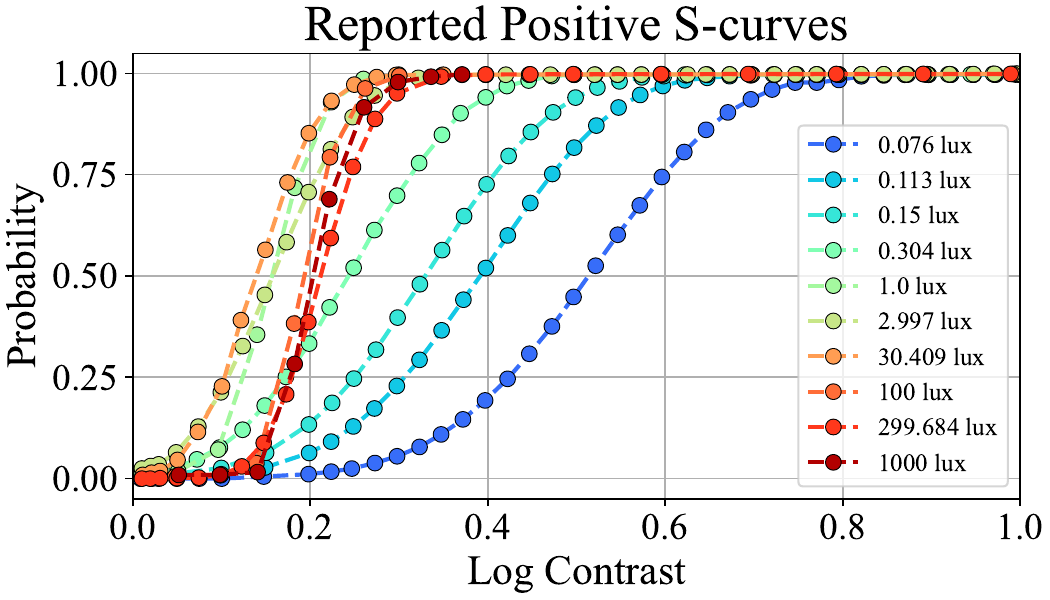}
    \caption{Reported positive event S-curves, for the same EC model compiled from \citep{mcreynolds2024stepresponse, IMX636sample}.}
    \label{fig:reported s-curves}
\end{figure}

We include a compilation of reported S-curves for positive events \citep{mcreynolds2024stepresponse, IMX636sample}, which can be found in Fig. \ref{fig:reported s-curves}. The S-curves for negative events are similar. These sources collected the S-curve data using the same model of EC that we have used, or a test sample if it, though smaller ROIs were used instead of the entire sensor array. Neither source discusses accounting for outlier pixels, which may be the cause of the leftward shift of S-curves on the order of 1 lux.

\section{Noise2Params: Parameter Estimation} \label{section: param determ}
 
\begin{figure*} 
    \centering
    \newsavebox{\StaticScenes}
    \newsavebox{\NoiseDistribution}
    \newsavebox{\SquareWaveStimuli}
    \newsavebox{\SCurveFamily}
     
    \newcommand{\fontone}{\fontsize{14pt}{16pt}\selectfont}
    \newcommand{\fonttwo}{\fontsize{13pt}{14pt}\selectfont}
    \newcommand{\fontthree}{\fontsize{10pt}{11pt}\selectfont}
    \savebox{\StaticScenes}{%
    \begin{tikzpicture}[scale=1]
        \filldraw[fill=black!90, draw=black!70] (0,0) -- (1,1) -- (1,3) -- (0,2) -- cycle;
        \filldraw[fill=black!60, draw=black!50] (0.5,0) -- (1.5,1) -- (1.5,3) -- (0.5,2) -- cycle;
        \filldraw[fill=black!25, draw=black!40] (1,0) -- (2,1) -- (2,3) -- (1,2) -- cycle;
        \filldraw[fill=white, draw=black!40] (1.5,0) -- (2.5,1) -- (2.5,3) -- (1.5,2) -- cycle;
        \node[align=center, font=\fontone] at (1.25,3.7) {\textbf{Static uniform scenes}\\ \textbf{at different intensities}};
    \end{tikzpicture}%
    }
     
    
    \savebox{\NoiseDistribution}{%
    \begin{tikzpicture}[scale=1]
        \draw[thick, ->] (0,0.25) -- (0,2.9);
        \draw[thick, ->] (0,0.25) -- (4.5,0.25);
        \node[font=\fontone, rotate=90, anchor=south] at (-0.35,1.5) {$P(\text{noise})$};
        \node[font=\fontone, anchor=north] at (2.25,-0.5) {Intensity (lux)};
        \foreach \y in {0.5, 1.0, 1.5, 2.0, 2.5} { \draw (0,\y) -- (-0.08,\y); }
        \node[font=\fonttwo, anchor=north] at (0.25,0.15) {$10^{-2}$};
        \node[font=\fonttwo, anchor=north] at (1.5,0.15) {$10^{-1}$};
        \node[font=\fonttwo, anchor=north] at (2.75,0.15) {$10^{0}$};
        \node[font=\fonttwo, anchor=north] at (4.0,0.15) {$10^{1}$};
        \foreach \x in {0.25, 1.5, 2.75, 4.0} { \draw (\x,0.25) -- (\x,0.08); }
        \pgfmathsetmacro{\xmin}{0.25}
        \pgfmathsetmacro{\xmax}{4.0}
        \pgfmathsetmacro{\decades}{3}
        \foreach \decade in {0,1,2} {
            \foreach \minor in {2,3,4,5,6,7,8,9} {
                \pgfmathsetmacro{\xpos}{\xmin + (\xmax-\xmin) * (ln(\minor)/ln(10) + \decade) / \decades}
                \draw (\xpos,0.25) -- (\xpos,0.17);
            }
        }
        \foreach \point in {
            (0.25,0.38), (0.45,0.40), (0.65,0.42), (0.85,0.45), (1.05,0.50),
            (1.25,0.58), (1.45,0.70), (1.65,0.90), (1.85,1.20), (2.05,1.55),
            (2.25,1.85), (2.45,2.05), (2.65,2.05), (2.85,1.85), (3.05,1.50),
            (3.25,1.10), (3.45,0.75), (3.65,0.55), (3.85,0.45), (4.05,0.40)
        } { \fill[color=red] \point circle (1.8pt); }
        \foreach \point in {
            (0.25,0.42), (0.45,0.45), (0.65,0.48), (0.85,0.52), (1.05,0.60),
            (1.25,0.72), (1.45,0.90), (1.65,1.15), (1.85,1.50), (2.05,1.90),
            (2.25,2.25), (2.45,2.45), (2.65,2.40), (2.85,2.10), (3.05,1.65),
            (3.25,1.20), (3.45,0.85), (3.65,0.62), (3.85,0.50), (4.05,0.44)
        } { \fill[color=blue] \point circle (1.8pt); }
        \node[anchor=west, font=\fontone] at (3.6,2.6) {
            \tikz[baseline=-0.5ex]{\fill[red] (0,0) circle (1.5pt);} Pos.
        };
        \node[anchor=west, font=\fontone] at (3.6,2.1) {
            \tikz[baseline=-0.5ex]{\fill[blue] (0,0) circle (1.5pt);} Neg.
        };
        \node[font=\fontone] at (2.0,3.35) {\textbf{Noise event distribution}};
    \end{tikzpicture}%
    }
     
    \savebox{\SquareWaveStimuli}{%
    \begin{tikzpicture}[
        scale=1,
        axislabel/.style={font=\fontone},
        curveA/.style={color={rgb,255:red,255;green,128;blue,0}, line width=1.8pt},
        curveC/.style={color={rgb,255:red,191;green,0;blue,64}, line width=1.8pt},
        planefill/.style={fill=white, fill opacity=0.95},
        planeedge/.style={draw=black!50, thin}
    ]
        \draw (1.4,3.8) node [align=center, font=\fonttwo]{\textbf{Square-wave} \\ \textbf{Intensity Stimuli}};
        \pgfmathsetmacro{\xscale}{0.5}
        \pgfmathsetmacro{\yscale}{0.65}
        \pgfmathsetmacro{\shearfactor}{0.4}
        \pgfmathsetmacro{\depthstep}{1.8}
        \pgfmathsetmacro{\IoLow}{0.5}
        \pgfmathsetmacro{\IoHigh}{1.3}
        \pgfmathsetmacro{\plotW}{4}
        \pgfmathsetmacro{\plotHmin}{0.17}
        \pgfmathsetmacro{\plotHmax}{3.15}
        \begin{scope}[cm={\xscale, \shearfactor*\xscale, 0, \yscale, (0, 0)}]
            \fill[planefill] (-0.1,\plotHmin) rectangle (\plotW+0.1, \plotHmax+0.1);
            \draw[planeedge] (\plotW+0.1, \plotHmin) -- (\plotW+0.1, \plotHmax+0.1) -- (-0.1, \plotHmax+0.1);
            \node[rotate=24] at (2,3.6) {\fonttwo $I_0$ High};
            \draw[thick, ->] (0,0.25) -- (0,3.15);
            \draw[thick, ->] (0,0.25) -- (4,0.25);
            \foreach \y in {0.5, 1.0, 1.5, 2.0, 2.5} { \draw (0,\y) -- (-0.15,\y); }
            \node[anchor=east, font=\fonttwo, fill=white, fill opacity=0.8, text opacity=1, inner sep=3pt] at (-0.1,\IoHigh) {$I_0$};
            \foreach \x in {0.5, 1.0, 1.5, 2.0, 2.5, 3.0, 3.5} { \draw (\x,0.25) -- (\x,0.1); }
            \draw[curveA, line width=1.2pt] 
                (0.25,\IoHigh) -- (1,\IoHigh) -- (1,1.5) -- (1.75,1.5) -- (1.75,\IoHigh) -- 
                (2.5,\IoHigh) -- (2.5,2.5) -- (3.25,2.5) -- (3.25,\IoHigh) -- (3.75,\IoHigh);
        \end{scope}
        \begin{scope}[cm={\xscale, \shearfactor*\xscale, 0, \yscale, (\depthstep, 0)}]
            \fill[planefill] (-0.1,\plotHmin) rectangle (\plotW+0.1, \plotHmax+0.1);
            \draw[planeedge] (\plotW+0.1, \plotHmin) -- (\plotW+0.1, \plotHmax+0.1) -- (-0.1, \plotHmax+0.1);
            \node[rotate=24] at (2,3.6) {\fonttwo $I_0$ Low};
            \draw[thick, ->] (0,0.25) -- (0,3.15);
            \draw[thick, ->] (0,0.25) -- (4,0.25);
            \foreach \y in {0.5, 1.0, 1.5, 2.0, 2.5} { \draw (0,\y) -- (-0.12,\y); }
            \node[anchor=east, font=\fonttwo, fill=white, fill opacity=0.8, text opacity=1, inner sep=3pt] at (-0.1,\IoLow) {$I_0$};
            \foreach \x in {0.5, 1.0, 1.5, 2.0, 2.5, 3.0, 3.5} { \draw (\x,0.25) -- (\x,0.15); }
            \draw[curveC, line width=1.2pt] 
                (0.25,\IoLow) -- (1,\IoLow) -- (1,1.5) -- (1.75,1.5) -- (1.75,\IoLow) -- 
                (2.5,\IoLow) -- (2.5,2.5) -- (3.25,2.5) -- (3.25,\IoLow) -- (3.75,\IoLow);
        \end{scope}
        \node[axislabel, rotate=90, anchor=south] at (-0.45, 1.1) {Intensity};
        \node[axislabel, anchor=north] at (3.25, 0.35) {Time};
    \end{tikzpicture}%
    }
     
    \newcommand{\scurvepoint}{1.8pt}
    \newcommand{\scurvethick}{1.4pt}
    \savebox{\SCurveFamily}{%
    \begin{tikzpicture}[
        scale=1,
        axislabel/.style={font=\fontone},
        curveA/.style={color={rgb,255:red,255;green,128;blue,0}, line width=\scurvethick},
        curveC/.style={color={rgb,255:red,191;green,0;blue,64}, line width=\scurvethick},
        datapointA/.style={curveA, fill={rgb,255:red,255;green,128;blue,0}},
        datapointC/.style={curveC, fill={rgb,255:red,191;green,0;blue,64}}
    ]
        \draw[thick, ->] (0.1,0) -- (0.1,2.5);
        \draw[thick, ->] (0.1,0) -- (3.5,0);
        \node[axislabel, rotate=90, anchor=south] at (-0.25,1.15) {$P(\text{event})$};
        \node[axislabel, anchor=north] at (1.75,-0.15) {$\ln(I/I_0)$};
        \node[anchor=east, font=\fonttwo] at (0.05,0.25) {0};
        \node[anchor=east, font=\fonttwo] at (0.05,2.15) {1};
        \draw[curveA, smooth, tension=0.7] plot coordinates {
            (0.1,0.25) (0.4,0.25) (0.6,0.35) (0.75,0.7) 
            (0.95,1.4) (1.1,1.9) (1.3,2.1) (1.7,2.15) (2.5,2.15) (3.3,2.15)
        };
        \foreach \point in {(0.4,0.25), (0.6,0.35), (0.75,0.7), (0.95,1.4), (1.1,1.9), (1.3,2.1), (1.7,2.15), (2.5,2.15)} {
            \fill[datapointA] \point circle (\scurvepoint);
        }
        \draw[curveC, smooth, tension=0.7] plot coordinates {
            (0.1,0.25) (0.6,0.25) (1.0,0.25) (1.3,0.3) (1.6,0.6) 
            (1.85,1.1) (2.1,1.65) (2.35,2.0) (2.6,2.12) (3.3,2.15)
        };
        \foreach \point in {(0.6,0.25), (1.0,0.25), (1.3,0.3), (1.6,0.6), (1.85,1.1), (2.1,1.65), (2.35,2.0), (2.6,2.12)} {
            \fill[datapointC] \point circle (\scurvepoint);
        }
        \node[anchor=west, font=\fonttwo] at (2.9,1.2) {
            \tikz[baseline=-0.5ex]{\draw[curveA] (0,0) -- (0.35,0); \fill[datapointA] (0.175,0) circle (\scurvepoint);} High $I_0$
        };
        \node[anchor=west, font=\fonttwo] at (2.9,0.7) {
            \tikz[baseline=-0.5ex]{\draw[curveC] (0,0) -- (0.35,0); \fill[datapointC] (0.175,0) circle (\scurvepoint);} Low $I_0$
        };
        \node[font=\fontone\bfseries] at (1.75,2.8) {S-Curve Family};
    \end{tikzpicture}%
    }
     
    \resizebox{\textwidth}{!}{%
    \begin{tikzpicture}[
        >=Stealth,
        flowbox/.style={
            draw, thick, rounded corners=3pt,
            minimum height=1.5cm,
            align=center,
            inner sep=5pt,
            fill=black!6
        },
        flowarrow/.style={->, line width=1.1pt, black!50, >=Stealth}
    ]
        \pgfmathsetmacro{\panelscale}{0.6}
        \pgfmathsetmacro{\lgap}{1}      
        \pgfmathsetmacro{\rgap}{1}      
        \pgfmathsetmacro{\rowsep}{0.25}   
        
        \pgfmathsetmacro{\cambodyW}{0.5}
        \pgfmathsetmacro{\cambodyH}{0.45}
        \pgfmathsetmacro{\cambodyHtop}{1.575}
        \pgfmathsetmacro{\camlensHtop}{\cambodyHtop+0.075}
        
        \pgfmathsetmacro{\camlensW}{0.25}
        \pgfmathsetmacro{\camlensH}{0.35}
        
        \pgfmathsetmacro{\cambodyHbot}{-2.3}
        \pgfmathsetmacro{\camlensHbot}{\cambodyHbot+0.075}
        \pgfmathsetmacro{\camX}{-0.15}
     
        \node[inner sep=0pt, anchor=south east] (staticscenes) 
            at (-\lgap+0.2, \rowsep+0.4) 
            {\scalebox{\panelscale}{\usebox{\StaticScenes}}};
     
        \node[inner sep=0pt, anchor=south west] (noisedist) 
            at (\rgap-0.08, \rowsep) 
            {\scalebox{\panelscale}{\usebox{\NoiseDistribution}}};
     
        \node[inner sep=0pt, anchor=north east] (sqwave) 
            at (-\lgap-0.15, -\rowsep) 
            {\scalebox{0.65}{\usebox{\SquareWaveStimuli}}};
     
        \node[inner sep=0pt, anchor=north west] (scurves) 
            at (\rgap, -\rowsep-0.5) 
            {\scalebox{\panelscale}{\usebox{\SCurveFamily}}};
     

        \draw[flowarrow] (\cambodyW +\camX, \cambodyHtop+.25) -- (\rgap - 0.1, \cambodyHtop+.25);
        \draw[flowarrow] (\cambodyW +\camX, \cambodyHbot+.25) -- (\rgap - 0.1, \cambodyHbot+.25);

        \fill[black!85] (\camX, \cambodyHtop) rectangle (\cambodyW - 0.05, \cambodyHtop+0.5);
        \fill[black!85] (\camX - \camlensW, \camlensHtop) rectangle (-0.05, \camlensHtop+.35);

        \fill[black!85] (\camX, \cambodyHbot) rectangle (\cambodyW - 0.05, \cambodyHbot+.5);
        \fill[black!85] (\camX - \camlensW, \camlensHbot) rectangle (-0.05, \camlensHbot+0.35);
     
        \draw[thick, black!50, dashed, ] (\camX - \camlensW - 0.1, \camlensHtop+.35) 
            -- (-\lgap , \camlensHtop+.8);
        \draw[thick, black!70, dashed, ] (\camX - \camlensW - 0.1, \camlensHtop) 
            -- (-\lgap , \camlensHtop-.5);
     
        \draw[thick, black!70, dashed] (\camX - \camlensW - 0.1, \camlensHbot) 
            -- (-\lgap , \camlensHbot-0.5);
        \draw[thick, black!50, dashed] (\camX - \camlensW - 0.1, \camlensHbot+0.35) 
            -- (-\lgap , \camlensHbot+0.8);
     
        \node[font=\small\bfseries, anchor=north] at (0.275, 0) {Event camera};
     

        \newcommand{\fontfour}{\fontsize{12pt}{14pt}\selectfont}
        \newcommand{\fontfive}{\fontsize{10pt}{12pt}\selectfont}
    
        \node[flowbox,
              anchor=west,                
              label={[yshift=0.1 cm,font=\bfseries]above:{\fontfour Noise2Params}}
             ] (noise2params)
             at ([xshift=0.4cm, yshift=0.2cm]noisedist.east) 
             {\fontfive Fit analytic \\ \fontfive model to data};   
     
        \draw[flowarrow] ([xshift=-0.5cm, yshift=0.2cm]noisedist.east) -- (noise2params.west);
     
        \node[flowbox, anchor=west,
            label={[yshift=0.1 cm,font=\bfseries]above:{\fontfour Result}}
            ] (paramvalues) 
            at ([xshift=0.6cm, yshift=-0.17 cm]noise2params.east) {\fontfive Camera-\\ \fontfive specific\\ \fontfive parameter\\ \fontfive values};
     
        \draw[flowarrow] (noise2params.east) -- ([yshift=0.17 cm]paramvalues.west);
     
        \draw[flowarrow, rounded corners=5pt] 
            ([xshift=0.5cm]scurves.east) -| (noise2params.south);
     
    \end{tikzpicture}%
    }
    \caption{Schematic flow chart of the parameter determination pipeline. The top row shows the primary path posited by this work where observations of uniform static scenes, at different intensities, allow for the construction of empirical noise event distributions, which are fitted against by the Noise2Params analytic model, resulting in the determination of camera-specific parameter values. This employs the our direct measurements of static scene noise events. The bottom row shows the additional inclusion of S-curve distributions, resulting from observations of modulated square-wave intensity stimuli, over different baseline intensities. We did not make direct observations of S-curve distributions, but employed available S-curve data to act as an additional bounding constraint on the parameter ranges. The plots shown here are illustrative and do not correspond to actual data points.}
    \label{fig:param_determ_flowchart}
\end{figure*}
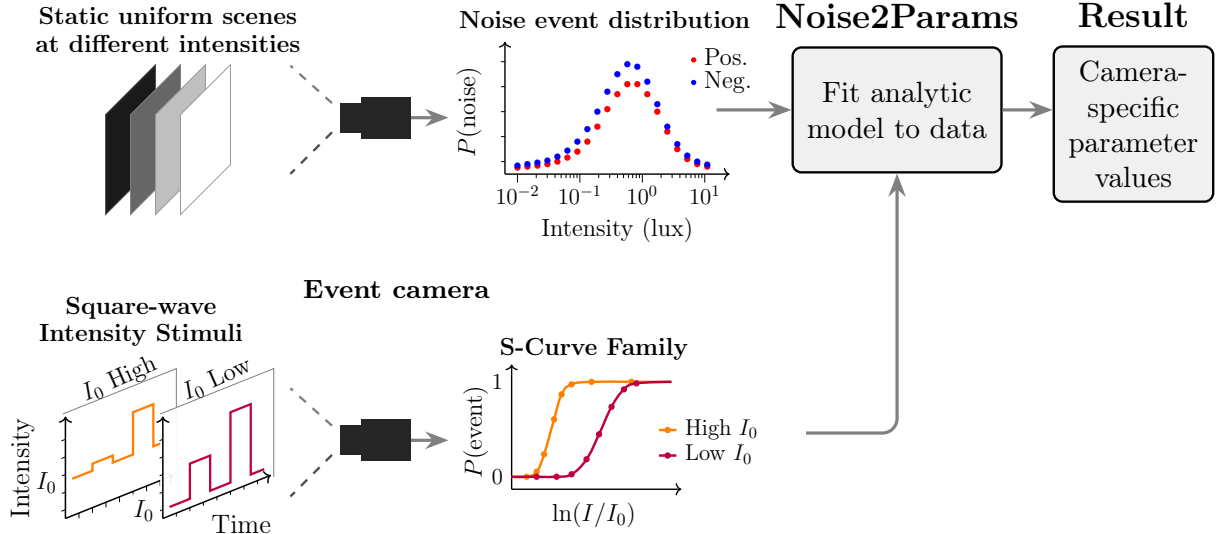

With the measured noise-event probability curves and reported S-curves in hand, we estimate the physical parameters $B$, $\alpha$, and $\theta$. We call this overall method Noise2Params, which is shown schematically in Fig.~\ref{fig:param_determ_flowchart}. 
\par
We model the leakage term as intensity dependent, $\theta=\theta(\lambda)$, where $\lambda$ is the mean photon count and $\lambda=\alpha I$ (with $I$ in lux). To probe the structure of $\theta(\lambda)$, we fixed $(B,\alpha)$ over a grid and, for each intensity level, numerically solved for the value of $\theta$ that matched the observed noise-event probabilities. We performed this inversion using all three probabilistic models, but relied primarily on the exact Poisson and saddle-point forms because they remain valid in the low-intensity regime.
 
The resulting $\theta$ estimates were well described by the parametric form
\begin{equation} \label{eq: analytic theta}
    \theta(\lambda)=c_1+c_2\sqrt{\lambda}+c_3\lambda.
\end{equation}
We discuss the physical interpretation of this structure in Section~\ref{section: discussion}.
 
With Eq.~\eqref{eq: analytic theta} fixed, we jointly fit $B$, $\alpha$, and $(c_1,c_2,c_3)$ by minimizing the RMSE against (i) the measured noise-event probability curves and (ii) the reported S-curves, using a single objective that weights both data sources equally. In some bias setting configurations, the measured noise probabilities did not approach zero at the lowest or highest intensities, indicating a non-negligible noise floor. To accommodate this behavior, we introduce an additive offset $c_V$ applied to the modeled probabilities. We set $c_V$ to the minimum observed probability in the corresponding dataset and retain it only when it improves the overall fit; otherwise we set $c_V=0$. The dependence of this floor value on bias settings is discussed in Section~\ref{s-section: non-default settings}.

At default bias settings we constrain $B$ to be the same for positive and negative events. Residual polarity asymmetries are captured by allowing $(c_1,c_2,c_3)$ (and, when needed, $c_V$) to differ between polarities; the fitted $\alpha$ is shared. The best-fit values are reported in Table~\ref{table:parameter_values}. 
The best-fit value of $\alpha$ is further supported by agreement with a radiometric order-of-magnitude cross-check (Section~\ref{s-section: alpha crosscheck}). 
\par
We omit reporting formal uncertainties for the best-fit values in Table~\ref{table:parameter_values}, for several interrelated reasons. First, the event probability distributions are sharply sensitive to small parameter changes, with $B$ in particular exerting an exponential influence, so that conventionally propagated intervals would correspond to predicted distributions varying by orders of magnitude and would convey little practical information. Second, the fit combines sets of observations with asymmetric uncertainty information: our measured noise-event probabilities carry both temporal and spatial variances, whereas the externally reported S-curves are available without experimental uncertainties. The appropriate prescription for combining these into a single uncertainty budget is not obvious. Third, the empirical distributions are themselves ensemble statistics over the sensor pixel array, which exhibits parameter heterogeneity, so that the observed variance convolves per-pixel fitting uncertainty with the pixel-to-pixel variation of the parameters. Without an independent characterization of the per-pixel parameter distributions, disentangling these contributions to yield a well-defined uncertainty on an array-averaged parameter value is not tractable within our framework. While we adopt simple parametric descriptions of the heterogeneity in $B$ and $\theta(\lambda)$ later in this work, which are themselves acknowledged there to be incomplete, per-pixel variation likely extends to the remaining fitted parameters as well as to quantities outside the fit, such as the refractory time $R$ and the timing uncertainty of individual events. We therefore report point estimates.
\par
\begin{figure*}[!ht]
    \centering
    \includegraphics[width=0.6\linewidth]{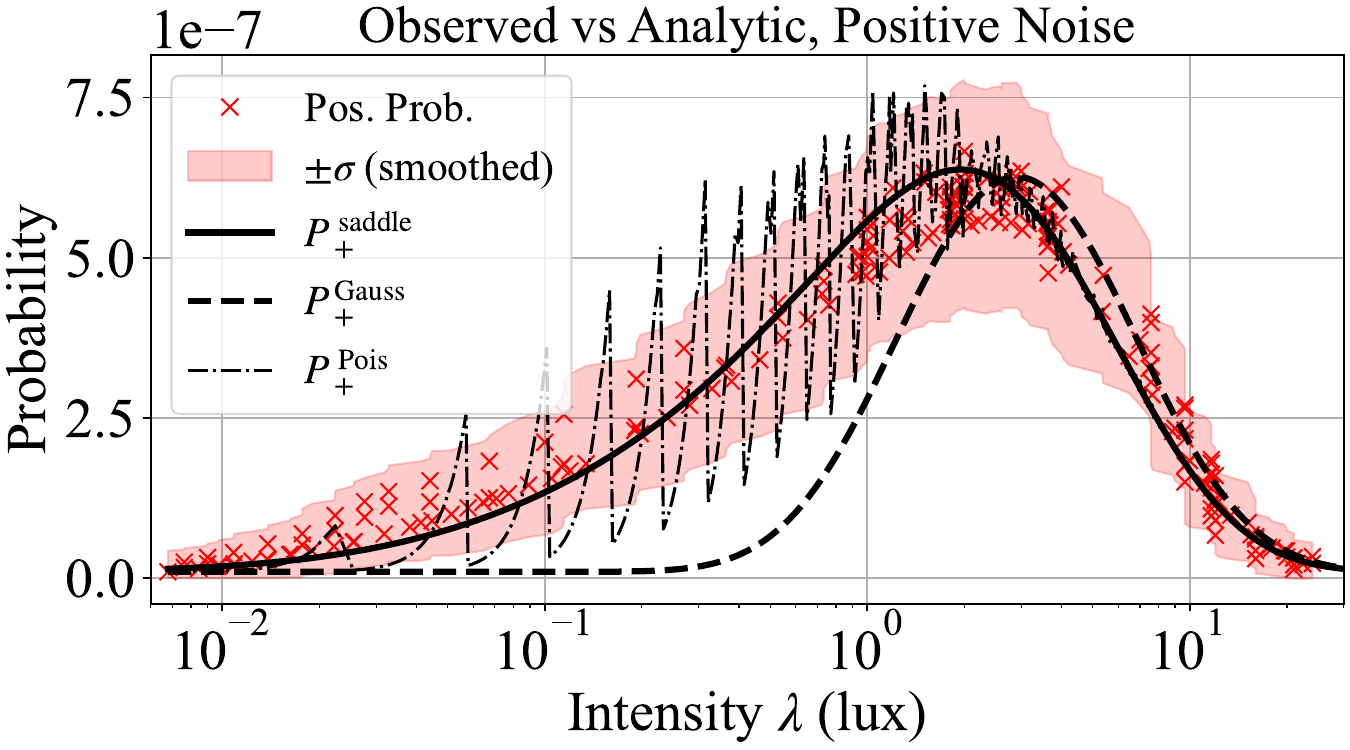}
    \caption{Saddle-point, Gaussian, and Poisson distributions for positive noise events, computed with the fitted parameters of Table~\ref{table:parameter_values}, shown alongside the empirical probabilities and temporal uncertainty band. The saddle-point distribution (solid line) agrees excellently with the data (RMSE: $4.52 \times10^{-8}$, $\chi^2_\nu: 0.14$, $R^2: 0.96$). The Gaussian distribution (dashed line) captures only the mid to high intensity regime. The Poisson distribution (dash-dotted line) exhibits discontinuities at certain intensities; see text for discussion.}
    \label{fig:prob saddle dist pos fitted params}
\end{figure*}
\begin{figure*}[!ht]
    \centering
    \includegraphics[width=0.6\linewidth]{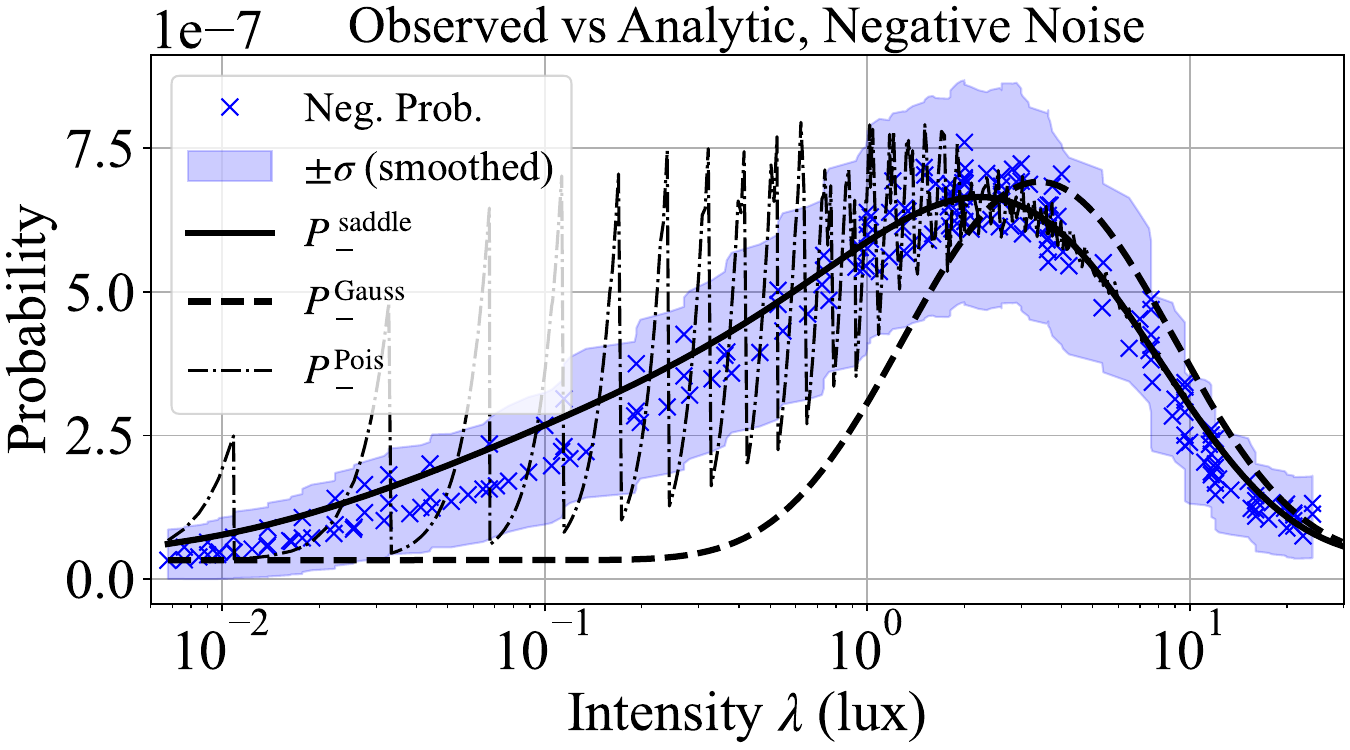}
    \caption{Same as Fig.~\ref{fig:prob saddle dist pos fitted params} but for negative noise events. The saddle-point distribution again shows excellent agreement with the data (RMSE: $4.34 \times10^{-8}$, $\chi^2_\nu: 0.15$, $R^2: 0.97$).}
    \label{fig:prob saddle dist neg fitted params}
\end{figure*}
\begin{figure*} 
    \centering
    \includegraphics[width=1\linewidth]{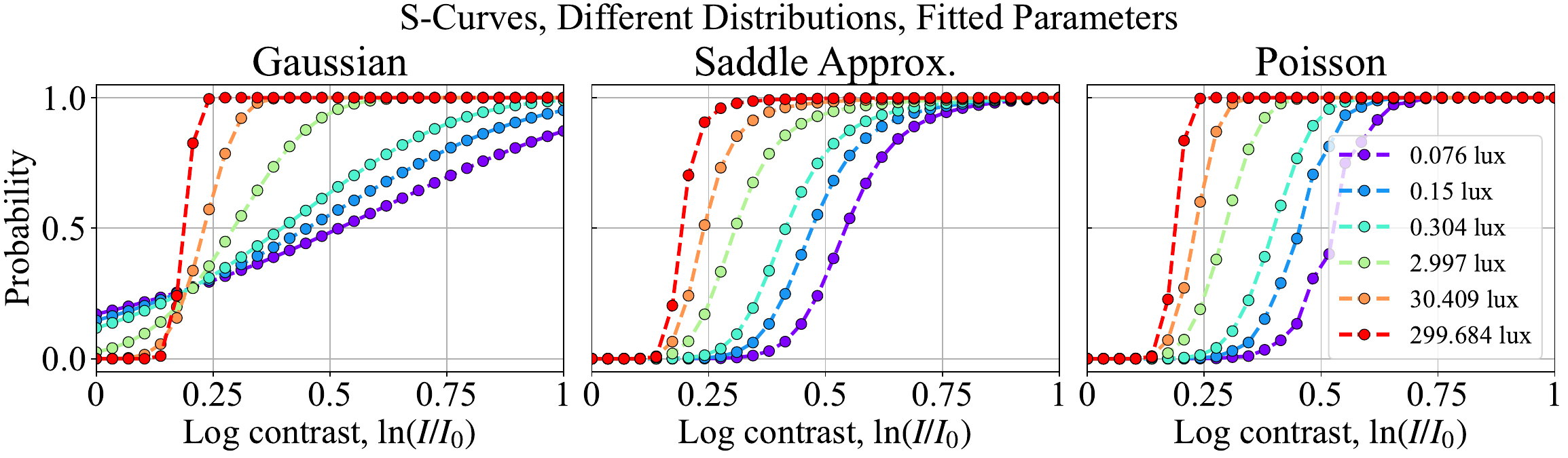}
    \caption{Heaviside event distributions, or the S-curve family, for each of the distribution methods, computed with fitted parameters and 1000 sampled pixels. Again, the three distributions agree well at high intensities, but only the Poisson and the saddle are consistent at low intensities. Compare with the reported S-curve observations in Fig.~\ref{fig:reported s-curves}.}
    \label{fig:fitted S-curves}
\end{figure*}

The fitted noise-event probability curves are shown in Figs.~\ref{fig:prob saddle dist pos fitted params} and~\ref{fig:prob saddle dist neg fitted params}, and the corresponding fitted S-curves are shown in Fig.~\ref{fig:fitted S-curves}, where the S-curves are computed by averaging over the pixel-to-pixel threshold sampling of $B$ described below. Across the full intensity range, the Poisson and saddle-point models match the observations closely, while the Gaussian approximation agrees only at mid to high intensities. Aside from the explicit $\lambda$-dependence in $\theta(\lambda)$, all parameters are assumed to be intensity independent.
\par
 
\begin{table} 
    \centering
    \caption{Best-fit values of the parameters for default bias settings, for positive and negative events.}
    \label{table:parameter_values}
    \renewcommand{\arraystretch}{1.2}
    \begin{tabular}{lcc}
        \hline
        Parameter & Positive & Negative \\
        \hline
        $B$       & 0.15                 & 0.15              \\
        \hline
        $\alpha$  & 4.5                  & 4.5               \\
        \hline
        $c_1$     & 18.92                & 16.42             \\
        \hline
        $c_2$     & 35.49                & 37.42             \\
        \hline
        $c_3$     & 0.439                & 0.0676            \\
        \hline
        $c_V$     & $9.57\cdot10^{-9}$   & $3.18\cdot10^{-8}$\\
        \hline
    \end{tabular}
    
\end{table}

    
 
\begin{figure*} 
    \centering
    \includegraphics[width=1\linewidth]{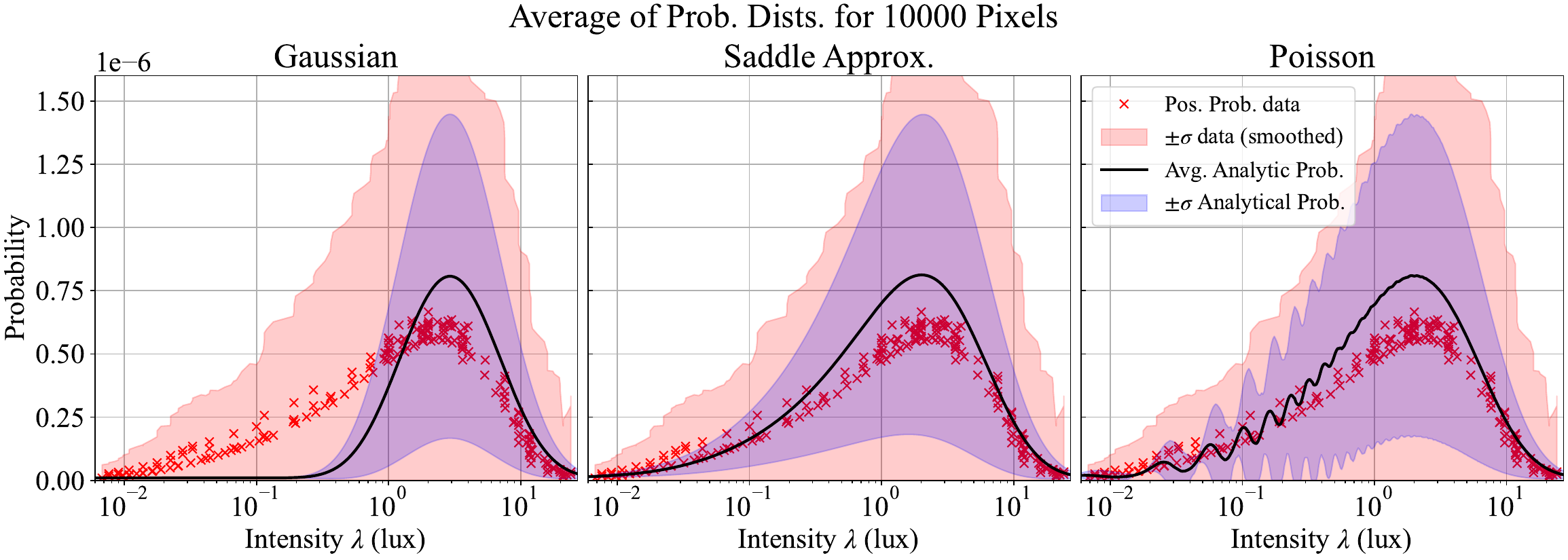}
    \caption{ Comparison of $\hat{P}_+$, with spatial uncertainty band, with averages of $P_+$ for 10,000 simulated pixels, via each of the probability distributions, computed using fitted parameters. The per-pixel threshold $B$ is sampled from $\mathcal{N}_{\left[0,\infty\right)}(\mu_B=0.15,\;\sigma_B=0.0045)$; all other parameters are held fixed. The averaging substantially smooths the discontinuities visible in the single-pixel Poisson distribution (cf.\ Fig.~\ref{fig:prob saddle dist pos fitted params}).}
    \label{fig: Poisson avg}
\end{figure*}

The exact Poisson noise-event distribution curves exhibit small discontinuities resulting from the summation bounds involving floor and ceiling operators (Eqs.~\eqref{eq: Poisson dist pos} and~\eqref{eq: Poisson dist neg}), reflecting the discrete nature of photon counts. Note that the vertical segments visible in Figs.~\ref{fig:prob saddle dist pos fitted params} and~\ref{fig:prob saddle dist neg fitted params} are artifacts of plotting these discontinuities and do not represent actual parts of the distribution. The calculations above also assume a single fixed threshold $B$ for all pixels, whereas real sensors exhibit pixel-to-pixel threshold variation.  To assess the impact of this heterogeneity, we average the modeled probabilities over an ensemble of pixels with thresholds
\begin{equation}
    B_i \sim \mathcal{N}_{\left[0,\infty\right)}(\mu_B = 0.15,\sigma_B=0.0045),
\end{equation}
that is, a normal distribution with mean $\mu_B$ and standard deviation $\sigma_B$, truncated to $\left[0,\infty\right)$ to enforce $B_i>0$. This value of $\sigma_B$ is taken from the manufacturer-stated typical pixel-to-pixel variation, $\sigma_B = 0.0045 = 3\% \text{ of } B = 0.15$\citep{sony2025imx636aamr}. The resulting averaged curves (Fig.~\ref{fig: Poisson avg}) substantially smooth the Poisson discontinuities.
\par
As can be seen in Fig.~\ref{fig: Poisson avg}, the averaged-model variability remains smaller than the observed variability in $\hat{P}_\pm$, and the averaged curves exhibit a slightly higher peak. This suggests that the true pixel-to-pixel parameter variation is both larger than the nominal $\sigma_B$ and may follow a distribution that is non-Gaussian or imbalanced (for example, skewed). Moreover, the averaged Poisson model shows increased variability at low intensities relative to the saddle-point approximation, qualitatively closer to the behavior of $\hat{P}_\pm$, which supports the need for the exact (discrete) treatment when modeling low-intensity behavior.

\section{Synthetic Data Generation and Deep-Learning Reconstruction}
\label{sec: noise_event_images}
 
\subsection{Construction of Noise Images}
Given that noise events are intensity dependent, by per-pixel accumulation of an event feed from a static scene, for a given integration time, one can construct a ``noise image,'' where brighter pixels of the image correspond to more events detected at that pixel of the sensor array. In the case of non-uniform scenes, details of the scene clearly emerge in the noise image, as was demonstrated by Cao \textit{et al.}~\citep{cao2025noise2image}. To generate experimental noise images, EC recordings were taken of greyscale (non-uniform) images displayed on the same LED screen as in Section~\ref{subsection: setup} with screen brightness maximized (all experimental noise images were generated at this brightness). Given a particular integration time, events were accumulated over a random clip, with the length of the clip being the integration time, from the recording. 
\par
To enable comparison with noise images constructed from experimental data, we generated synthetic noise images using the noise-event probability models. First, we determined a mapping from the greyscale pixel values of the displayed images (integers in the range 0--255, with 255 corresponding to white) to the intensity reported by the EC's illuminometer (see Section~\ref{s-section:greyscale_to_intensity_mapping}). This mapping was then used to convert each greyscale image into an array of synthetic intensity values $\{I_i\}$. Before this conversion, the greyscale images were resized to match the EC resolution of $1280 \times 720$. The intensity values were mapped to average photon counts $\{\lambda_i\}$ via $\lambda=\alpha I$, as before ($\alpha$ was considered a constant across the array).
\par
The per-pixel $B_i$ are sampled as was done earlier, except different values of $\sigma_B$ were used depending on the probability model, as detailed below. To clarify, $B_i$ is used to determine both $\pm B$, that is $+B=B_i$ and $-B=-B_i$. Additional pixel-to-pixel variance was introduced in the leakage term by setting
\begin{equation}
    \theta_{i\pm}(\lambda_i) = X_i\cdot\theta_\pm(\lambda_i), \qquad
X_i \sim \mathcal{N}(\mu_X = 1,\sigma_X^2),
\end{equation}
so that each pixel's leakage is a random multiplicative perturbation of the baseline $\theta_\pm(\lambda_i)$.
\par
For the Gaussian and saddle-point distributions, we used
\[
\sigma_B = 0.0065 \approx 4.\overline{3}\% \text{ of } B = 0.15,
\qquad
\sigma_X = 0.001,
\]
chosen to align both the variance and the maximum value of event counts between the synthetic and experimental noise images (this corresponds to a somewhat larger $\sigma_B$ than the manufacturer-reported value stated earlier). For the Poisson distribution, slightly smaller values,
\[
\sigma_B = 0.006, \qquad \sigma_X = 0.0005,
\]
were used to obtain comparable agreement, reflecting the intrinsically larger variance of the Poisson model. However, as noted in Section~\ref{section: param determ}, this simplistic heterogeneity model results in slightly higher mean event counts than the observed values. For a given choice of probability model, the per-pixel synthetic event probabilities for the respective polarities were then computed as
\begin{equation}
  P_{i\pm} = P(\lambda_i, B_i, \theta_i(\lambda_i); \alpha),  
\end{equation}
where $P(\cdot)$ denotes the corresponding noise-event probability function and $\alpha$ is taken to be a uniform value.
\par
To incorporate pixel dead time, the effective total event probability at pixel $i$ was modeled as
\[
P^{\text{eff}}_{i} = \frac{\Sigma P_i}{1 + \Sigma P_i\,R},
\qquad
\Sigma P_i \equiv P_{i+} + P_{i-},
\]
with $R$ the pixel dead time (see Section~\ref{subsection: setup}). The effective per-polarity probabilities were then taken to be
\begin{equation} \label{eq: p_eff_i_pm}
    P^{\text{eff}}_{i\pm}
    = \left(\frac{\Sigma P_i}{1 + \Sigma P_i\,R}\right)
      \left(\frac{P_{i\pm}}{\Sigma P_i}\right)
    = \frac{P_{i\pm}}{1 + \Sigma P_i\,R}.
\end{equation}
Finally, the synthetic per-pixel event counts for each polarity were generated by sampling from a binomial distribution with success probability \(P^{\text{eff}}_{i\pm}\) and number of trials equal to the number of microseconds in the integration time.
\par
\begin{figure*} 
    \centering
    \includegraphics[width=1\linewidth]{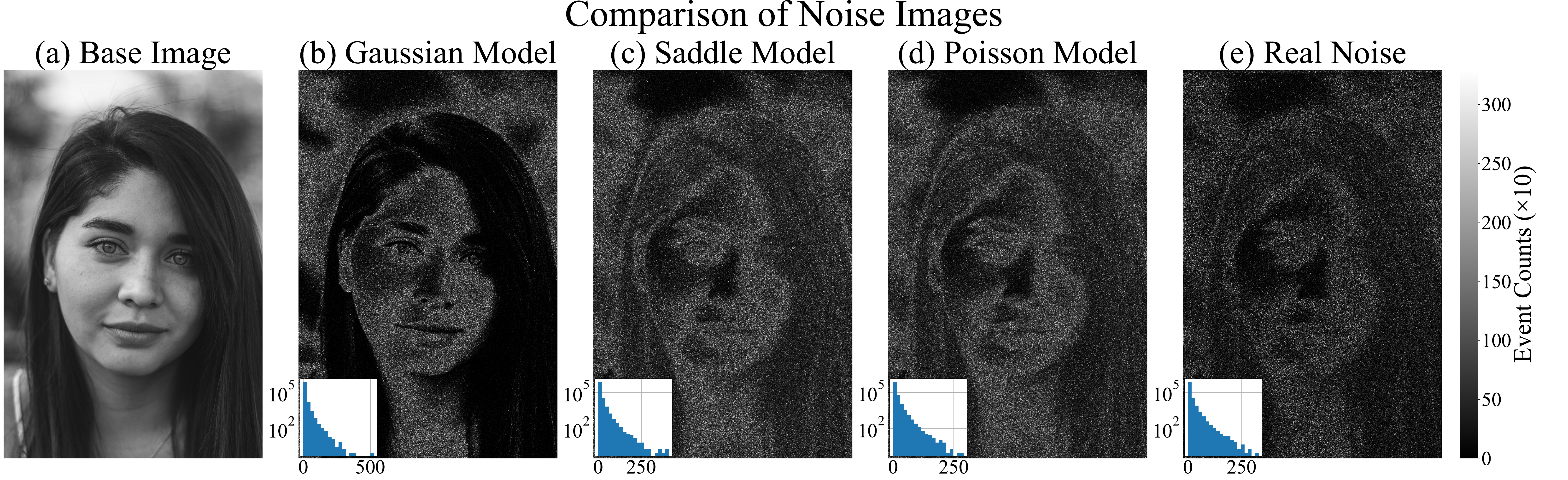}
    \caption{(a) A base greyscale image, (b--d) corresponding synthetic noise images from the different probability models, and (e) corresponding experimental noise image, all generated with an integration time of 5 s. The displayed noise images have been scaled up by a factor of 10 for greater visual clarity. The histogram insets show the per-pixel event count frequencies (unscaled), with a logarithmic $y$-axis.}
    \label{fig:noise_images_comparison_sampled}
\end{figure*}

A representative example of an experimental noise image and the corresponding synthetic noise images are shown in Fig.~\ref{fig:noise_images_comparison_sampled}. The synthetic noise images produced by all probability models agree well with the experimental noise in the bright regions of the base image. In the darker regions, however, only the saddle-point and Poisson models remain in good agreement with the experiment. The Gaussian synthetic noise image systematically underestimates the event counts in these regions, indicating its breakdown in the low-intensity regime.
\par
Although the saddle-point and Poisson synthetic images reproduce the experimental data in terms of overall variance and maximum event counts, the experimental noise images exhibit larger regional fluctuations in areas of higher event counts and slightly lower mean counts in such areas. This behavior suggests that the variance of the event counts may itself be intensity dependent, which is not captured by our model, while the higher mean of the synthetic counts are an artifact of our simplistic per-pixel sampling, as mentioned ealier, and unrelated to the probability models themselves.

\subsection{Recovering Images from Noise via Deep Learning}
With the generation of experimental and synthetic noise images established, we next used a convolutional neural network (CNN) to reconstruct the underlying greyscale images from the noise images, following the general approach of Cao \textit{et al.}~\citep{cao2025noise2image}. Specifically, we employed a U-Net architecture with self-attention, adapted from \citep{cao2025noise2image}, and trained it on datasets of synthetic noise images generated from each of the probability models, as well as on a dataset of experimental noise images. The base images used are adaptations of the datasets employed by Cao\emph{~et~al.}~ \citep{cao2025noise2image}, which are themselves derived from the DIV2K image dataset~ \citep{Agustsson_2017_CVPR_Workshops} and human
portraits from Unsplash. A detailed description of the CNN architecture, training procedure, and dataset information, together with the nuanced but important differences between the approach of Cao \textit{et al.} and ours, is provided in Section~\ref{s-sec:recon_from_noise_images}.
\par
\begin{figure*}[h]
    \captionsetup[subfigure]{labelformat=empty}
    \newcommand{\imagesizeone}{0.19}
        \centering
    \begin{subfigure}{\imagesizeone\textwidth}
        \caption{Base Image}
        \includegraphics[width=\textwidth]{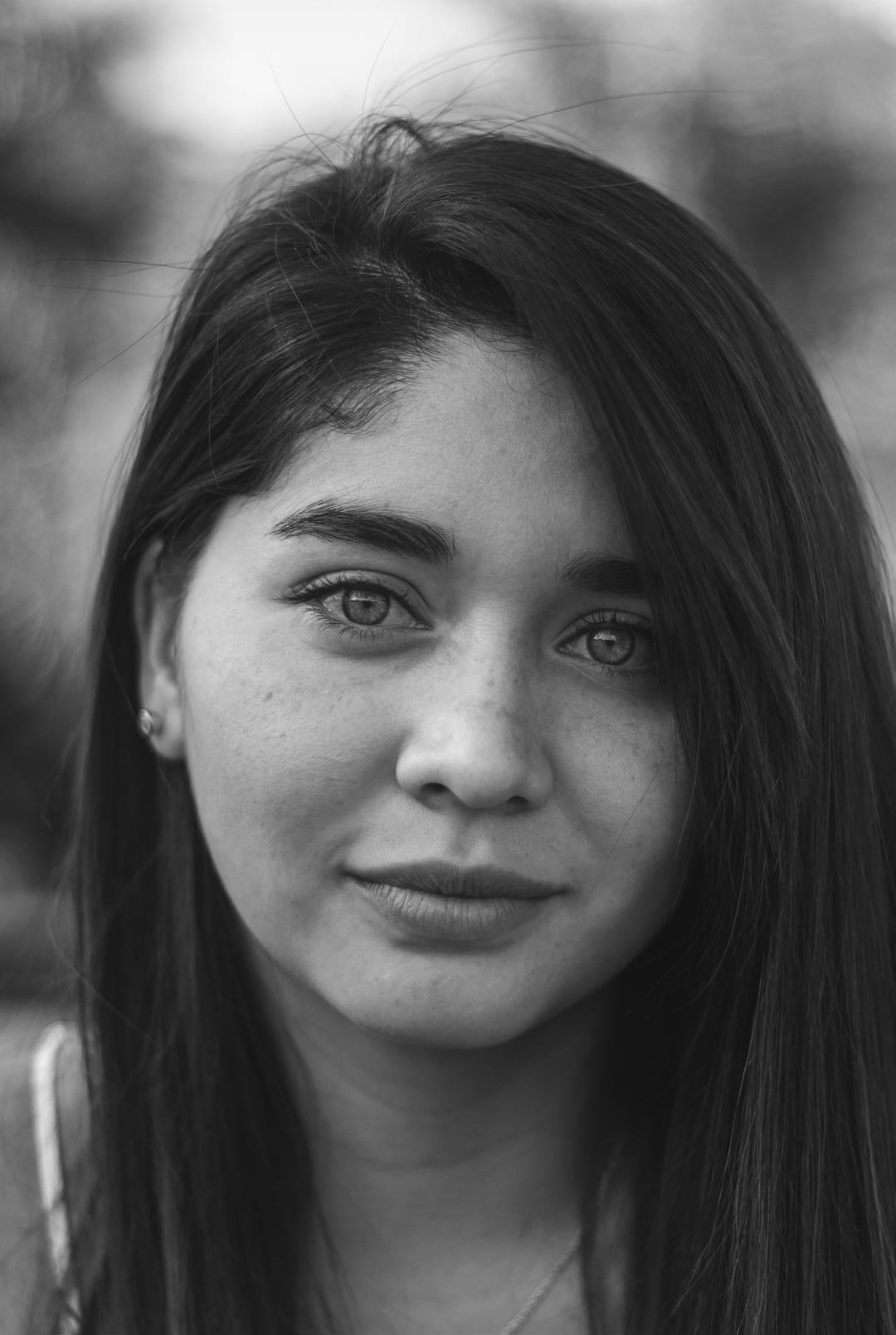}
    \end{subfigure}
    \begin{subfigure}{\imagesizeone\textwidth}
        \caption{Gaussian Model}
        \includegraphics[width=\textwidth]{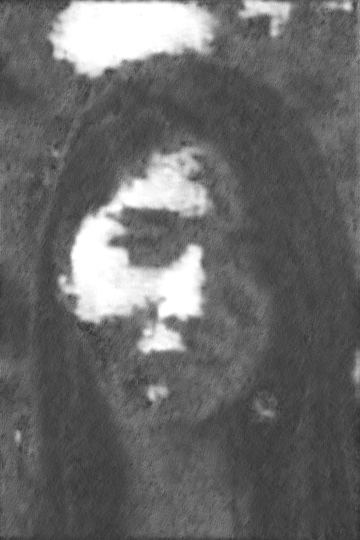}
    \end{subfigure}
    \begin{subfigure}{\imagesizeone\textwidth}
        \caption{Saddle-Point Model}
        \includegraphics[width=\textwidth]{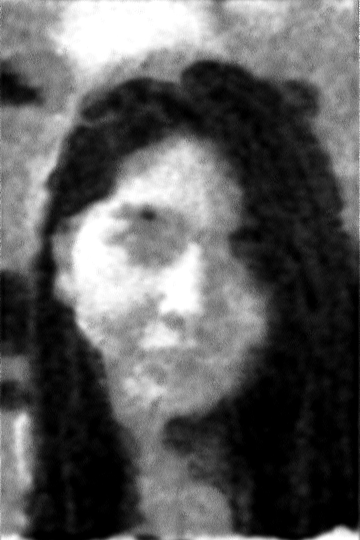}
    \end{subfigure}
    \begin{subfigure}
        {\imagesizeone\textwidth}
        \caption{Mixed Data}
        \includegraphics[width=\textwidth]{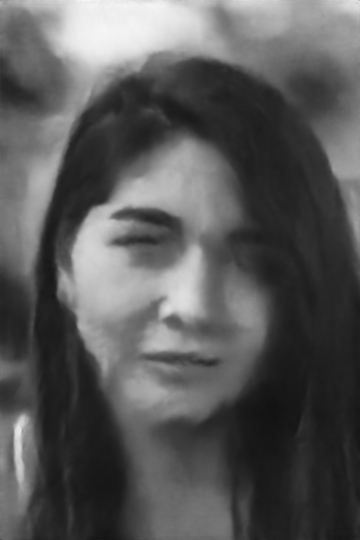}
    \end{subfigure}
    \begin{subfigure}   
        {\imagesizeone\textwidth}
        \caption{Experimental Data}
        \includegraphics[width=\textwidth]{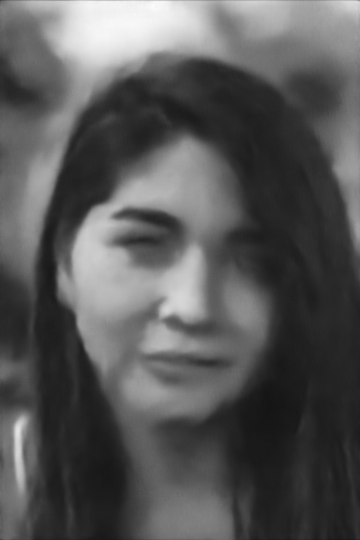}
    \end{subfigure}
 
    \begin{subfigure}{\imagesizeone\textwidth}
        \includegraphics[width=\textwidth]{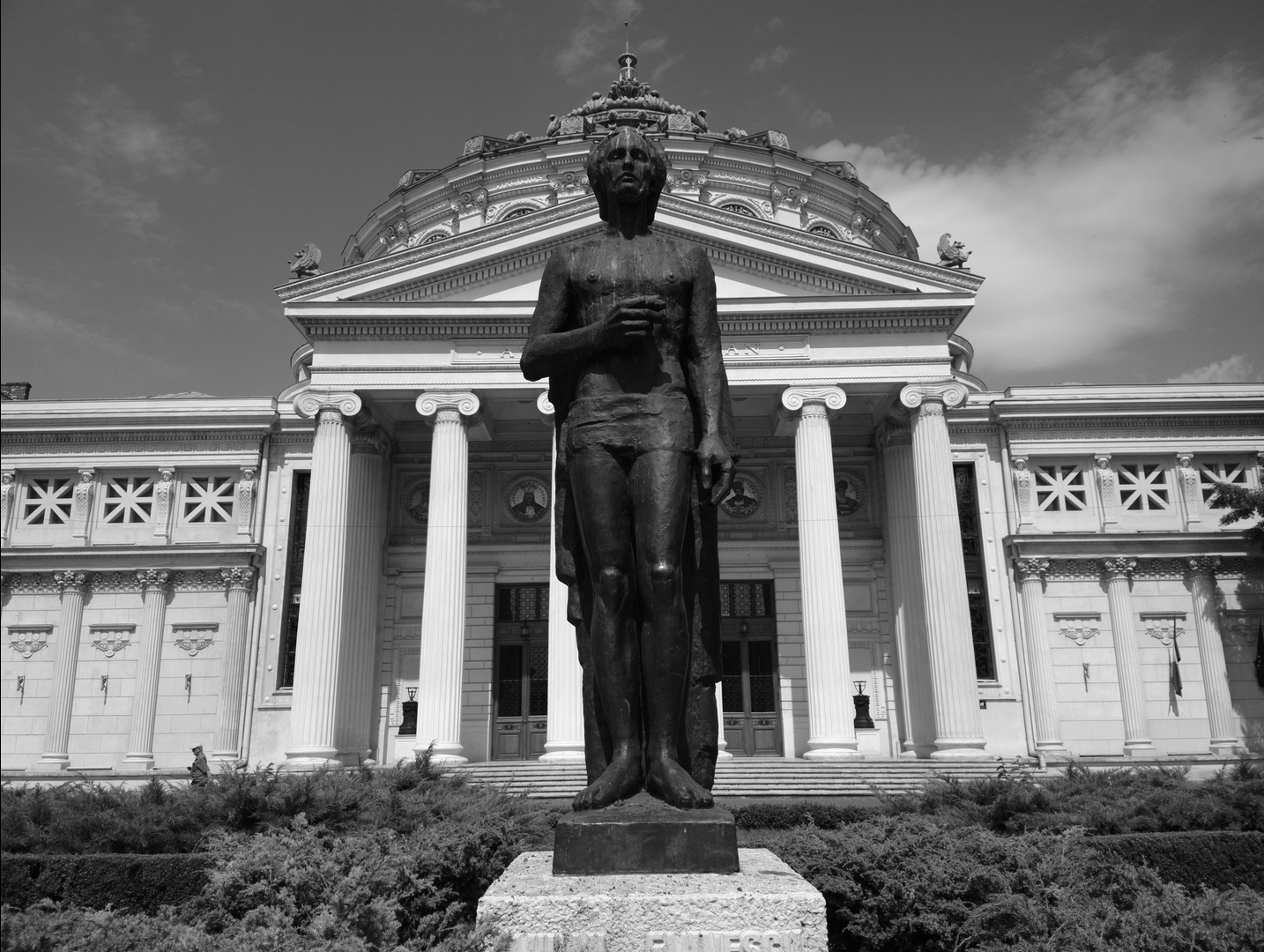}
    \end{subfigure}
    \begin{subfigure}{\imagesizeone\textwidth}
        \includegraphics[width=\textwidth]{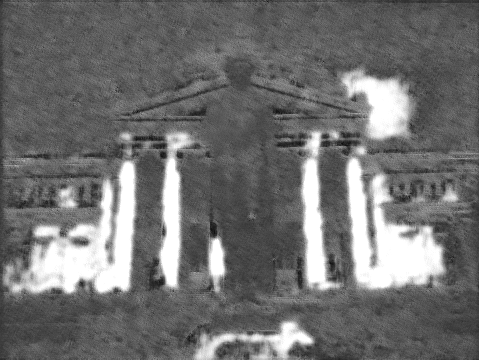}
    \end{subfigure}
    \begin{subfigure}{\imagesizeone\textwidth}
        \includegraphics[width=\textwidth]{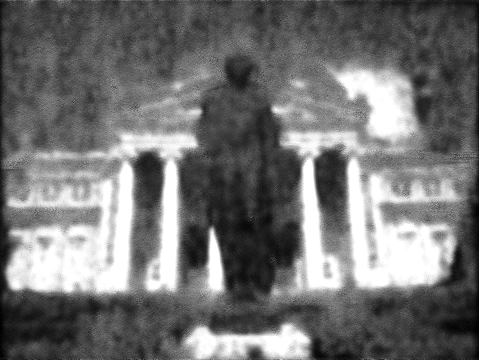}
    \end{subfigure}
    \begin{subfigure}{\imagesizeone\textwidth}
        \includegraphics[width=\textwidth]{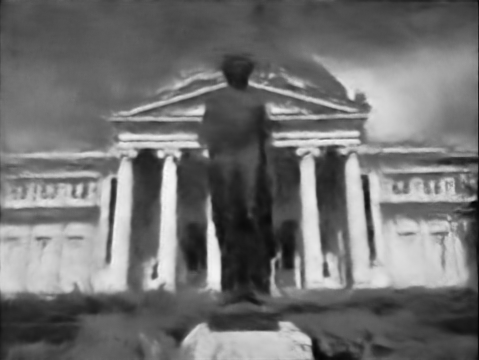}
    \end{subfigure}
    \begin{subfigure}{\imagesizeone\textwidth}
        \includegraphics[width=\textwidth]{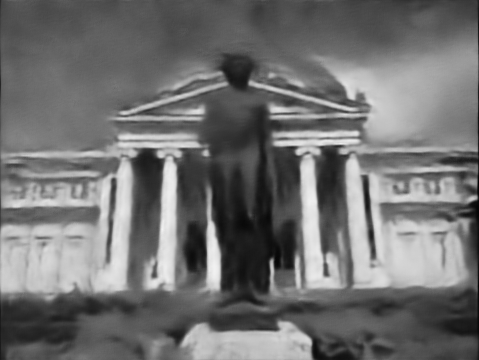}
    \end{subfigure}
 
    \begin{subfigure}   
        {\imagesizeone\textwidth}
        \includegraphics[width=\textwidth]{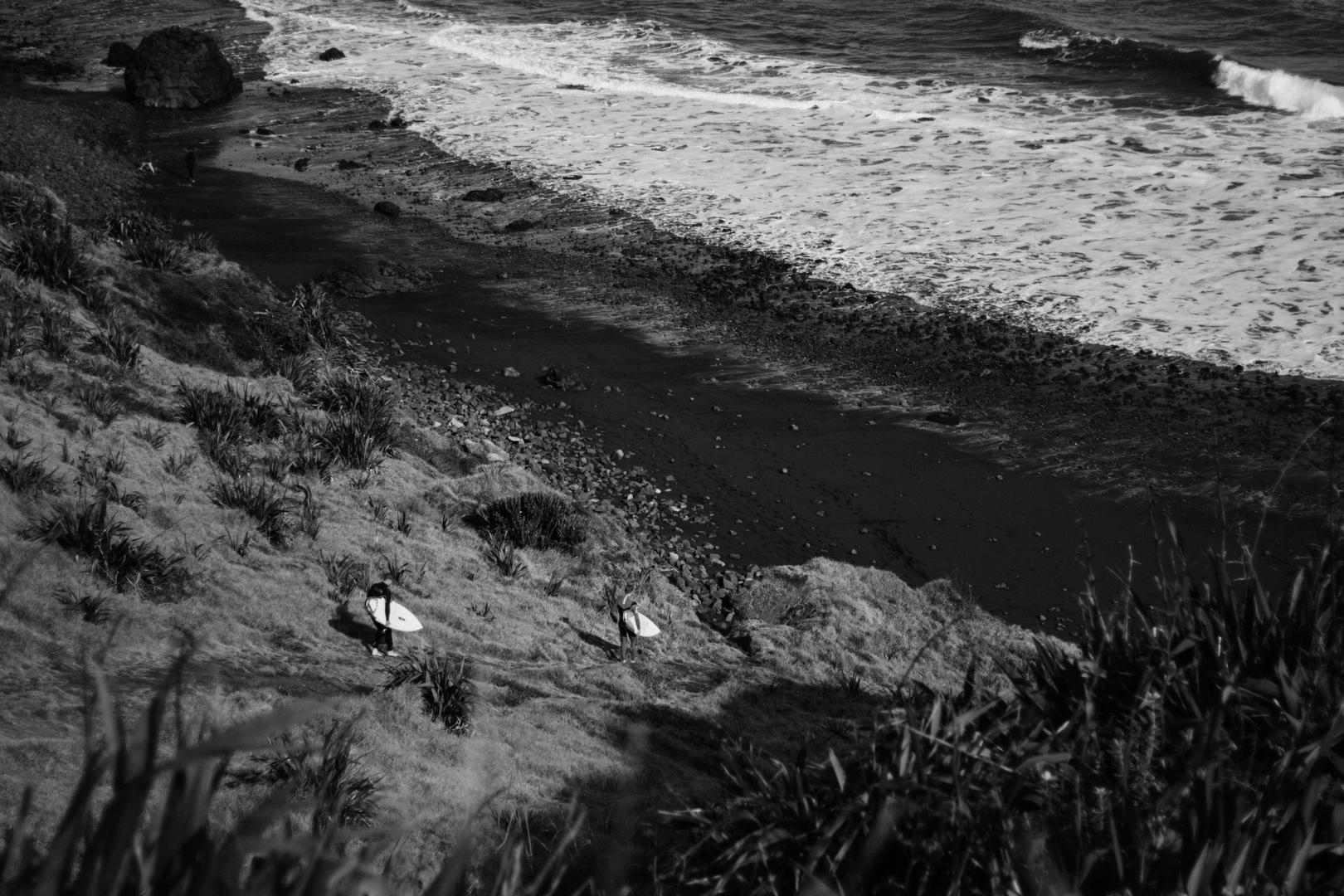}
    \end{subfigure}
    \begin{subfigure}   
        {\imagesizeone\textwidth}
        \includegraphics[width=\textwidth]{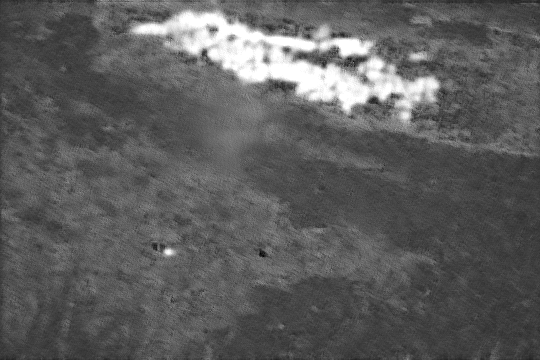}
    \end{subfigure}
    \begin{subfigure}   
        {\imagesizeone\textwidth}
        \includegraphics[width=\textwidth]{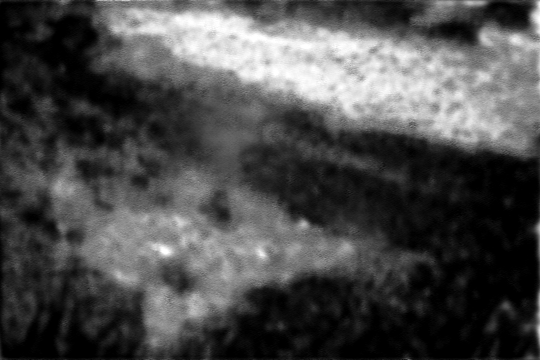}
    \end{subfigure}
    \begin{subfigure}   
        {\imagesizeone\textwidth}
        \includegraphics[width=\textwidth]{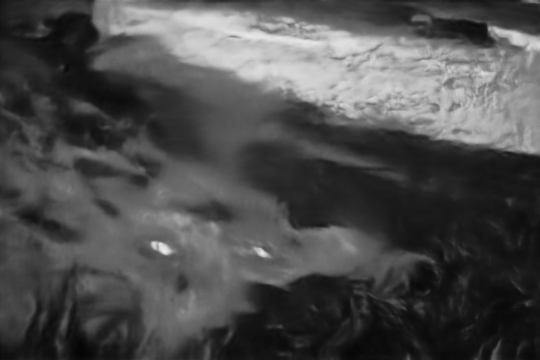}
    \end{subfigure}
    \begin{subfigure}   
        {\imagesizeone\textwidth}
        \includegraphics[width=\textwidth]{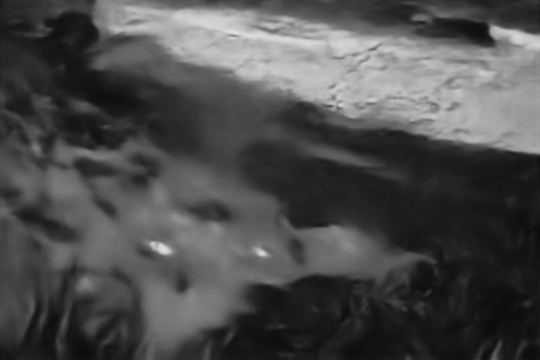}
    \end{subfigure}
 
    \begin{subfigure}   
        {\imagesizeone\textwidth}
        \includegraphics[width=\textwidth]{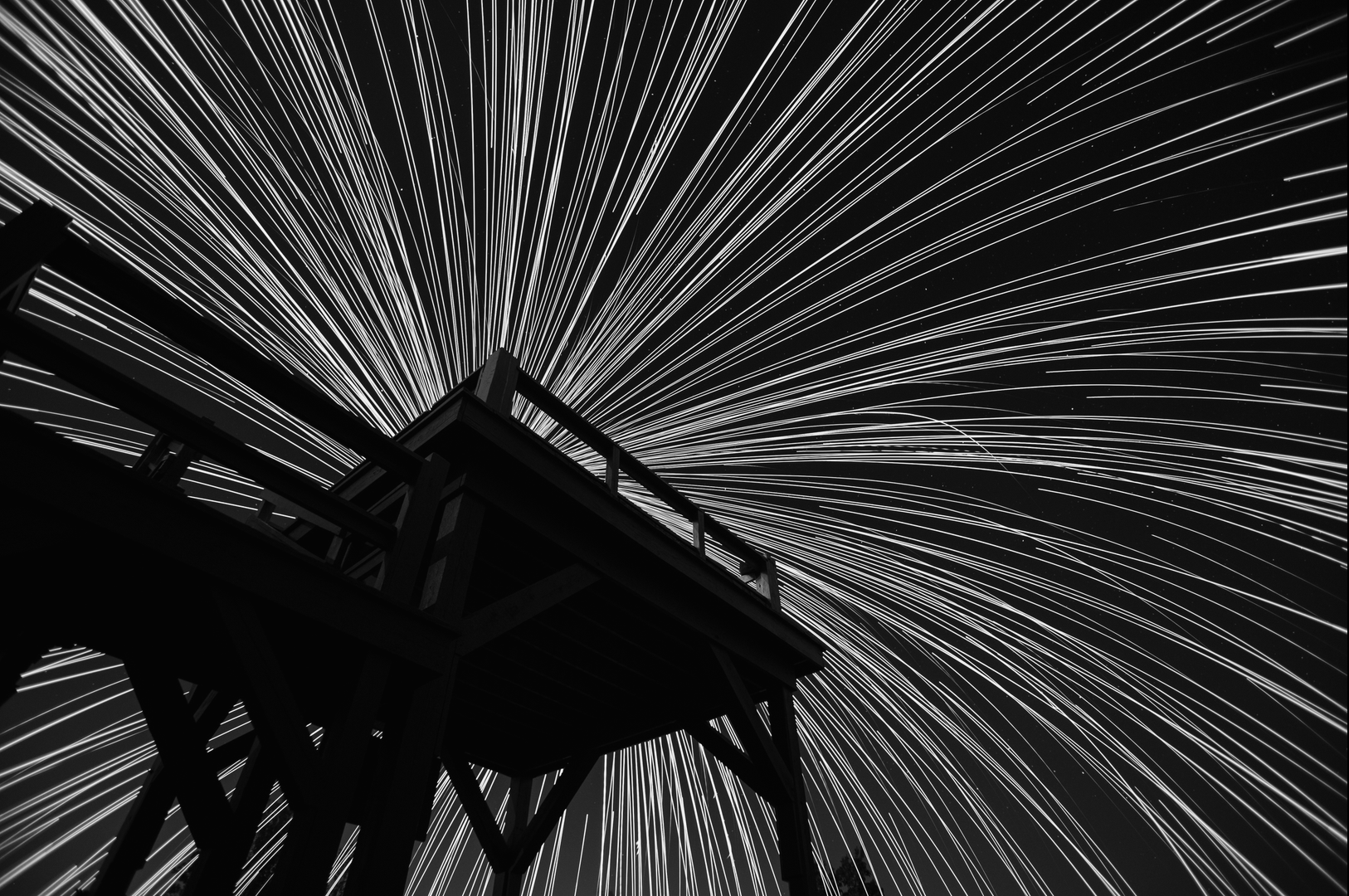}
    \end{subfigure}
     \begin{subfigure}   
        {\imagesizeone\textwidth}
        \includegraphics[width=\textwidth]{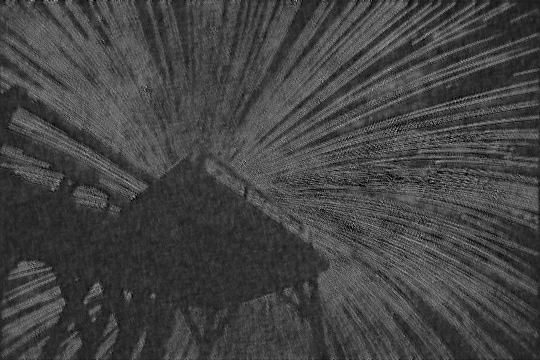}
    \end{subfigure}
     \begin{subfigure}   
        {\imagesizeone\textwidth}
        \includegraphics[width=\textwidth]{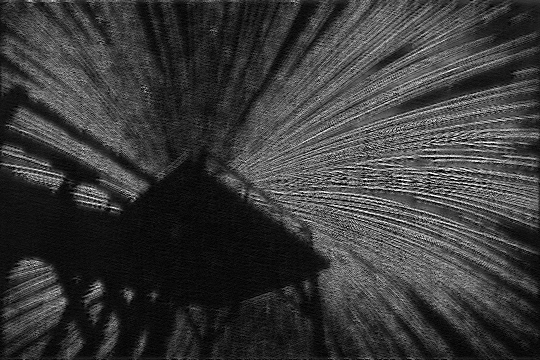}
    \end{subfigure}
    \begin{subfigure}   
        {\imagesizeone\textwidth}
        \includegraphics[width=\textwidth]{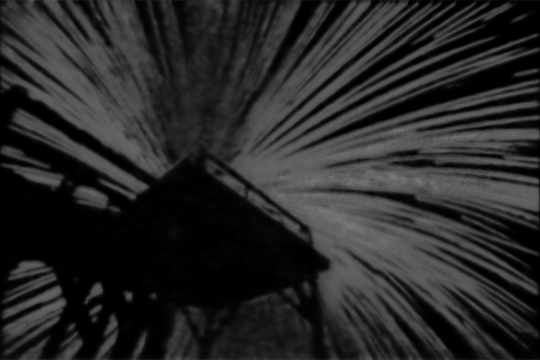}
    \end{subfigure}
    \begin{subfigure}   
        {\imagesizeone\textwidth}
        \includegraphics[width=\textwidth]{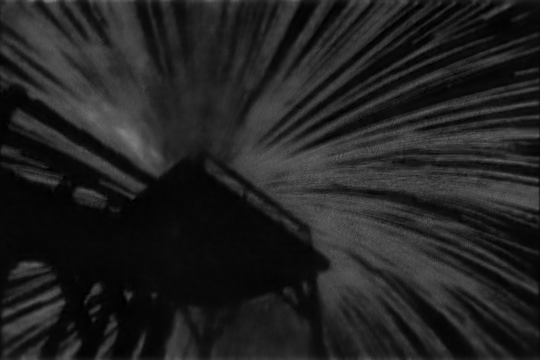}
    \end{subfigure}
    
    \caption{Base greyscale images and reconstructed greyscale images (from noise images generated from experimental recordings of the base images, 5 s integration time), by CNNs trained on Gaussian model synthetic data, saddle-point model synthetic data, mixed data (combination of saddle-point model synthetic data and experimental data), and purely experimental data. The first-row base image was from the training set, while the second through fourth-row base images were part of the validation set.}
    \label{fig:all_reconstrcuction_comparison}
\end{figure*}

The central motivation for this approach is to compare the probability models by evaluating how well a CNN trained on synthetic data generalizes to experimental noise images. If a given probabilistic model is sufficiently accurate, then training the CNN exclusively on synthetic data generated from that model should enable it to learn features that transfer to experimental data, thereby producing high-quality reconstructions of the underlying greyscale images from experimental noise images. Example reconstructions obtained from CNNs trained on synthetic data generated using the Gaussian model, on synthetic data generated using the saddle-point model, and on experimental noise images are shown in Fig.~\ref{fig:all_reconstrcuction_comparison}, as well as from a CNN trained on a mixed dataset (discussed later in this section). 
\par
No synthetic noise image datasets were generated using the Poisson model, because its evaluation is substantially more computationally intensive and more prone to numerical instability in our implementation, while its benefits over the saddle-point model \emph{for this use case} are likely marginal. As a result, the Poisson model is omitted from the deep-learning-based comparisons presented here.

\begin{figure*}
    \centering
    \includegraphics[width=0.9\linewidth]{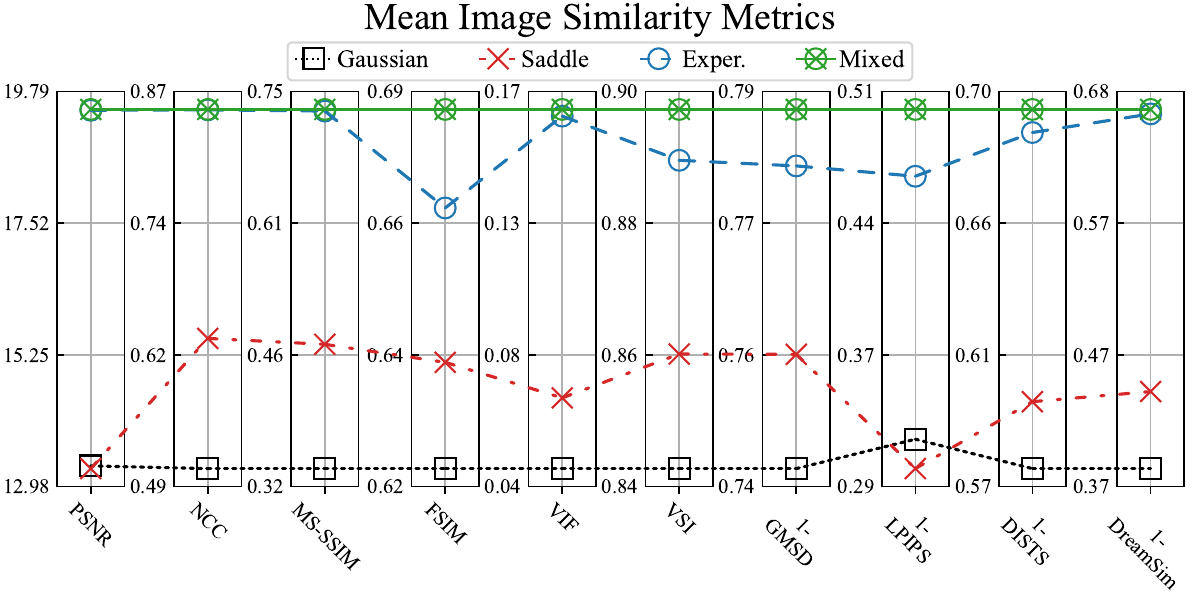}
    \caption{Mean image similarity metric (ISM) scores for each of the CNNs for the test dataset. Seven classical metrics and three deep-learning-based metrics were employed. The classical metrics are: Peak Signal-to-Noise Ratio (PSNR) \citep{PSNR}, Normalized Cross-Correlation (NCC) \citep{Brown1992SurveyRegistration}, Multiscale Structural Similarity Index Measure (MS-SSIM) \citep{ms-ssim}, Feature Similarity Index Measure (FSIM) \citep{fsim}, Visual Information Fidelity (VIF) \citep{vif}, Visual Saliency-Induced Index (VSI) \citep{vsi}, and Gradient Magnitude Similarity Deviation (GMSD) \citep{gmsd}. The deep-learning metrics are Learned Perceptual Image Patch Similarity (LPIPS) \citep{zhang2018unreasonableeffectivenessdeepfeatures}, Deep Image Structure and Texture Similarity (DISTS) \citep{dists}, and DreamSim \citep{fu2023dreamsimlearningnewdimensions}, all of which are trained to mimic human perceptual preferences. For most ISMs, a larger score indicates greater similarity; however, for GMSD and the three deep-learning metrics, a lower score indicates greater similarity. For these four ISMs, we have plotted one minus the mean score, so that a higher vertical placement in the graphic indicates greater similarity for all ISMs.}
    \label{fig:simarlity_metric_scores}
\end{figure*}
 
\par
As can be seen in Fig.~\ref{fig:all_reconstrcuction_comparison}, there is a significant difference in the quality of the reconstructions between the different CNNs based on their training datasets. While all CNNs are able to recover shapes and outlines of objects, correctly reconstructing black levels and details in dark regions of the image is where the CNNs differ. The Gaussian model CNN reconstructs some brighter regions clearly, but loses details in the darker regions of the image, whereas these details are much more clearly preserved by the saddle-point model CNN. It should be noted, though, that both the Gaussian and saddle-point model CNN reconstructions contain artifacts near the edges of some shapes, likely a consequence of incompletely capturing the variance of the experimental data, as noted previously. As could be expected, the experimental data CNN trained on experimental noise images produces the highest-quality reconstructions. 
\par
A noteworthy feature of the reconstructions is that while the reconstructions from the experimentally trained CNN have much greater overall quality than those of the Gaussian or saddle-point CNNs, they tend to have a ``smoothed'' appearance, while in certain aspects, the saddle-point CNN's reconstructions preserve a greater degree of texture or sharpness at the edges of some shapes (this can be seen most clearly in Fig.~\ref{fig:all_reconstrcuction_comparison} in the rays of light in the images in the bottom row and in the water in the images in the third row). This ``smoothing'' in the experimentally trained case is a well-documented effect of pixel-wise MSE and related losses in CNN-based reconstructions, which drive the network toward conditional mean estimates and therefore average out high-frequency, locally structured details such as texture \citep{BlauMichaeli2018PerceptionDistortion, Mustafa_2022_WACV, Nagano2021NoiseReductionSEM}. In the case of the synthetically trained saddle-point model CNN, these details are preserved due to the CNN failing to generalize to the variance of the experimental data, leading to overall lower quality but markedly less averaging out of the high-frequency features.
\par
In light of this, we additionally trained a CNN on a mixed dataset that consisted of synthetic saddle-point model and experimental noise images. Comparisons of reconstructions by the mixed CNN and by the purely experimental CNN can be seen in Fig.~\ref{fig:all_reconstrcuction_comparison}, along with the other reconstructions. A quantitative comparison of the performances of the CNNs is shown by their mean scores on a variety of image similarity metrics (ISMs) in Fig.~\ref{fig:simarlity_metric_scores}, comparing reconstructions with base images of the test dataset (Section~\ref{s-sec:supp_datasets}). ISMs evaluate how similar the predicted images are to the corresponding ground-truth image. Nearly all the ISMs evaluate the saddle-point CNN as performing better than the Gaussian CNN, though both are largely outperformed by the mixed and purely experimental CNNs. Between the latter two, while several ISMs score the mixed and experimental CNNs as equivalent, a majority find that the mixed CNN performs better.
\par
Given the better performance, in terms of static image reconstruction from noise event images, of the saddle-point CNN over the Gaussian CNN, due to the Gaussian model's invalidity at low intensities, this provides further evidence of the saddle-point model's superiority as the more generalizable model. While the overall quality of the saddle-point CNN's reconstructed images is low compared to the reconstructions from the experimentally trained CNN, we have demonstrated that there is real practical utility in utilizing the saddle-point model for noise image reconstructions, as when employed in a mixed training set of experimental and saddle-point model synthetic data, the resulting CNN tangibly outperforms the purely experimentally trained CNN. Given the high degree of similarity in the synthetic noise images between the saddle-point and the Poisson models, these conclusions can be extended to the Poisson model as well. Higher-quality reconstructions from a synthetically trained saddle-point model CNN are likely achievable using more sophisticated per-pixel sampling approaches than the simplistic ones employed here.

\section{Discussion} \label{section: discussion}
 

\subsection{Further Remarks} \label{subsection: Further Verifications}
With the parameter values established, we now examine their physical plausibility and discuss limitations of the analysis.
\par

\textbf{Physical Interpretation of $\theta(\lambda)$.} We deduced the functional form of $\theta(\lambda)=c_1+c_2\sqrt{\lambda}+c_3\lambda$ from the numerically determined $\theta$ values. Based on the nature of this form, as well as how the coefficients $c_1$, $c_2$, and $c_3$ change with respect to the bias settings (discussed in Section~\ref{s-section: non-default settings}), we can make inferences about the physical interpretation of these components.
\par
\emph{$c_1$ (light-independent offset).} Because $c_1$ does not depend on $\lambda$, it represents a baseline offset that exists even in darkness. The fact that $c_1$ is controlled mainly by `bias\_hpf' suggests this offset is set by the part of the pixel that removes slow changes (the ``high-pass'' side): as `bias\_hpf' is increased, that circuitry likely allows more unwanted background current to influence the event decision, making $c_1$ grow. The consistent difference between positive and negative events (positive $c_1$ larger) indicates that the two event polarities are not implemented in perfectly identical ways. Since these background currents typically increase with temperature, $c_1$ is expected to be the most temperature-sensitive coefficient.
 
\par
\emph{$c_2$ (shot-noise coupling).} The $\sqrt{\lambda}$ dependence is characteristic of photon shot noise, the unavoidable randomness in photon arrival. Thus, $c_2$ quantifies how strongly these random fluctuations translate into the event decision and can produce spurious events in a static scene. The slight decrease of $c_2$ as `bias\_hpf' increases suggests that stronger ``high-pass'' settings reduce how much of this randomness appears at the thresholding stage (for example, by reducing the effective amplification of fast fluctuations). The weak dependence on `bias\_fo' indicates that this effect is governed mainly by the differencing/thresholding part of the pixel rather than the slower reference-setting path. The small polarity difference (negative $c_2$ slightly larger) again suggests minor asymmetries between the two event pathways.
 
\par
\emph{$c_3$ (brightness-dependent leakage).} The linear dependence on $\lambda$ indicates an effect that grows in proportion to the mean brightness: as the scene becomes brighter, a larger fraction of the average signal effectively ``leaks'' into the offset term. The strong dependence of $c_3$ on `bias\_fo' implies that this leakage is set primarily by the circuitry that defines the pixel's slow reference level (the ``low-pass'' side): changing `bias\_fo' changes how much the average intensity can influence the event condition. The strong polarity asymmetry (positive $c_3$ larger than negative) suggests this brightness-dependent leakage is more pronounced for increasing intensity than for decreasing intensity, consistent with the two polarities having different internal signal paths.
\par
As noted in Section~\ref{s-section: non-default settings}, there are exceptions to the trends on which we are basing these inferences (particularly at the extremes of the bias settings), and the parameter space of `bias\_fo' and `bias\_hpf' is large in comparison to our sampling density. The true nature of $\theta(\lambda)$ may be more complex, but this form captures much of the first-order behavior, especially near the default bias settings. 
 
\textbf{Caveats.} Our stated results should be interpreted with several limitations in mind. Near the default bias settings our model provides consistent parameter estimates; however, at more extreme bias values we observe anomalous behavior (Section~\ref{s-section: non-default settings}). In particular, the inferred threshold $B$ remains stable near the default `bias\_diff\_on/off' setting but becomes less reliable at the extremes. Estimates of $\theta(\lambda)$ (and hence of $c_1,c_2,c_3$) do not necessarily degrade in the same way, but additional unexpected effects appear.
 
One such effect is strong polarity asymmetry. Under most conditions the measured positive and negative noise-event probabilities are comparable in magnitude, yet for `bias\_fo'$\,=0$ and `bias\_hpf'$\,=+120$ the peaks differ by several orders of magnitude: $\max \hat{P}_+ \approx 7\times 10^{-8}$ while $\max \hat{P}_- \approx 2\times 10^{-5}$. An even more extreme case occurs for `bias\_fo'$\,=-35$ and `bias\_hpf'$\,=+120$ (Fig.~\ref{s-fig: fo -35 hpf +120}), where $\max \hat{P}_+ \approx 1\times 10^{-12}$ and $\max \hat{P}_- \approx 3\times 10^{-7}$. In this configuration positive noise events are suppressed so strongly that, away from the peak region, we often observe no positive noise events at all.
 
A potential contributor is that we have implicitly treated the bias controls as independent. This approximation appears reasonable near the defaults, but it may break down at extreme settings. For example, when `bias\_fo' and `bias\_hpf' are driven far from their default values, the effective sensitivity (and therefore the relevant threshold parameter $B$) may change even if `bias\_diff\_on/off' is held fixed.
 
There is also a systematic discrepancy in the fitted S-curve families. While the predicted S-curves derived from the Heaviside-event model broadly match the reported shapes, the ordering of the intermediate-intensity curves differs. In the reported S-curve data (Fig.~\ref{fig:reported s-curves}), the S-curves at 2.997 and 30.409~lux lie to the left of the 299.684~lux curve, whereas in our fitted family (Fig.~\ref{fig:fitted S-curves}) the curves appear ordered monotonically from left to right with decreasing reference intensity. This disagreement could indicate either that $B$ is not strictly constant with respect to intensity, or that the chosen parameterization of $\theta(\lambda)$ is not flexible enough to capture the full leakage behavior. However, we suspect that the cause of this reported behavior may instead be due to the impact of outlier pixels reporting improperly high event counts in this intensity regime, as neither of the sources of S-curve data employed in Fig.~\ref{fig:reported s-curves} discusses the filtering of outlier pixels.
 
Finally, although we incorporate reported S-curve measurements for the default-bias positive-polarity case \citep{mcreynolds2024stepresponse, IMX636sample}, comparable S-curve measurements are not available to us for non-default bias settings. As a result, those fits lack a key external constraint. Direct S-curve measurements under such cases would be particularly valuable for diagnosing and potentially resolving the anomalies described above.

\subsection{Practical Guidance} \label{subsection: Practical Guidance}
 
\subsubsection{Interpretation of S-Curves} \label{subsubsection: S-curve interpretation}
As previously noted, while S-curves have become a somewhat common approach in attempts to determine $B$, there have been differing interpretations of how to properly determine $B$ from S-curves, with some authors taking a 50\% probability level interpretation and others taking 100\%. Our model of the Heaviside event probability distributions shows that interpreting the S-curves is not as straightforward as picking the correct probability threshold to determine $B$. 
\par
For instance, we determined the value of $B$ to be approximately 0.15, which is not near the 50\% or 100\% probability thresholds on any of our computed probability curves (Fig.~\ref{fig:fitted S-curves}). If anything, a log contrast of 0.15 appears to be where the higher-intensity S-curves go to 0. We instead deduce a few general properties useful for correctly interpreting S-curves. 
\par
\begin{figure} 
  \centering
  \begin{minipage}[b]{0.48\textwidth}
    \centering
    \includegraphics[width=\textwidth]{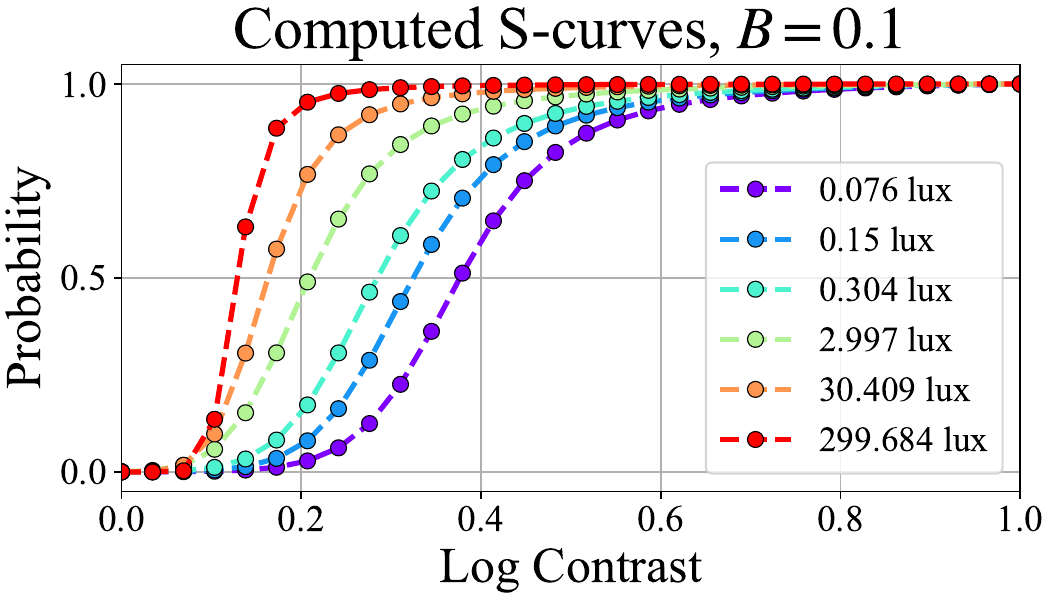}
  \end{minipage}
  \begin{minipage}[b]{0.48\textwidth}
    \centering
    \includegraphics[width=\textwidth]{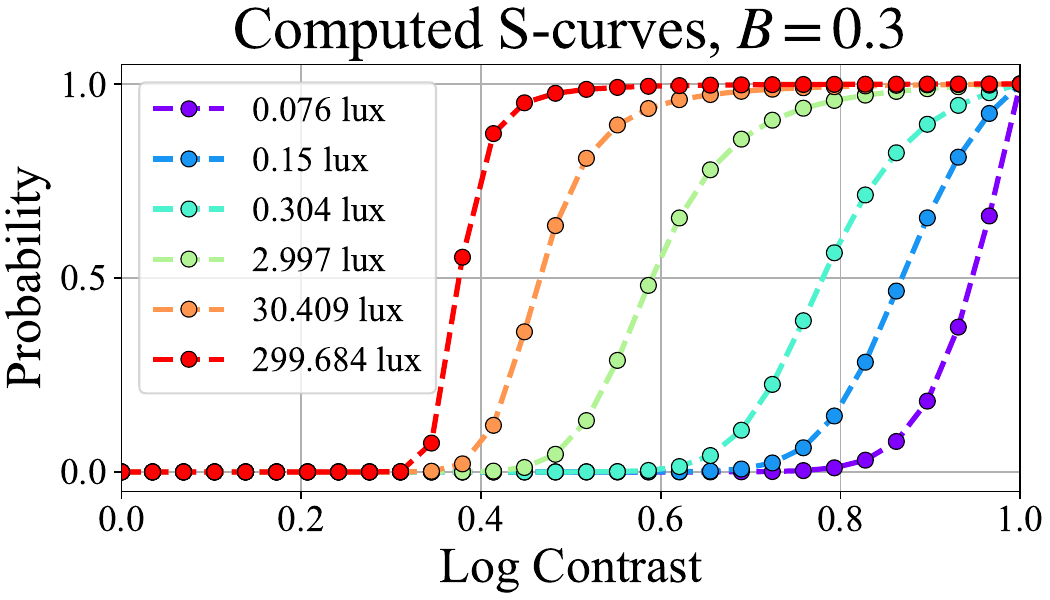}
  \end{minipage}
 
  \caption{Computed S-curve families (saddle-point approximation, 100 sampled pixels as in Section~\ref{section: param determ}) for $B=0.1$ (top) and $B=0.3$ (bottom), with $\alpha=4.5$ and $\theta(\lambda)$  as determined for the default bias settings. Increasing $B$ shifts the S-curves rightward; decreasing $B$ shifts them leftward.}
  \label{fig: S-curves for B=0.1 and B=0.3}
\end{figure}

$B$ primarily influences the horizontal placement of the S-curves; a larger $B$, requiring a larger log contrast to saturate the probability, shifts the curves to the right, while a smaller $B$ shifts them to the left, as can be seen in Fig.~\ref{fig: S-curves for B=0.1 and B=0.3}. It is important to note that we have taken $B$ to be a constant value and that the phenomenon of lower-intensity S-curves being further right of higher-intensity ones is not a result of $B$ increasing at lower intensities, as might naively be suggested if one only considers the apparent event conditions, Eqs.~(\ref{eq:1}) and~(\ref{eq:2}). This phenomenon is the result of the presence of $\theta$ in the fundamental conditions, Eqs.~(\ref{eq:3}) and~(\ref{eq:4}). It might be said that ``the effective threshold increases at lower intensities,'' but we caution that such language will likely serve to perpetuate confusion around this topic.
\par
The higher-intensity curves should be ``nearer'' to the value of $B$, though they need not actually reach the value of $B$ itself. All else being equal, a reasonable general rule is that the correct $B$ is the one that aligns the horizontal position of the high-intensity curves with observations.
 
\par
\begin{figure} 
  \centering
  \begin{minipage}[b]{0.48\textwidth}
    \centering
    \includegraphics[width=\textwidth]{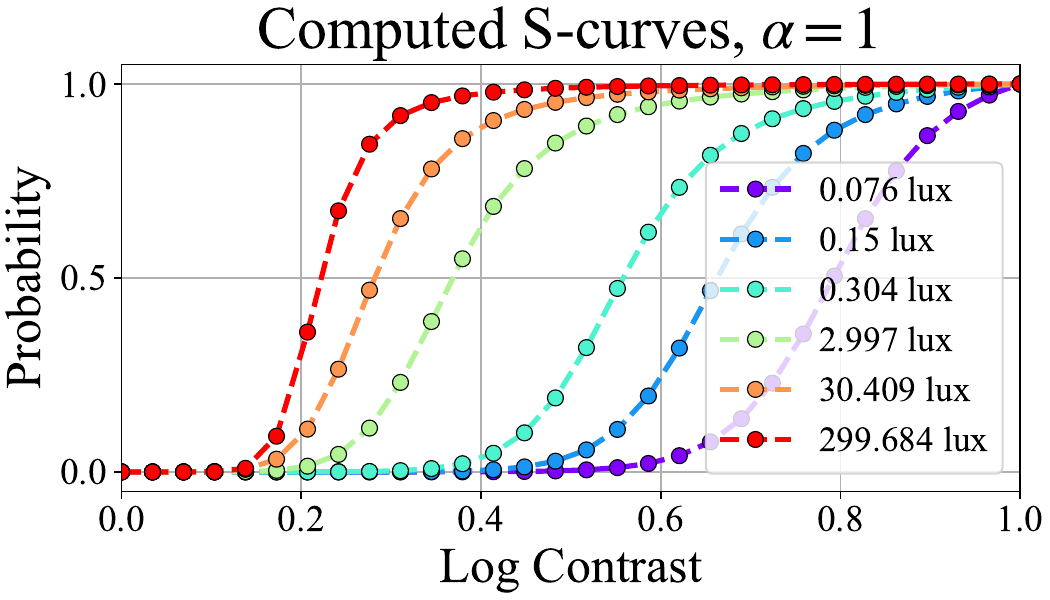}
  \end{minipage}
  \begin{minipage}[b]{0.48\textwidth}
    \centering
    \includegraphics[width=\textwidth]{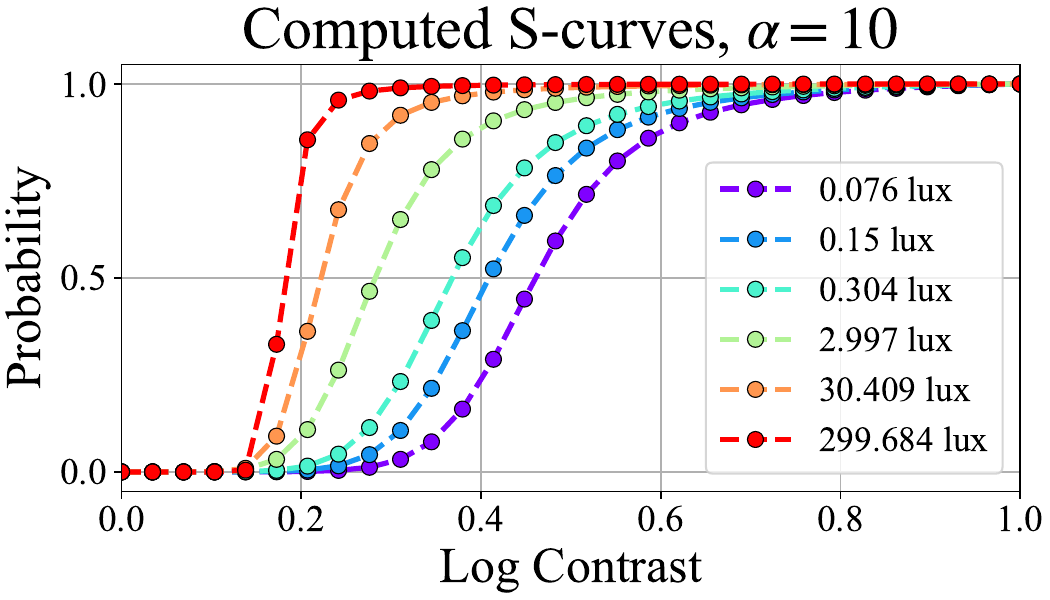}
  \end{minipage}
 
  \caption{Computed S-curve families (saddle-point approximation, 100 sampled pixels as in Section~\ref{section: param determ}) for $\alpha=1$ (top) and $\alpha=10$ (bottom), with $B=0.15$ and $\theta(\lambda)$  as determined for the default bias settings. A larger $\alpha$ causes the S-curves from different baseline intensities to be more tightly packed together.}
  \label{fig: S-curves for alpha=1 and alpha=10}
\end{figure}

The conversion factor $\alpha$ primarily influences how tightly packed the S-curves, from different $I_0$ intensity levels, are to each other. A smaller $\alpha$ results in more sparsely packed S-curves, while a larger $\alpha$ brings them more tightly together, as seen in Fig.~\ref{fig: S-curves for alpha=1 and alpha=10}.
 
\begin{figure} 
  \centering
  \begin{minipage}[b]{0.48\textwidth}
    \centering
    \includegraphics[width=\textwidth]{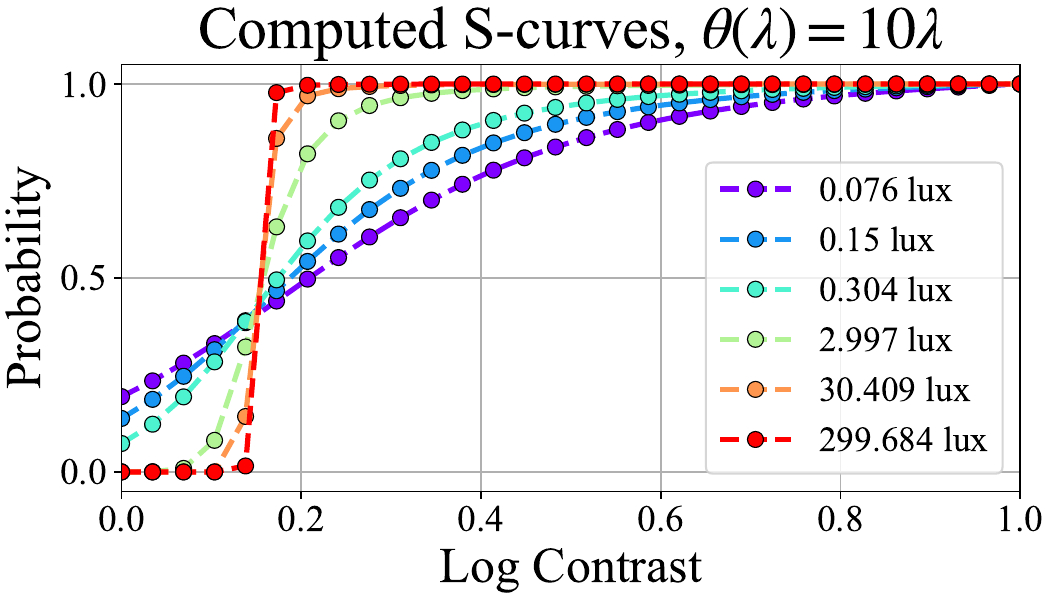}
  \end{minipage}
  \begin{minipage}[b]{0.48\textwidth}
    \centering
    \includegraphics[width=\textwidth]{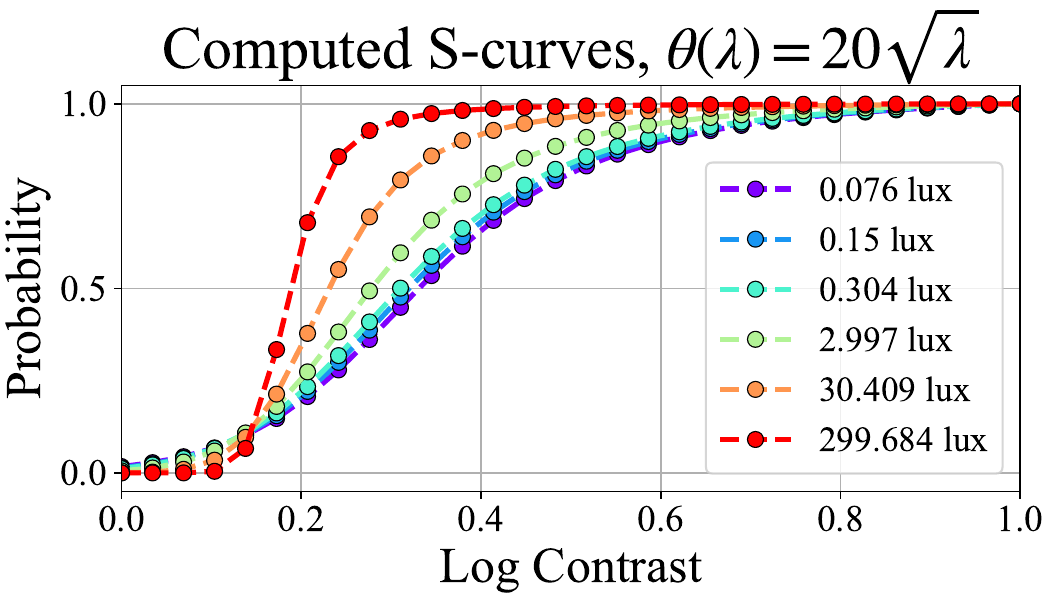}
  \end{minipage}
 
  \caption{Computed S-curve families (saddle-point approximation, 100 sampled pixels as in Section~\ref{section: param determ}) for $\theta(\lambda)=10\lambda$ (top) and $\theta(\lambda)=20\sqrt{\lambda}$ (bottom), with $B=0.15$ and $\alpha=4.5$. Different functional forms of $\theta(\lambda)$ produce markedly different S-curve shapes, illustrating its complex influence on the curve family.}
  \label{fig: S-curves for different theta expressions}
\end{figure}

The effect of $\theta(\lambda)$ on S-curves is more complex. Unlike $\alpha$, which scales the spacing between all S-curves in a fairly uniform manner, $\theta$ influences the curves selectively, altering how S-curves at different baseline intensities $I_0$ relate to one another. As a result, $\theta$ can substantially reshape the S-curve family, for instance by compressing some curves together while leaving others widely separated. Different functional forms of $\theta$ produce markedly different S-curve families, as illustrated in Fig.~\ref{fig: S-curves for different theta expressions}. A concrete example of this selective influence is visible in the default bias settings case (Figs.~\ref{fig:fitted S-curves}, \ref{fig: S-curves for B=0.1 and B=0.3}, and~\ref{fig: S-curves for alpha=1 and alpha=10}), where the gap between the 0.304 and 2.997~lux S-curves is notably larger than between neighboring curves at other intensities. Despite this complexity, for sufficiently well-behaved forms of $\theta$, the "step" of the higher-intensity S-curves remains near the correct value of $B$.
 
We caution that while these are the aspects of the S-curves that $B$, $\alpha$, and $\theta$ influence \textit{primarily}, their influences are not strictly independent and may overlap, though typically marginally. For example, in Fig.~\ref{fig: S-curves for B=0.1 and B=0.3}, the different values of $B$, even though they have the most dramatic impact on the horizontal location of the curves, do exert some influence on their density.
\par
Therefore, we advise that the \textit{most} correct way to interpret an S-curve is to find the values of the parameters that give computed S-curves that align best with the observed ones. In the case that this approach is not desirable, $B$ can be estimated, though not particularly accurately, as a value ``near'' (often slightly below) the step of the high-intensity S-curves. While the horizontal spread of the S-curves can be used to estimate whether $\alpha$ is small or large, the exact values of $\alpha$ and $\theta$ require more direct fitting to establish an accurate estimate. 
 
\subsubsection{How to Use Noise2Params}
 
Having presented our probabilistic model for the description of EC behavior and the results of Noise2Params in determining camera-specific parameter values, here we summarize the process and give advice on the implementation of Noise2Params. 
\par
\textbf{Experimental.} The minimum requirement of Noise2Params is EC recordings of static, uniform scenes at varying levels of ambient intensity. The scene should be illuminated with light dominated by wavelengths in the camera's optimal quantum efficiency regime (for commercial ECs this typically coincides with the most sensitive regime of human vision). Care should be taken to ensure that any light sources present are devoid of flickering. An accessible control light source, known to be flicker-free, such as an incandescent bulb on DC power, can be used to compare against other light sources to ensure the absence of unexpected flickering. 
\par
\textbf{Parameter Determination.} With the recordings taken, and after the removal of any outlier pixels (approaches to which are detailed in Section~\ref{s-section: hot pixels}), the positive and negative noise event probabilities from a recording, at a given intensity, can be estimated by means of Eq.~(\ref{eq: prob estimate}). With the estimated probability $\hat{P}_\pm$ found as a function of intensity, the parameters of the distributions, $B$, $\alpha$, and (the coefficients of) $\theta$, can be determined by fitting to $\hat{P}_\pm$. The use of appropriate bounds (such as restricting $B\in\left(0,1\right]$) when performing the fit is key to ensuring the meaningfulness of the results and to guarding against overfitting. Order of magnitude bounds for $\alpha$ may be estimated by a similar approach as we took in Section~\ref{subsection: Further Verifications}. Bounds for $\theta$ are less apparent, but they are less critical, as $\theta$ can be solved for numerically (given placeholder values of $B$ and $\alpha$) at each given intensity. This can then be used to determine an functional form of $\theta(\lambda)$. This form will likely have coefficients of its own, which are then among the parameters as a whole to be fitted. 
\par
\textbf{S-Curve Use.} If one has access to S-curve data for an EC, this can be used to provide even tighter bounds for the parameters, such as the reported S-curves in Fig.~\ref{fig:reported s-curves} suggesting the tighter bounds $B\in\left(0,0.3\right]$. Furthermore, the parameter fitting can be performed against the noise event probability distributions and the S-curves simultaneously, providing an additional level of confidence. We stress, however, that the use of S-curves in the parameter determination is not strictly necessary, and that it can be performed against $\hat{P}_\pm$ alone, as long as sufficiently appropriate parameter bounds can be determined.
\par
\textbf{Choice of Distribution.} Either the Poisson or the saddle-point approximation distributions should be used when performing the fit; both should yield similar results for the same parameter values. If the Poisson and saddle-point distributions differ significantly for a given choice of parameter values, this may indicate an erroneous value. Computationally, the saddle-point distributions are likely to be less intensive than the Poisson ones (one numerical root-finding evaluation vs.\ many double summations). Hence, we recommend employing the saddle-point distributions for the fit, and then evaluating the Poisson distribution for comparison. Given their invalidity at low intensities, the Gaussian distributions should only be employed for the fitting if the high-intensity regime is the only one of concern. 
\par
\textbf{Numerical Considerations.} Regarding the numerical evaluation of the saddle-point and Poisson distributions, due to the typically small magnitude of the empirical probabilities (a peak on the order of $10^{-7}$ at default bias settings), care should be taken to achieve numerical stability. In our experience, anomalous intermediate results were sometimes encountered, such as a significant difference between $P_+$ and $P_-$, despite their analytic equivalence. In all cases, these were found to be erroneous and attributable to numerical issues, which were then remediable by the use of more careful numerical approaches. 
 
\subsection{Applications and Impact} \label{subsection: applications}
The scope of our approach is as broad as the use cases of ECs, particularly in photon-limited or high-dynamic-range regimes where the fidelity of individual events matters.
\par
\emph{Calibration, metrology, and quality assurance.} Estimating $(B,\alpha,\theta,c_V)$ from a uniform, static scene enables per-pixel calibration maps and device screening without specialized targets or strobed setups. Spatial maps of $B$ and $\theta$ can equalize thresholds across the array, guide bias selection to reduce false-event heterogeneity, and serve as factory acceptance tests via S-curve consistency. Because our approach captures both low- and high-intensity limits, it remains valid across illumination levels, a property not guaranteed by earlier heuristics.
\par
\emph{Model-based denoising and reconstruction.} The fitted parameters define a likelihood for the observed event stream, enabling maximum-likelihood or Bayesian filters for denoising and improvement in image reconstructions. Inverse methods currently reliant on ad hoc regularizers can instead draw priors from our photon-statistics model, improving fidelity in both sparse and saturated regimes.
\par
\emph{Low-flux bioimaging.} Many biophotonics applications operate in the low-intensity regime, including single-molecule localization microscopy (SMLM) \citep{cabriel2023event}, fluorescence correlation measurements, and label-sparse reporters, where individual events can correspond to biologically meaningful fluctuations. Our framework supports setting $B$ for high event specificity, relates events to molecular photon budgets via $\alpha$, and enables probabilistic discrimination between noise events and events arising from actual dynamics.
\par
\emph{Astronomy, space surveillance, and night-sky sensing.} ECs can mitigate saturation from bright sources while preserving sensitivity to dim targets, and have shown promise in astronomical imaging and space-domain-awareness settings \citep{yadav2025neuromorphicastronomy,afshar2020eventSSA,oliver2025eventNoiseSDA}. Our model characterizes low-intensity and dark-current-dominated behavior, including potentially temperature-dependent leakage through $\theta$, and supports post hoc rejection of spurious events in photometric and tracking estimates.
\par
\emph{High-speed metrology and tracking.} Applications such as particle image velocimetry \citep{wang2020stereoEventPTV}, laser speckle sensing, and fast microfluidic tracking demand event-level timing precision. Likelihood-based event validation from our model suppresses false triggers during rapid contrast changes, raising effective temporal resolution without the aggressive smoothing that would blur dynamics.
\par
\emph{Cross-sensor and hybrid systems.} When ECs are paired with TCs \citep{BrandliChristian2014Rhvd}, other frame-based sensors, or SPAD arrays \citep{muglikar2025eventSPAD}, $(\alpha,\theta)$ provides a physically interpretable bridge between units, enabling cross-calibration and sensor fusion consistent with photon-counting statistics. This is useful for HDR fusion pipelines and for aligning event rates to photon count rates in hybrid low-light systems.
\par
\emph{Synthetic data and benchmarks.} A physically grounded generator of noise and Heaviside events yields more realistic simulators for training and evaluating reconstruction and recognition algorithms. Because the parameters are measurable per device, simulators can be matched to specific hardware rather than relying on universal, potentially inaccurate noise assumptions.
\par
These modes share a theme: by converting photon-statistical structure into identifiable parameters, ECs can be operated with specification-driven settings and analyzed via model-based inference rather than heuristic thresholds.

\section{Conclusion} \label{section: conclusion}
In this work, we developed a foundational analytic model for event cameras grounded in the discrete, Poisson nature of photon arrivals. The model describes EC behavior across a unified probabilistic framework, encompassing both static scene noise events and the event response to Heaviside (square-wave) intensity stimuli that underlie S-curves. Three formulations of the probability distributions were derived: an exact Poisson treatment valid at all intensity regimes, a computationally efficient saddle-point approximation with comparable accuracy, and a Gaussian approximation appropriate for mid to high intensities. That these distinct EC phenomena arise from a single set of physically interpretable parameters illustrates a connection that, to our knowledge, has not previously been explored analytically.
\par
Enabled by this model, we proposed Noise2Params, a method for determining camera-specific values of the log-contrast threshold $B$, the lux-to-photon conversion factor $\alpha$, and the leakage term $\theta$ (which was found to be intensity dependent, with a parametric form $\theta(\lambda) = c_1 + c_2\sqrt{\lambda} + c_3\lambda$), via error minimization against empirical noise-event probability distributions. Noise2Params can be performed using solely static scene noise observations, offering an experimentally accessible approach to parameter determination that requires no specialized dynamic light sources; it can also be supplemented with S-curve data for additional constraint. The fitted parameter values are reported in Table~\ref{table:parameter_values}.
\par
We validated the model and the determined parameter values through several independent means. The Poisson and saddle-point distributions showed strong quantitative agreement with measured noise-event probabilities across the full intensity range (Section~\ref{section: param determ}). We further supported the model and demonstrated its practical utility by training convolutional neural networks on synthetic noise images generated from our probability distributions: a CNN trained on a mixed dataset of saddle-point synthetic and experimental noise images outperformed a CNN trained on experimental data alone across a majority of image similarity metrics (Section~\ref{sec: noise_event_images}). Additionally, our model clarified the interpretation of S-curves, showing that the determination of $B$ from S-curves is more nuanced than selecting a fixed probability threshold, and that the apparent intensity dependence of the threshold is attributable to $\theta$ rather than to a varying $B$ (Section~\ref{subsubsection: S-curve interpretation}).
\par
While Noise2Params addresses an immediate need in the EC community for accessible and rigorous parameter determination, the underlying model is intended to be foundational in a broader sense. Because it is built from first principles of photon statistics rather than empirical curve fitting, it extends naturally to any EC operating regime in which photon-level considerations are relevant. We expect the most impactful applications to be in domains where individual events carry semantic weight, including low-light microscopy, faint-object astronomy, and high-speed metrology (Section~\ref{subsection: applications}).
\par
Several avenues for future work remain. These include the use of $\theta(\lambda)$ models of greater complexity, parameter estimation at higher precision (including per-pixel determination of parameters beyond $B$), investigation of environmental factors such as temperature for inclusion in the model, and extension of the framework to multi-spectral or wavelength-resolved EC operation. We anticipate that the probabilistic foundation established here will serve as a basis for continued development of quantitative, model-driven approaches to event camera science.


\section*{Funding}
This work was supported by the NASA MIRO CUNY program at Hunter College. 

\section*{Data and Code Availability}
Code and data associated with the deep-learning noise image reconstruction portion of this paper are available: \url{https://github.com/yagoiroot/Noise2Params}

\bibliography{Bib}


\subfile{supplemental_info}

\AddToHook{enddocument/afteraux}{%
\immediate\write18{
cp output.aux main.aux
}%
}
\end{document}

%% file: supplemental_info.tex
\clearpage
\onecolumn

\setcounter{figure}{0}
\setcounter{table}{0}
\setcounter{equation}{0}
\setcounter{section}{0}
\renewcommand{\thefigure}{S\arabic{figure}}
\renewcommand{\thetable}{S\arabic{table}}
\renewcommand{\theequation}{S\arabic{equation}}
\renewcommand{\thesection}{S\arabic{section}}

\renewcommand{\theHsection}{S\arabic{section}}
\renewcommand{\theHsubsection}{\theHsection.\arabic{subsection}}
\renewcommand{\theHsubsubsection}{\theHsubsection.\arabic{subsubsection}}
\renewcommand{\theHfigure}{S\arabic{figure}}
\renewcommand{\theHtable}{S\arabic{table}}
\renewcommand{\theHequation}{S\arabic{equation}}

\ifSubfilesClassLoaded{\maketitle}{\section*{Supplemental Information}}

\section{Derivation of Distributions} \label{section: derivation of distributions}
We derive three forms of the event probability distributions, first using an exact Poisson treatment, then a Gaussian approximation, and lastly a saddle-point approximation. We derive these for the general case that $I\ne I_0\rightarrow\lambda\ne\lambda_0$, as the distributions for the static scene case, in which $I=I_0$, can be found as a simplification of these general results. 
 
\subsubsection*{Exact Poisson}
Given that $n,n_0$ are Poisson-distributed random variables, we can construct exact expressions for $P_\pm$ by summing over the products of all probabilities for $n,n_0$ that satisfy the conditions. Let $s_0^+,s^+$ denote the sets of allowed values of $n_0, n$ for the condition $Z_+>0$ and $s_0^-,s^-$ denote the sets of allowed values of $n_0, n$ for the condition $Z_-<0$ ($Z_\pm$ is given by Eq.~(\ref{m-eq:Z_pm lambda dependent})). The probabilities are then 
\begin{equation}
    P_\pm=\sum_{n_0\in s_0^\pm, n\in s^\pm}P(n_0,n).
\end{equation}
The positive event condition $Z_+>0 \rightarrow n>e^{B}\left(n_0+\theta(\lambda_0)\right)-\theta(\lambda)$. Since $n$ is evidently an integer, $s^+$ is all values of $n$ which are greater than $\floor{e^{B}\left(n_0+\theta(\lambda_0)\right)-\theta(\lambda)}+1$ (the $+1$ is necessary to meet the condition of strictly greater than 0, as the floor term could be 0 itself). We have moved the restriction on $n$ and $n_0$ to just $n$, so $s_0^+$ is just its usual integer values. The probability of a positive noise event is then 
\begin{equation}\label{eq: P_+ pois}
    P_+^{\text{Pois}}(\lambda,\lambda_0)=\sum_{n_0=0}^\infty \sum_{n=\floor{e^{B}\left(n_0+\theta(\lambda_0)\right)-\theta(\lambda)}+1}^\infty \frac{e^{-2 \lambda}\lambda^{n+n_0}}{n!n_0!}.
\end{equation}
\par
The negative event condition $Z_-<0\rightarrow n<e^{-B}\left(n_0+\theta(\lambda_0)\right)-\theta(\lambda)$. Here we now have an upper limit on the values of $n$, so $s^-$ is all the integer values of $n$ less than $\ceil{e^{-B}\left(n_0+\theta(\lambda_0)\right)-\theta(\lambda)}-1$. Here we have used the ceiling $-1$ to avoid including the case $n=n_0-\theta(1-e^{-B})$ (again, so as to meet the strict inequality). Again, we have moved the restriction on $n$ and $n_0$ to just $n$, so $s_0^-$ is just its usual integer values. The probability of a negative event is then
\begin{equation}\label{eq: P_- pois}
    P_-^{\text{Pois}}(\lambda,\lambda_0)=\sum_{n_0=0}^\infty \frac{e^{-\lambda}\lambda^{n_0}}{n_0!} \sum_{n=0}^{\ceil{e^{-B}\left(n_0+\theta(\lambda_0)\right)-\theta(\lambda)}-1} \frac{e^{-\lambda}\lambda^{n}}{n!}.
\end{equation}
\par
Given that they are very different in form, it may be surprising that (\ref{eq: P_+ pois}) and (\ref{eq: P_- pois}) give equivalent values, that is, positive and negative events are equally likely. To see this, consider the following.
\par
First, recognize that the positive event condition can be recast so that $P_+=P\left(n>e^{B}\left(n_0+\theta(\lambda_0)\right)-\theta(\lambda)\right)$. We can also rearrange the negative event condition as $P_-=P\left(n_0>e^B n+\theta(\lambda_0)e^B-\theta(\lambda)\right)$. Given that the joint law of independent Poisson random variables is invariant under swapping the two coordinates  \citep{ross2014probability}, it follows that $P_-=P\left(n>e^B n_0+\theta(\lambda_0)e^B-\theta(\lambda)\right)$. In other words, there is nothing special about $n$ or $n_0$; we could just as easily call one the other and nothing about the situation changes besides the label, and as long as we meet $Z_+>0$ and $Z_-<0$, we are free to move the restriction to $n$ or $n_0$. Now, notice that this expression for $P_-$ is exactly equal to the recast $P_+$, hence $P_+=P_-$ and one may choose whichever form, (\ref{eq: P_+ pois}) or (\ref{eq: P_- pois}), one wishes. We have left off the ``Pois'' labels for $P_\pm$ here, as this equivalence between positive and negative event probabilities is an exact property of the distribution, which any valid approach to construction of the distributions should preserve.
\par
Of course, this has all been done under the assumption that positive and negative thresholds have been taken to be the same value just with opposite signs, that is, $B_+=B, B_-=-B$. If this were not the case, that is, if $|B_+|\ne|B_-|$, then one must track the thresholds more carefully and cannot, in general, use (\ref{eq: P_+ pois}) or (\ref{eq: P_- pois}) interchangeably.
\par
In addition to $P_\pm$, one could construct the distribution for the null probability $P_0$, that is, the probability that no noise events are detected. This could be done via direct considerations of $Z_\pm$, $P_0=P(Z_+<0)P(Z_->0)$, or by exploiting normalization, $P_0=1-P_+-P_-=1-2P_\pm$. However, in this work, we are chiefly concerned with $P_\pm$, and so only include their explicit forms. 
 
\subsubsection*{Gaussian Approximation Distributions}
 
While the distributions (\ref{eq: P_+ pois}) and (\ref{eq: P_- pois}) are exact, they are unwieldy and computationally intensive to evaluate. Depending on the intensity regime of interest, certain approximations can be made, yielding analytically simpler solutions. Here we present an approach based on the one taken in  \citep{cao2025noise2image}.
\par
If we are only concerned with the mid to bright light regime, $\lambda\gtrsim10$, then the Poisson distributions for $n, n_0$ can be approximated as normal distributions, $n, n_0 \sim \mathcal{N}(\lambda,\lambda)$. It then follows that $Z_\pm$ are also normally distributed random variables. From the properties of mean and variance, we have
\begin{equation}
    Z_\pm
\sim \mathcal{N}\left(\left(\lambda+\theta(\lambda)-\theta(\lambda_0)e^{\pm B}\right),\left(\lambda+\lambda_0e^{\pm2B}\right)\right)=G_\pm(Z_\pm=z_\pm).
\end{equation}
\par
The conditions for a positive or negative event remain the same, $Z_+>0$ and $Z_-<0$, respectively, but the Gaussian approximation allows us to replace the summations in the probabilities with integrals. Hence, the probability of positive events is $P(Z_+>0)=\int_0^\infty G_+(z_+)\,\text{d}z_+$ and the probability of negative events is $P(Z_-<0)=\int_{-\infty}^0 G_-(z_-)\,\text{d}z_-$. Performing the integrals, this yields
\begin{equation}
    P_\pm^{\text{Gaus}}(\lambda,\lambda_0) \approx \frac{1}{2}\pm\frac{1}{2}\text{erf}\left(\frac{\lambda-e^{\pm B}(\lambda_0+\theta(\lambda_0))+\theta(\lambda)}{\sqrt{2\left(\lambda+e^{\pm 2B}\lambda_0\right)}}\right),
\end{equation}
where $\text{erf}$ denotes the error function. As with the exact Poisson distributions, for the same $B,\;\theta, \; \lambda$, and $\lambda_0$, the distributions $P_\pm^\text{Gaus}$ give equivalent values for positive and negative events. 
 
\subsubsection*{Saddle-Point Approximation Distributions}
While the Gaussian approximation distributions are a great deal simpler and easier to compute than the exact Poisson distributions, they are only valid in the mid to high intensity regimes. If we are interested in the low-intensity regime, $\lambda \lesssim 10$, the Gaussian approximation is not valid and we must use a different approach. A natural choice is the saddle-point approximation method, which we apply here. 
\par
The cumulant generating function (CGF) for $Z_\pm$ is given by 
\begin{equation}
    \kappa_\pm(t)\;=\;\ln\left(\text{E}\left(e^{tZ_\pm}\right)\right)\;=\;\lambda\left(e^t-1\right)+\lambda\left(e^{-te^{\pm B}}-1\right)+t\theta(\lambda)-te^{\pm B}\theta(\lambda_0).
\end{equation}
The probability density function (PDF) is then 
\begin{equation}
    f_{Z_\pm}(z_\pm)=\frac{1}{2\pi i}\int_{c-i\infty}^{c+i\infty} e^{\kappa_\pm(t)-tz_\pm}\,\text{d}t.
\end{equation}
Let $P_+$ denote the cumulative distribution function (CDF) $P(Z_+>0)$ and let $P_-$ denote the CDF $P(Z_-<0)$. Then
\begin{equation} \label{eq: saddle integral 2}
    P_\pm=\int_{0,-\infty}^{\infty,0} f_{Z_\pm}(z_\pm)\,\text{d}z_\pm=\pm\frac{1}{2\pi i}\int_{c-i\infty}^{c+i\infty} \frac{1}{t}e^{\kappa_\pm(t)}\,\text{d}t.
\end{equation}
The integral is dominated by the saddle point $s_\pm$, which maximizes the exponent, that is, $s_\pm$ is determined by the solution to the saddle-point equation, $\kappa_\pm'(s_\pm)=0$ (assuming $\kappa_\pm(t)$ is sufficiently well behaved). We now take a Taylor expansion of $\kappa_\pm(t)$ around $s_\pm$:
\begin{equation}
    \kappa_\pm(t)=\kappa_\pm(s_\pm)+\kappa_\pm'(s_\pm)(t-s_\pm)+\frac{1}{2}\kappa_\pm''(s_\pm)(t-s_\pm)^2+\dots.
\end{equation}
Since we are looking for solutions with $\kappa_\pm'(s_\pm)=0$, we can neglect the first derivative term and retain only the zeroth- and second-order terms,
\begin{equation}
    \kappa_\pm(t)\simeq\kappa_\pm(s_\pm)+\frac{1}{2}\kappa_\pm''(s_\pm)(t-s_\pm)^2,
\end{equation}
which we can substitute back into the integral in (\ref{eq: saddle integral 2}). However, since we Taylor expanded $\kappa_\pm(t)$, we must also Taylor expand the $1/t$ term to the same order, resulting in the integral (we drop the $\pm$ subscript on the $s$ for brevity here)
\begin{equation}
    P_\pm = \pm\frac{1}{2\pi i}\int_{c-i \infty}^{c+i \infty} \left[\frac{1}{s}-\frac{t-s}{s^2}+\frac{(t-s)^2}{s^3} \right]\exp{\left(\kappa_\pm(s)+\frac{1}{2}\kappa_\pm''(s)(t-s)^2\right)}\,\text{d}t.
\end{equation}
Let us assume that $\kappa''(t)>0$ and choose $c=\text{Re}(s)$, then we can make the substitution $t-s=-iy$ and switch the integration bounds to real values,
\begin{equation}
    P_\pm = \pm\frac{\exp(\kappa_\pm(s))}{2\pi}\int_{-\infty}^{\infty} \left[\frac{1}{s}+\frac{iy}{s^2}+\frac{y^2}{s^3} \right]\exp{\left(-\frac{1}{2}\kappa_\pm''(s)y^2\right)}\,\text{d}y.
\end{equation}
Now the integrals are all of the familiar Gaussian integral form. Taking the leading-order result, we have the saddle-point approximation probability distributions
\begin{equation}  \label{eq:saddle dist}
    P_\pm^{\text{saddle}}(\lambda,\lambda_0)\simeq \pm \frac{1}{ s_\pm\sqrt{2\pi\kappa_\pm''(s_\pm)}}e^{\kappa_\pm(s_\pm)},
\end{equation}
where the $\pm$ subscript has been reintroduced and $s_\pm$ must be found by numerically solving the saddle-point equation, $\kappa_\pm'(s_\pm)=0$.

\section{Non-default Bias Settings} \label{section: non-default settings}
 
\subsection{Noise Distributions}
In addition to the observed noise event distributions at default biases, we also made observations of the distributions using non-default biases. The first bias settings that we consider are `bias\_diff\_on' and `bias\_diff\_off'. In order to keep the symmetry between $\pm B$, we incremented `bias\_diff\_on/off' away from their default value by the same amount. Increasing the settings increases $B$ (the EC becomes less sensitive), while decreasing the settings decreases $B$ (the EC becomes more sensitive). The observed noise distributions for `bias\_diff\_on/off': $\pm 10$ can be seen in Fig.~\ref{fig: noise prob diff on/off +- 10}. Observations were made across the range of `bias\_diff\_on/off': $-35$ to $+105$. As might be expected, more sensitive settings lead to a marked increase in noise events. It is worth noting that at $+105$, noise events were suppressed entirely over the time scales at which we made observations. 
 
\begin{figure}[ht]
  \centering
  \begin{minipage}[b]{0.49\textwidth}
    \centering
    \includegraphics[width=\textwidth]{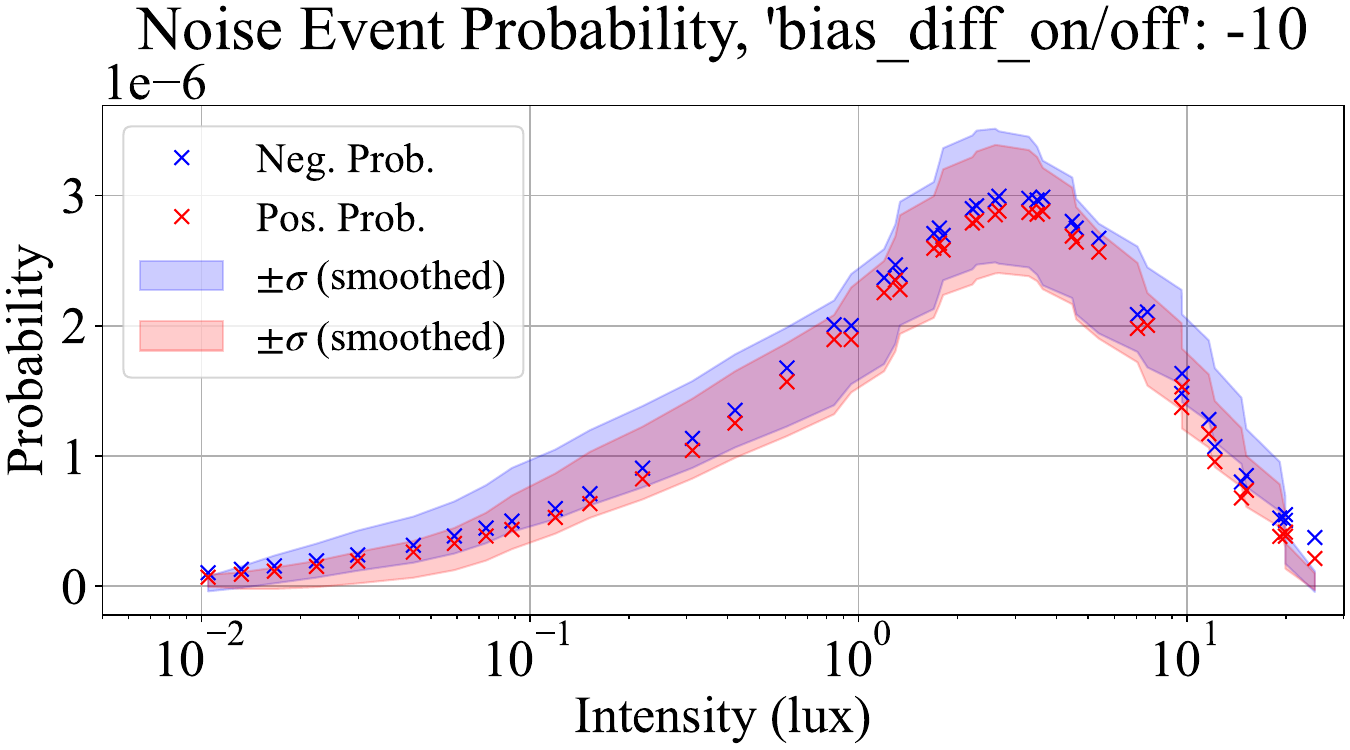}
  \end{minipage}
  \begin{minipage}[b]{0.49\textwidth}
    \centering
    \includegraphics[width=\textwidth]{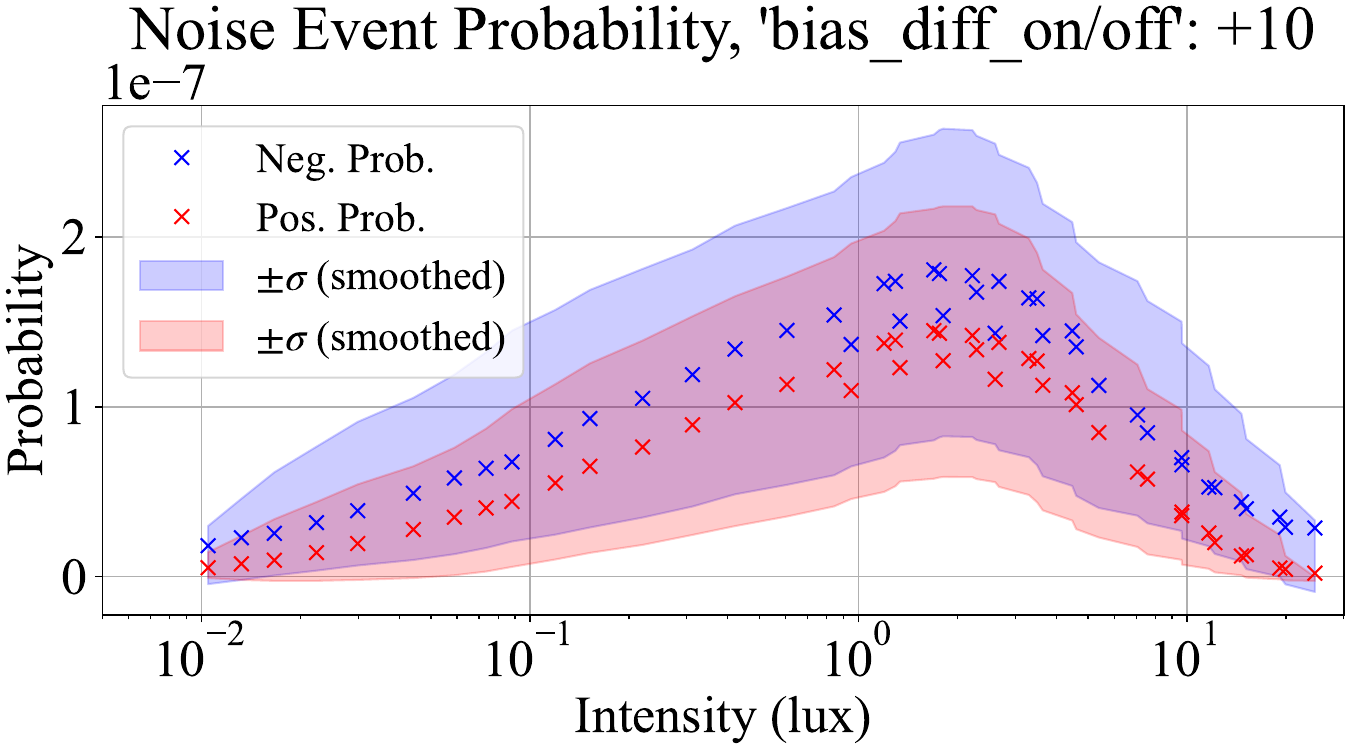}
  \end{minipage}
 
  \caption{Observed $\hat{P}_\pm$, `bias\_diff\_on/off': $-10$ on the left, `bias\_diff\_on/off': $+10$ on the right. Notice the difference in magnitude of the probabilities between the settings. The uncertainty bands were smoothed using a Savitzky-Golay filter (window size 25, order 2) for visual clarity.}
  \label{fig: noise prob diff on/off +- 10}
\end{figure}
 
The next bias settings we consider are `bias\_hpf', which controls the high-pass filter (HPF) of the EC circuitry, and `bias\_fo', which controls the low-pass filter (LPF). Lowering `bias\_hpf' lowers the HPF frequency, widening the bandwidth, while raising it shortens the bandwidth. Raising `bias\_fo' increases the LPF frequency, widening the bandwidth, while lowering it shortens the bandwidth. The observed noise distributions for `bias\_fo': $-35$ and $+55$ can be seen in Fig.~\ref{fig: noise prob fo}, which form the range of `bias\_fo' values for which observations were made. The observed noise distributions for `bias\_hpf': $+10$ and $+60$ can be seen in Fig.~\ref{fig: noise prob hpf}. Observations were made across the range of `bias\_hpf': 0 to 120. Combinations of `bias\_hpf' and `bias\_fo' throughout their ranges were also observed. 
 
\begin{figure}[ht]
  \centering
  \begin{minipage}[b]{0.49\textwidth}
    \centering
    \includegraphics[width=\textwidth]{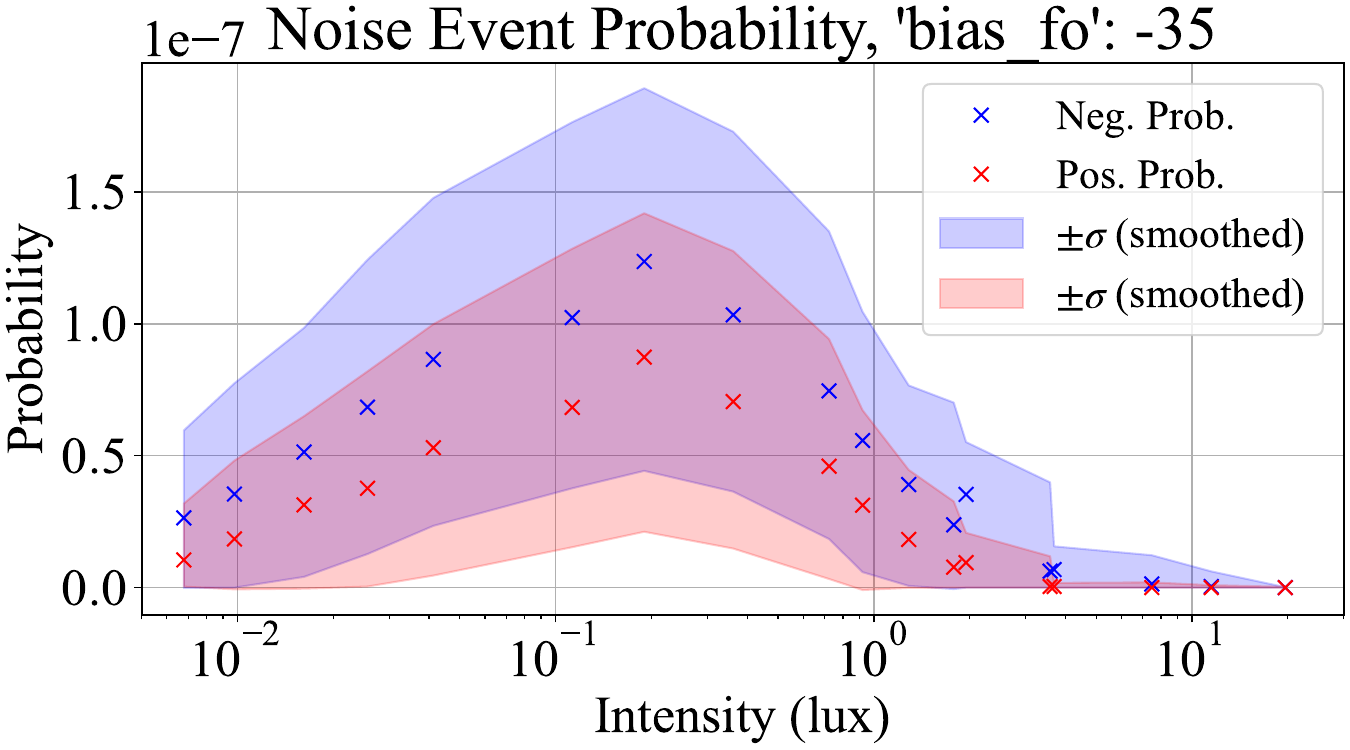}
  \end{minipage}
  \begin{minipage}[b]{0.49\textwidth}
    \centering
    \includegraphics[width=\textwidth]{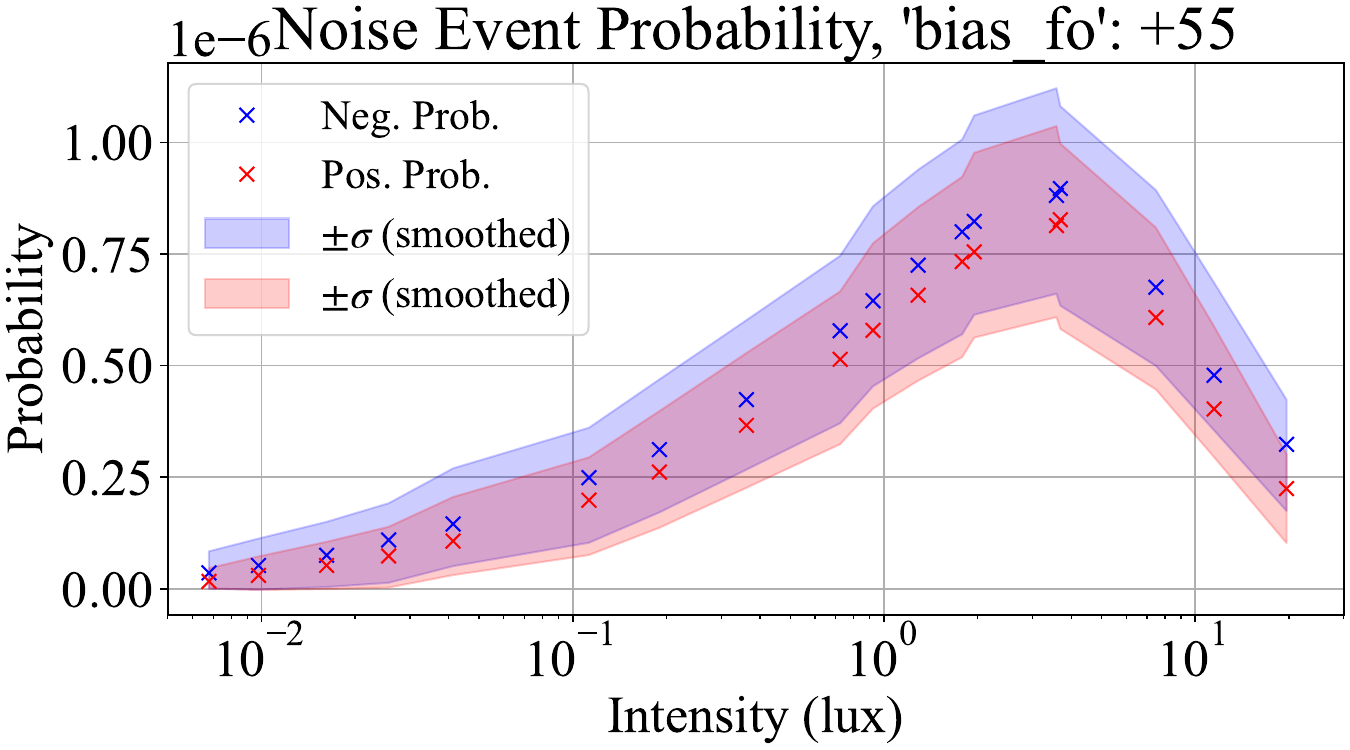}
  \end{minipage}
 
  \caption{Observed $\hat{P}_\pm$, `bias\_fo': $-35$ on the left, `bias\_fo': $+55$ on the right. Notice the shift in the location of the peak in addition to different magnitudes. The uncertainty bands were smoothed using a Savitzky-Golay filter (window size 5, order 2).}
  \label{fig: noise prob fo}
\end{figure}
 
\begin{figure}[H]
  \centering
  \begin{minipage}[b]{0.49\textwidth}
    \centering
    \includegraphics[width=\textwidth]{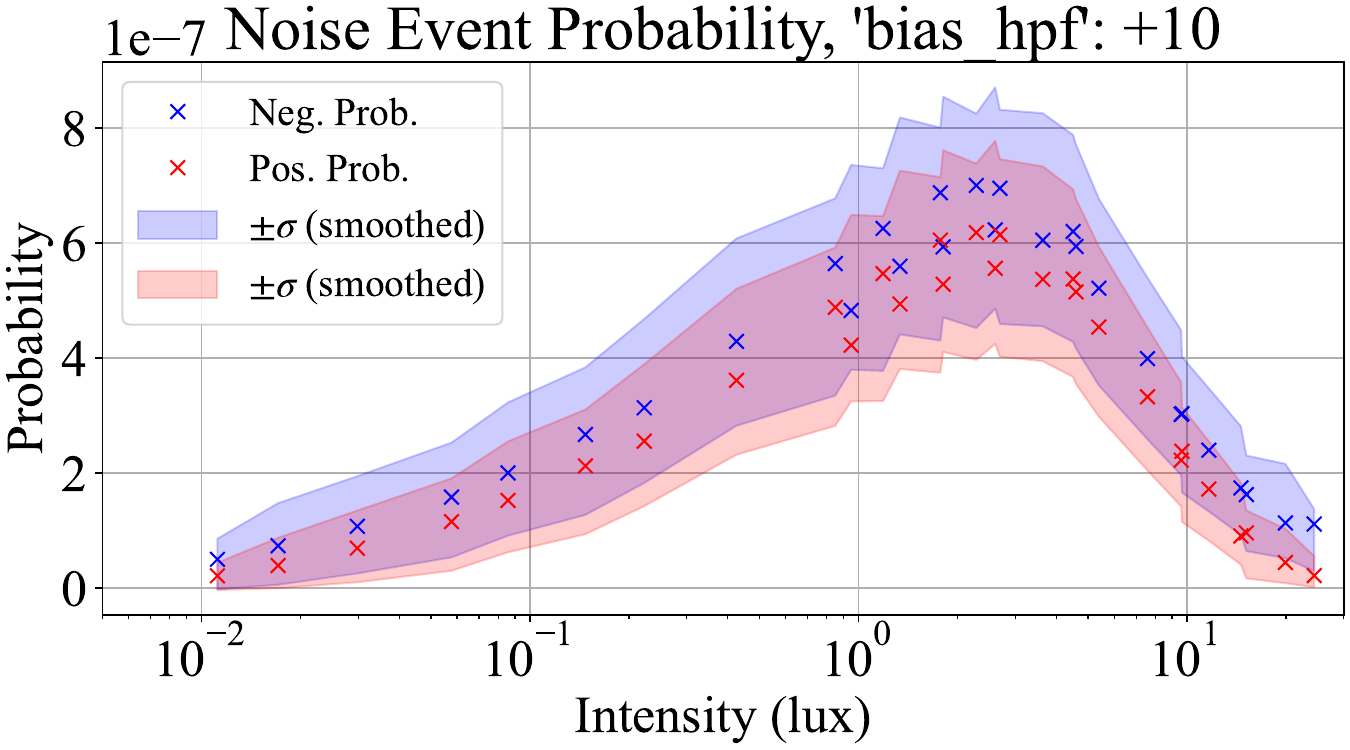}
  \end{minipage}
  \begin{minipage}[b]{0.49\textwidth}
    \centering
    \includegraphics[width=\textwidth]{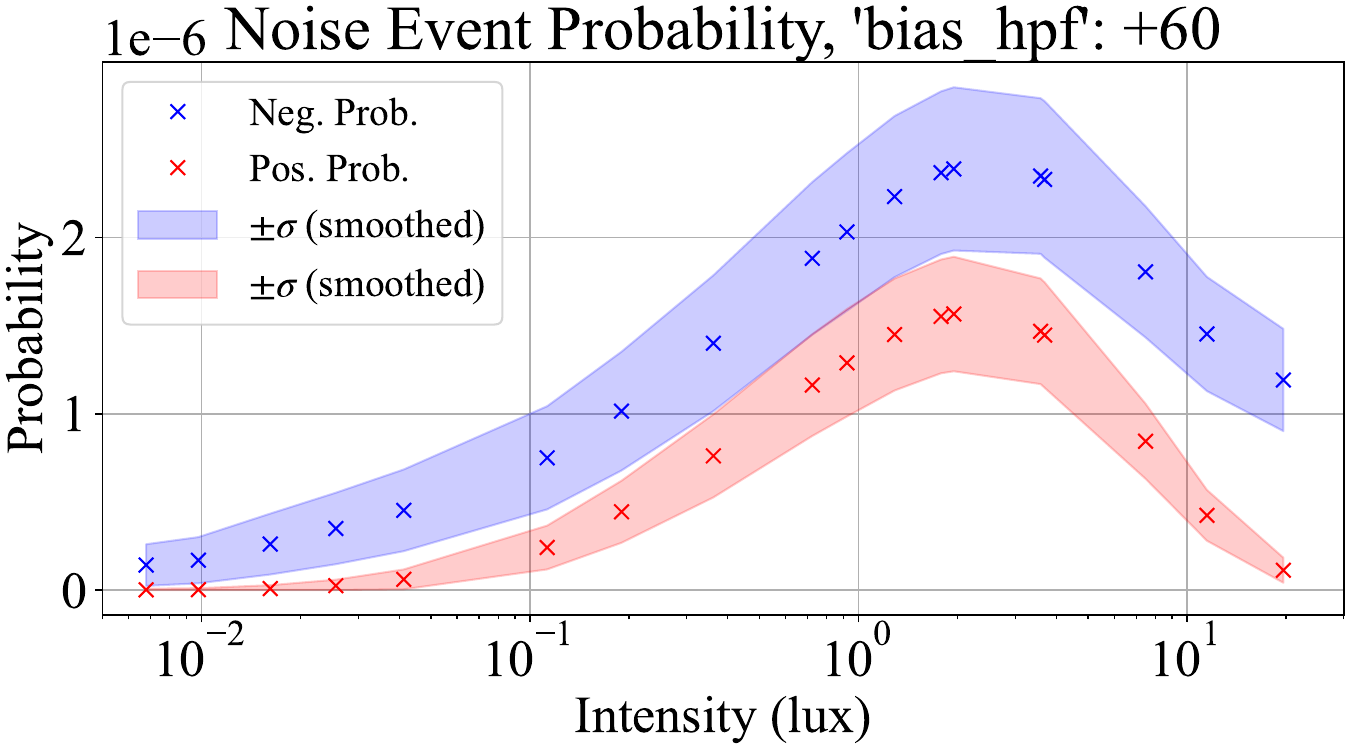}
  \end{minipage}
 
  \caption{Observed $\hat{P}_\pm$, `bias\_hpf': $+10$ on the left, `bias\_hpf': $+60$ on the right. Notice that while the magnitude changes, the location of the peak stays nearly the same. The uncertainty bands were smoothed using a Savitzky-Golay filter (window sizes 5 and 2 for the $+10$ and $+60$ cases, respectively, order 2).}
  \label{fig: noise prob hpf}
\end{figure}
 
\begin{figure}[ht]
  \centering
  \begin{minipage}[b]{0.49\textwidth}
    \centering
    \includegraphics[width=\textwidth]{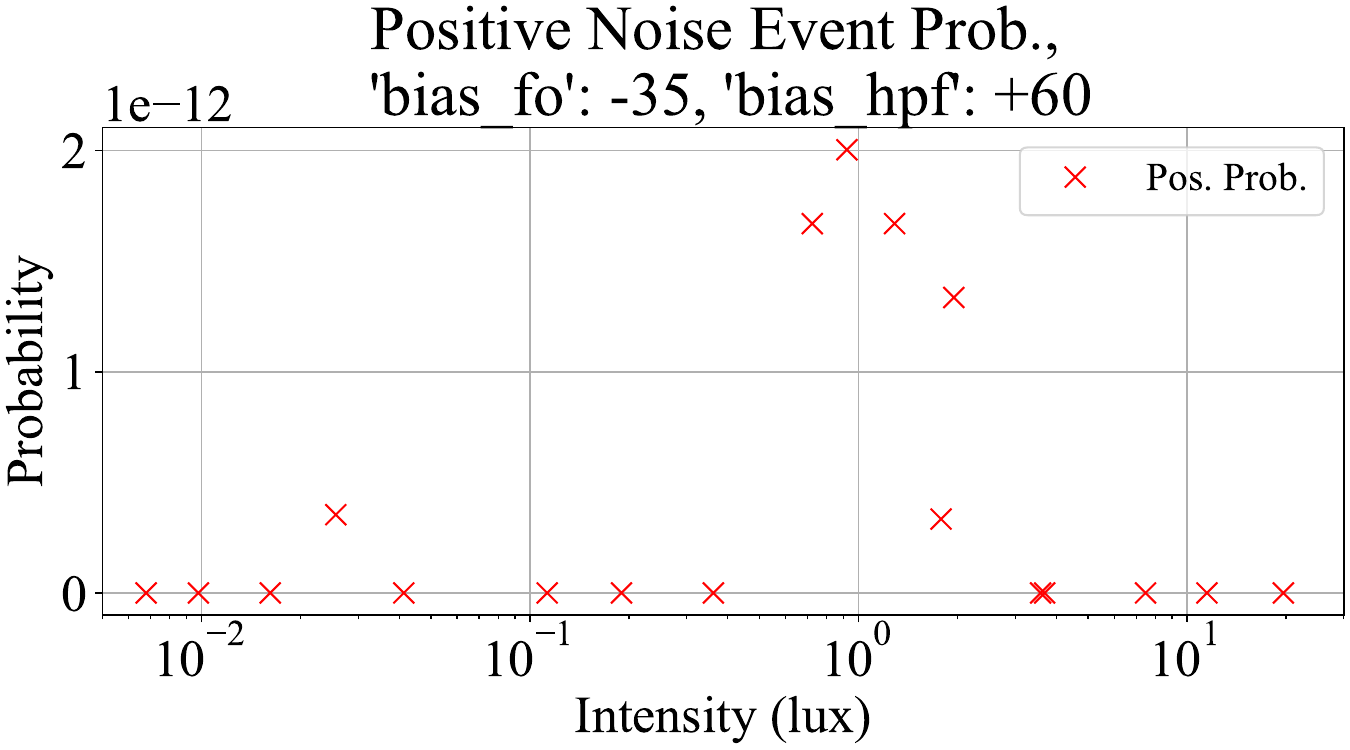}
  \end{minipage}
  \begin{minipage}[b]{0.49\textwidth}
    \centering
    \includegraphics[width=\textwidth]{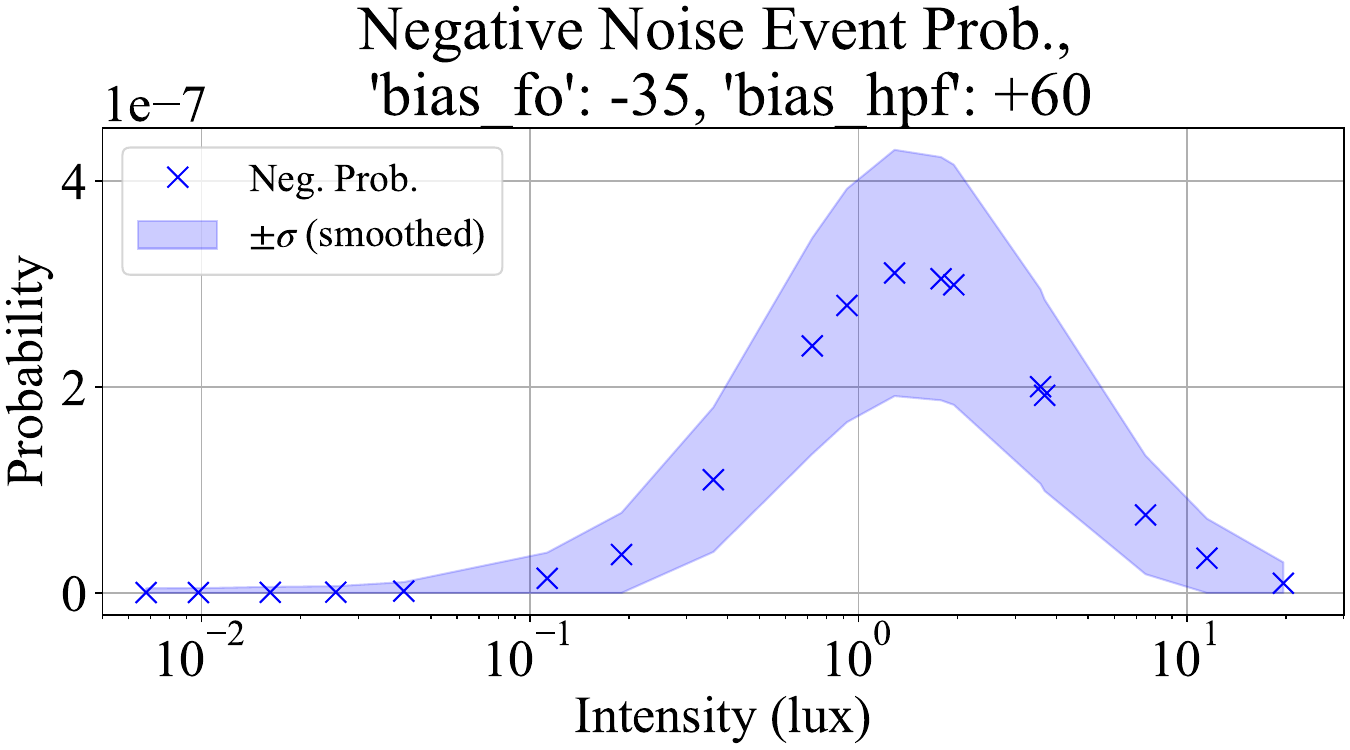}
  \end{minipage}
 
  \caption{Observed $\hat{P}_\pm$ for `bias\_fo': $-35$ and `bias\_hpf': $+120$. The uncertainty band is omitted for the positive events, as the magnitude of the band eclipses the data points themselves. The negative event uncertainty band was smoothed using a Savitzky-Golay filter (window size 2, order 2).}
  \label{fig: fo -35 hpf +120}
\end{figure}
 
The last bias setting we consider is `bias\_refr', which controls the length of the refractory period of the pixels. Decreasing `bias\_refr' increases the refractory period, while increasing `bias\_refr' decreases it. Observations were made across the range of `bias\_refr' from $-20$ to $+200$. For the most part, non-default `bias\_refr' settings left $\hat{P}_\pm$ unchanged from the default case, though the event rates themselves changed slightly. The refractory times for different `bias\_refr' settings can be seen in Table~\ref{tab: impact of bias_refr}. The relationship between them suggests a reciprocal form, and so we further fit the following expression for the refractory time, 
\begin{equation}
    R=\frac{1530.72}{b_r+22.97}+12.45,
\end{equation}
where $b_r$ is the `bias\_refr' setting value, which can be seen in Fig.~\ref{fig: refr time fit}.
 
\begin{table}[ht]
\centering
\begin{tabular}{|clllllll|}
\hline
\multicolumn{8}{|c|}{Impact of bias\_refr} \\[1mm]  \hline
\multicolumn{1}{|c|}{bias\_refr} & \multicolumn{1}{l|}{-20} & \multicolumn{1}{l|}{-15} & \multicolumn{1}{l|}{-5} & \multicolumn{1}{l|}{0} & \multicolumn{1}{l|}{50} & \multicolumn{1}{l|}{100} & 200 \\[1mm] \hline
 
\multicolumn{1}{|l|}{refractory time ($\mu$s)} & \multicolumn{1}{l|}{528} & \multicolumn{1}{l|}{205} & \multicolumn{1}{l|}{97} & \multicolumn{1}{l|}{79} & \multicolumn{1}{l|}{33} & \multicolumn{1}{l|}{25} & 20 \\[1mm]  \hline
\end{tabular}
\caption{The refractory time resulting from different `bias\_refr' settings.}
\label{tab: impact of bias_refr}
\end{table}
 
\begin{figure}[H]
    \centering
    \includegraphics[width=0.6\linewidth]{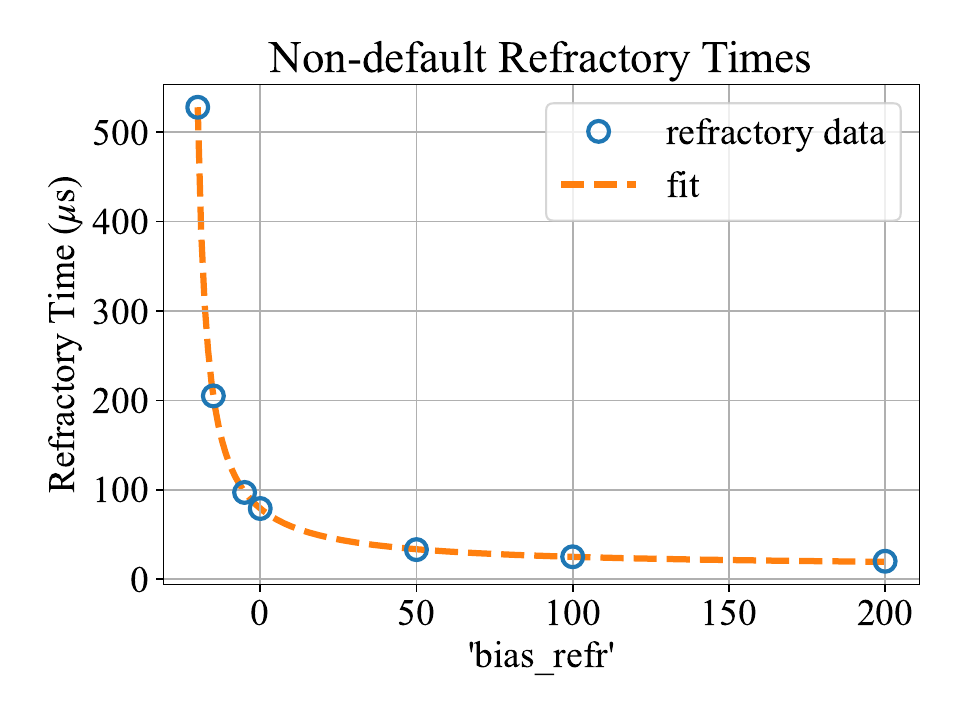}
    \caption{Refractory times at different values of the `bias\_refr' setting, along with a reciprocal-form fit.}
    \label{fig: refr time fit}
\end{figure}
 
\subsection{Parameter Determination} \label{subsection: impact of bias settings}
\textbf{Log Contrast Threshold, $B$.}
Having established the success of parameter determination for the default bias case, we can also consider the non-default cases. We make the assumption that the bias settings are independent of each other. Inspired by the observation that the HPF- and LPF-related bias settings can have a stronger impact on the location of the distribution peak, whereas `bias\_diff\_on/off' more significantly impacts the distribution magnitude, we assume that $\theta$ is influenced by `bias\_fo' and `bias\_hpf', while $B$ is primarily influenced by `bias\_diff\_on/off'. Given this, we can establish how `bias\_diff\_on/off' changes $B$ by fitting it to the observed noise for the non-default settings, while keeping the parameters of $\theta$ fixed. 
\par
The results of this approach can be seen in Fig.~\ref{fig:linear fit to B}. The reported standard deviations are calculated from the set of numerically found $B$ values, for each of the individual data points, at a given bias setting. However, for the `bias\_diff\_on/off': $+105$ case, given that we observed no noise events, we were unable to directly fit $B$, and so instead estimated it by using the smallest $B$ value that yields a predicted probability low enough that the expected event counts, for the recording lengths used, are less than 1. The ``standard deviation'' of this case is included for illustrative purposes. 
\par
\begin{figure}[h!]
    \centering
    \includegraphics[width=0.7\linewidth]{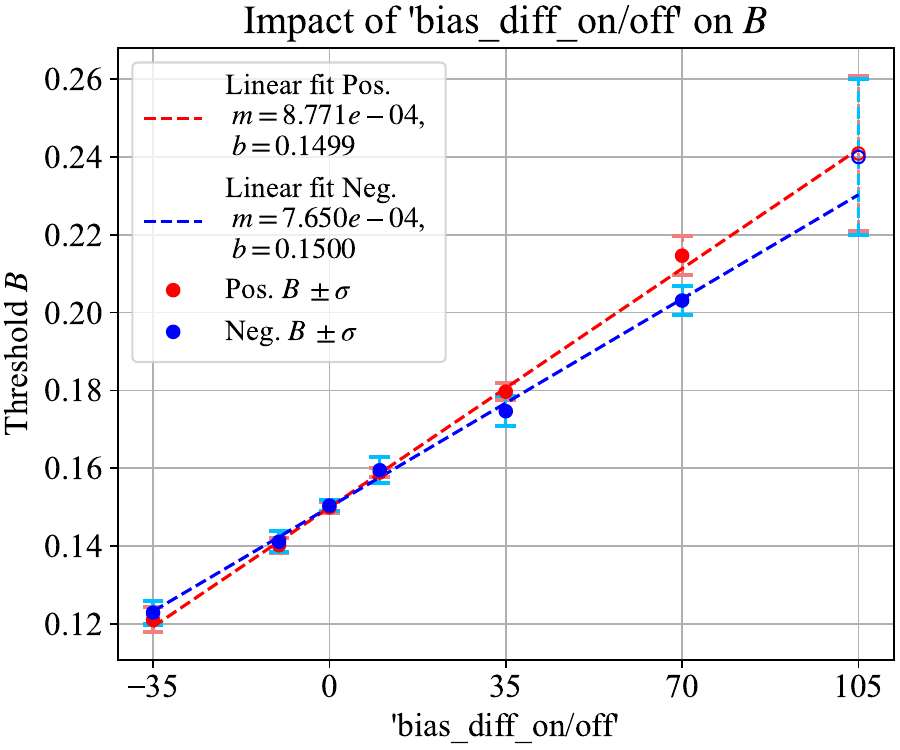}
    \caption{Linear fit to $B$ as a function of `bias\_diff\_on/off'. While both $\pm B$ exhibit linear behavior, they slightly diverge toward the extremes of the setting range. For `bias\_diff\_on/off': $+105$, no noise events were recorded, and $B$ in that case is estimated by the minimal value of $B$ for which the expected noise event count is $<1$ over the full length of the recording. The error bars for the $+105$ points are included solely for illustrative purposes.}
    \label{fig:linear fit to B}
\end{figure}
 
As can be seen, there is a strong linear relationship between $B$ and `bias\_diff\_on/off'. We have included separate linear fits for $\pm B$, as it appears that there is slight divergence between them at the extremes of the setting range, which can be impactful given that the effect of $B$ on the noise events is exponential. However, if one wishes to approximate them as equal, we can average the fits, which gives
\begin{equation}
B=(8.21\cdot10^{-4})k+0.15,
\end{equation}
where $k$ is the value of `bias\_diff\_on/off'. 
\par
Despite the strong linearity, we do wish to add a note of caution regarding the accuracy of the fits of $B$ at different bias settings. To measure the goodness of fit for the $B$ values, we computed two metrics: the reduced chi-square statistic, $\chi^2_\nu$, and the peak-relative root mean square error (Peak RRMSE), which is the standard RMSE value divided by the magnitude of the peak of the probability distribution. These metrics, as a function of the bias setting value, can be seen in Fig.~\ref{fig:metric of B fits}.
\par
As a reminder, the interpretation of $\chi^2_\nu$ is that if $\chi^2_\nu\approx1$, then the alignment of the fit and the data is in accordance with the variance of the error. If $\chi^2_\nu \ll1$, the fit still aligns with the data, but the error of the fit is much less than what the uncertainty of the data points themselves would indicate, possibly indicating that overfitting has occurred. If $\chi^2_\nu\gg1$, the fit is poor and the error of the fit is much greater than the data errors would indicate, suggesting that underfitting may have occurred. Regarding Peak RRMSE, the interpretation is that if the metric is close to 0, then the error of the fit is minimal compared to the magnitude of the data points, indicating, at an order-of-magnitude level, a more reliable fit; whereas if the Peak RRMSE $\gtrsim1$, then the error of the fit is comparable to the magnitude of the data, and so the fit captures the trend no better than a one-sigma scatter. 
\par
As seen in Fig.~\ref{fig:metric of B fits}, both $\chi^2_\nu$ and the Peak RRMSE take values that are indicative of more ideal fits for values of `bias\_diff\_on/off' that are near the default value of 0, whereas at the extremes of the setting range, the quality of the fit of $B$ values may decrease. 
Based upon this, while the linearity of $B$ in relation to `bias\_diff\_on/off' is clearly evident, we advise that the validity of our specific linear fits is strongest near the default `bias\_diff\_on/off' values.
 
\begin{figure}[h!]
    \centering
    \includegraphics[width=0.8\linewidth]{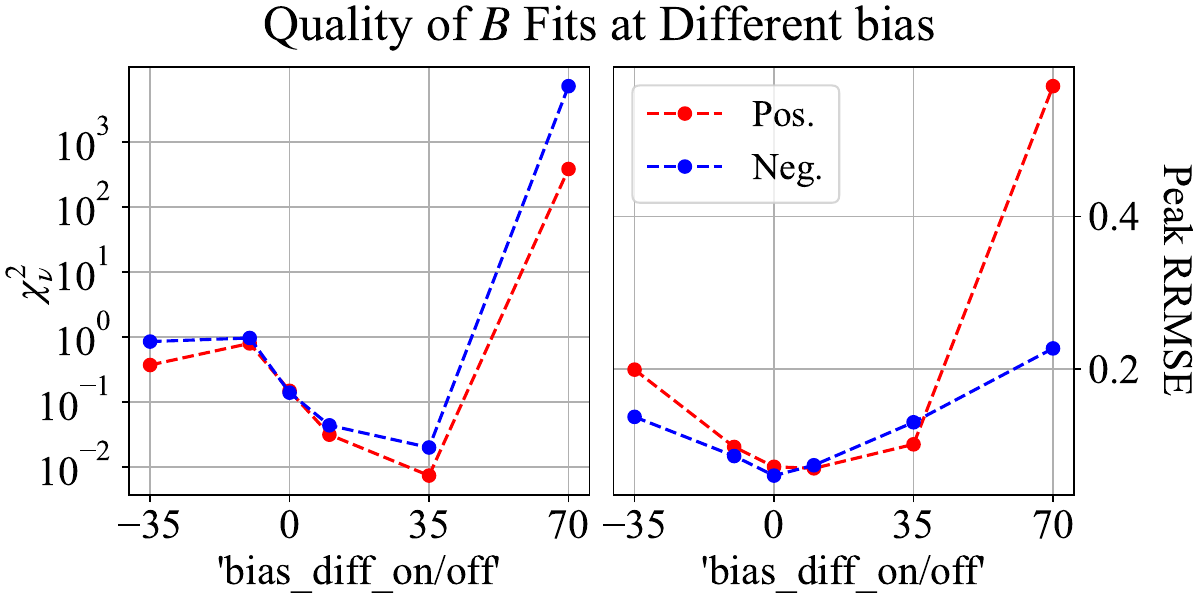}
    \caption{$\chi^2_\nu$ and Peak RRMSE goodness-of-fit metrics for our fits of $B$ at the different `bias\_diff\_on/off' values used. Ideal values for the metrics are $\chi^2_\nu\lesssim1$ and $\text{Peak RRMSE} \approx 0$, indicating that the quality of the fits is greatest for `bias\_diff\_on/off' values around the default value of 0.}
    \label{fig:metric of B fits}
\end{figure}
 
\textbf{Leakage Term, $\theta$.}
Moving to $\theta$ and its coefficients $c_1$, $c_2$, and $c_3$, we postulated that $\theta$ was influenced by `bias\_fo' and `bias\_hpf' primarily, and so by holding $B$ fixed, we can fit these parameters to the observations at different setting values to establish their dependency. However, here the relationship appears far less straightforward than $B$'s dependency on `bias\_diff\_on/off'; see Fig.~\ref{fig:impact of bias settings on theta parameters}. We can observe a few rough trends, however.
\par
 
\begin{figure}[h!]
    \centering
    
    \begin{tikzpicture}[scale=0.5]
    \node[inner sep=00pt] (far) at (4,4)
        {\includegraphics[width=1\textwidth]{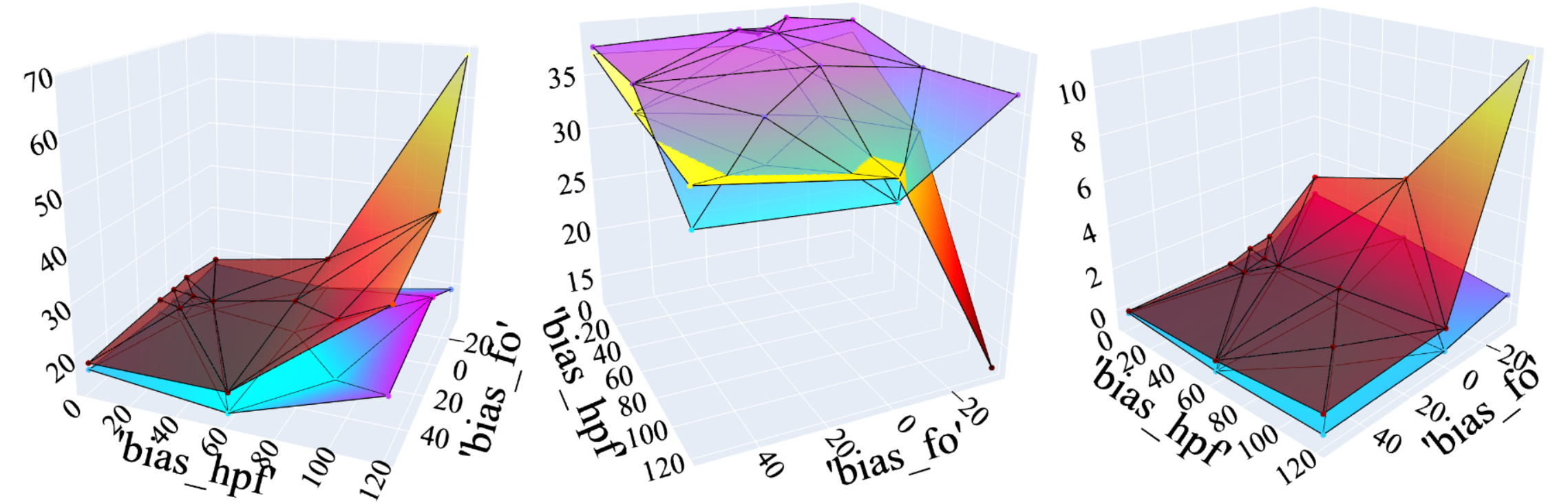}};
    \node[] at (-7,10) {\Large{$c_1$}};
    \node[] at (4,10) {\Large{$c_2$}};
    \node[] at (14,10) {\Large{$c_3$}};
    \node[] at (4,12) {\Large{Impact of Bias Settings on $\theta$ Coefficients}};
    
    \filldraw[fill=white, draw=gray] (-4.5,7.5) rectangle (-0.4,11);
    \shade[left color=cyan,right color=magenta] (-4.25,7.8) rectangle (-2.5,9);
    \shade[left color=red,right color=yellow] (-4.25,9.5) rectangle (-2.5,10.7);
    \node[] at (-1.5,10.1) {Pos.};
    \node[] at (-1.5,8.3) {Neg.};
    \end{tikzpicture}
    
    \caption{Dependency of $c_1$, $c_2$, and $c_3$, the coefficients of $\theta(\lambda)$, on the bias settings `bias\_fo' and `bias\_hpf'. The red-orange surfaces correspond to the values for the positive event case; the blue-purple surfaces correspond to the negative event case. The surface ``on top'' is partially translucent.}
    \label{fig:impact of bias settings on theta parameters}
\end{figure}
 
$c_1$ appears to be primarily a function of `bias\_hpf' and fairly independent of `bias\_fo', with the dependence on `bias\_hpf' appearing convex parabolic. Aside from anomalous behavior at maximum `bias\_hpf' and minimum `bias\_fo', $c_2$ also appears to be primarily a function of `bias\_hpf' and fairly independent of `bias\_fo', with the dependence being a slight decrease as `bias\_hpf' increases. $c_3$ has a similar overall shape as $c_1$, though of different magnitude and with the dependency swapped: $c_3$ appears primarily a function of `bias\_fo' and fairly independent of `bias\_hpf'. We caution that, due to the relatively sparse sampling of the `bias\_fo' and `bias\_hpf' parameter space, the actual trends may be obfuscated. 
\par
Unlike the case of $B$, we also do not see any consistent decrease in the goodness of fit for the coefficients of $\theta$ as we move further from the default values. Without any concrete analytical relationship between `bias\_hpf', `bias\_fo', and the coefficients of $\theta$, we simply provide the fitted values for the coefficients and $c_V$ across all measured bias setting combinations in Table~\ref{tab:theta_coefficients_nondefault}. The default bias setting case (first row) corresponds to the values reported in the main text (Table~\ref{m-table:parameter_values}).
\par
\textbf{Conversion Factor, $\alpha$.}
We did not find any compelling evidence that $\alpha$ was influenced by the bias settings, and a value of $\alpha=4.5$ fitted well in all cases.
\par
\textbf{Vertical Offset, $c_V$.} 
The ``noise floor'' vertical offset is accounted for by the addition of $c_V$, the value of which is chosen as the value of the lowest probability data point (across the intensity range). If the inclusion of the $c_V$ term does not increase the accuracy of the fit, its value is simply left at 0, which is also the case if there is a probability data point with a value of 0. 
\par
When nonzero, $c_V$ is generally larger for negative noise events than for positive ones. Given that $c_V$ appears broadly proportional to overall event counts, this is consistent with the tendency for negative noise event counts to be higher than positive ones. Conditions in which there are more noise events overall are associated with a larger $c_V$ value, such as a more sensitive $B$. This relationship between $c_V$ and `bias\_diff\_on/off' (which controls the sensitivity of $B$) can be seen in Fig.~\ref{fig:c_v vs bias_diff_on/off}, and appears to exhibit exponential decay.
\par
The relationship between $c_V$ and the `bias\_hpf' and `bias\_fo' settings, which can be seen in Fig.~\ref{fig:c_v verses hpf and fo}, is less straightforward. While $c_V$ appears to be weakly affected by `bias\_fo' (slightly decreasing with more negative settings), it is much more strongly impacted by `bias\_hpf', though in different ways for positive and negative events. In the positive case, a lower `bias\_hpf' (closer to the default value) is associated with a higher $c_V$, while in the negative event case, $c_V$ peaks in the middle of the `bias\_hpf' range, with lower values at the extremes of the setting. However, as with the $\theta$ coefficients, the relatively sparse sampling of the parameter space may be obfuscating trends.

\begin{table}
    \centering
    \small
    \caption{Fitted values of the $\theta(\lambda)$ coefficients ($c_1$, $c_2$, $c_3$) and vertical offset $c_V$ for different combinations of `bias\_fo' and `bias\_hpf', with `bias\_diff\_on/off'$\,=0$ (i.e., $B=0.15$). Subscripts $+$ and $-$ denote positive and negative event polarities, respectively. A $c_V$ value of 0 indicates that the offset term did not improve the fit and was omitted.}
    \label{tab:theta_coefficients_nondefault}
     
    \setlength{\tabcolsep}{3.5pt}
    \renewcommand{\arraystretch}{1.15}
     
    \begin{tabular}{rr|rrrr|rrrr}
    \hline
    \multicolumn{2}{c|}{\textbf{Bias Settings}} & \multicolumn{4}{c|}{\textbf{Positive Events}} & \multicolumn{4}{c}{\textbf{Negative Events}} \\
    \texttt{fo} & \texttt{hpf} & $c_{1+}$ & $c_{2+}$ & $c_{3+}$ & $c_{V+}$ & $c_{1-}$ & $c_{2-}$ & $c_{3-}$ & $c_{V-}$ \\
    \hline
      0 &   0 & 18.92 & 35.49 &  0.439 & $9.57\!\times\!10^{-9}$  & 16.42 & 37.42 &  0.068 & $3.18\!\times\!10^{-8}$ \\
    $-10$ &   0 & 19.31 & 35.57 &  0.646 & 0                       & 15.85 & 38.88 &  0.082 & 0 \\
     10 &   0 & 19.05 & 36.54 &  0.078 & $2.3\!\times\!10^{-8}$  & 16.04 & 38.03 & $-0.124$ & $4.9\!\times\!10^{-8}$ \\
      0 &  10 & 18.17 & 36.44 &  0.245 & 0                       & 16.02 & 38.45 & $-0.113$ & 0 \\
      0 &  20 & 17.81 & 36.38 &  0.227 & 0                       & 15.85 & 38.37 & $-0.145$ & 0 \\
     10 &  10 & 18.01 & 37.01 & $-0.012$ & 0                     & 15.25 & 38.52 & $-0.234$ & 0 \\
      0 &  60 & 20.90 & 32.78 &  0.454 & 0                       & 14.03 & 37.08 & $-0.331$ & $1.4\!\times\!10^{-7}$ \\
     55 &   0 & 18.20 & 36.95 & $-0.061$ & $1.7\!\times\!10^{-8}$ & 16.88 & 37.61 & $-0.172$ & $3.6\!\times\!10^{-8}$ \\
     55 &  60 & 18.21 & 34.89 &  0.054 & 0                       & 14.23 & 37.18 & $-0.452$ & 0 \\
     55 & 120 & 39.02 & 33.29 &  0.516 & $5.8\!\times\!10^{-11}$ & 23.55 & 30.19 & $-0.377$ & $1.9\!\times\!10^{-9}$ \\
     27 &  90 & 26.62 & 31.49 &  0.299 & $8.4\!\times\!10^{-11}$ & 14.53 & 35.28 & $-0.602$ & $8.5\!\times\!10^{-9}$ \\
    $-35$ &   0 & 17.86 & 36.92 &  2.889 & 0                     & 16.65 & 38.12 &  1.967 & 0 \\
    $-35$ &  60 & 22.06 & 30.18 &  4.400 & 0                     & 15.10 & 35.97 &  1.378 & 0 \\
    $-35$ & 120 & 68.50 & 12.90 & 11.103 & 0                     & 19.31 & 36.56 &  0.549 & $3.5\!\times\!10^{-10}$ \\
      0 & 120 & 43.34 & 31.42 &  0.849 & $8.1\!\times\!10^{-12}$ & 26.31 & 29.53 & $-0.227$ & $6.6\!\times\!10^{-10}$ \\
    \hline
    \end{tabular}
\end{table}

\begin{figure}
    \centering
    \includegraphics[width=0.6\linewidth]{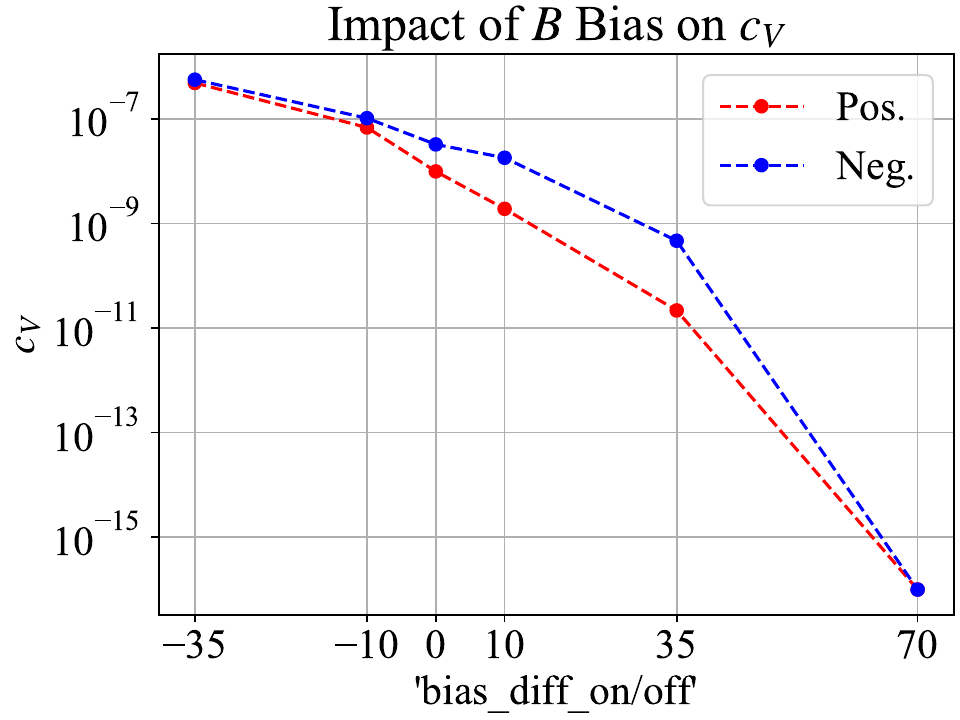}
    \caption{Relationship between $c_V$ and `bias\_diff\_on/off' setting, which most directly affects $B$ (all other bias settings are at default values here). Past `bias\_diff\_on/off' of $+35$ and $+70$, for positive and negative cases, respectively, $c_V$ takes a value of 0 (and so is not shown), as the minimum number of noise events in those cases is 0.}
    \label{fig:c_v vs bias_diff_on/off}
\end{figure}
 
\begin{figure}
    \centering
    \begin{tikzpicture}[scale=0.4]
    
    \node[inner sep=00pt] (far) at (4,4)
        {\includegraphics[width=1\textwidth]{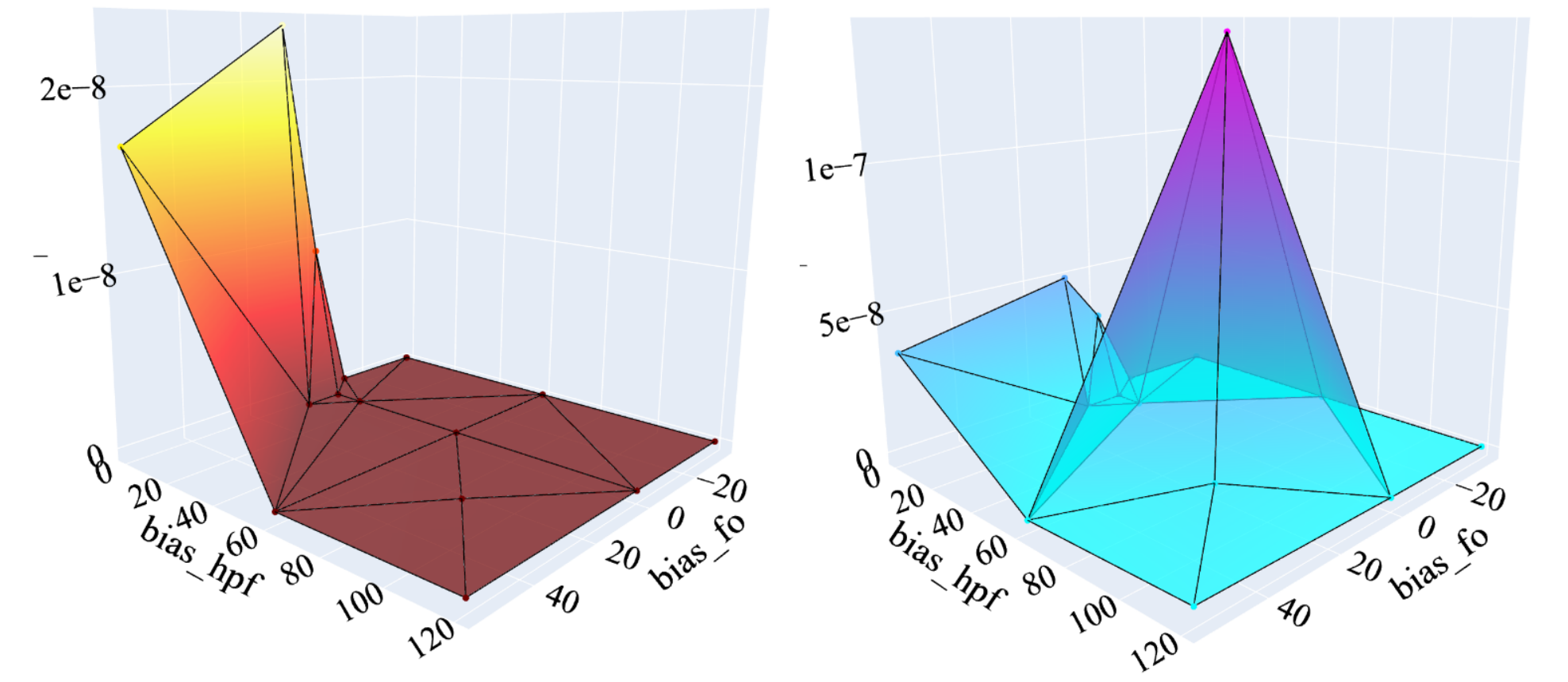}};
    \node[] at (4,13) {\Large Impact of Bias Settings on $c_V$};
    \filldraw[fill=white, draw=gray] (17,5) rectangle (22,9);
    \filldraw[fill=white, draw=white] (4,5) rectangle (5,6);
    \filldraw[fill=white, draw=white] (-12,5) rectangle (-11.5,6);
    \node[] at (-12,6) {\Large $c_V$};
    \node[] at (4,6) {\Large $c_V$};
    \shade[left color=cyan,right color=magenta] (17.5,5.5) rectangle (19.5,6.75);
    \shade[left color=red,right color=yellow] (17.5,7.25) rectangle (19.5,8.5);
    \node[] at (20.75,7.8) {Pos.};
    \node[] at (20.75,6) {Neg.};
    
    \end{tikzpicture}
    \caption{Dependency of the vertical offset term $c_V$ on the bias settings `bias\_fo' and `bias\_hpf'. For the most part, outside of values near the default bias settings (0 for both `bias\_fo' and `bias\_hpf'), $c_V$ is taken to be 0 (due to not improving the parameter fitting).}
    \label{fig:c_v verses hpf and fo}
\end{figure}

\section{Uncertainty vs Bin Size}
\label{section: B sigma vs bin size}
See Fig.~\ref{fig:elbow method}.

\begin{figure}[!ht]
    \centering
    \includegraphics[width=0.5\linewidth]{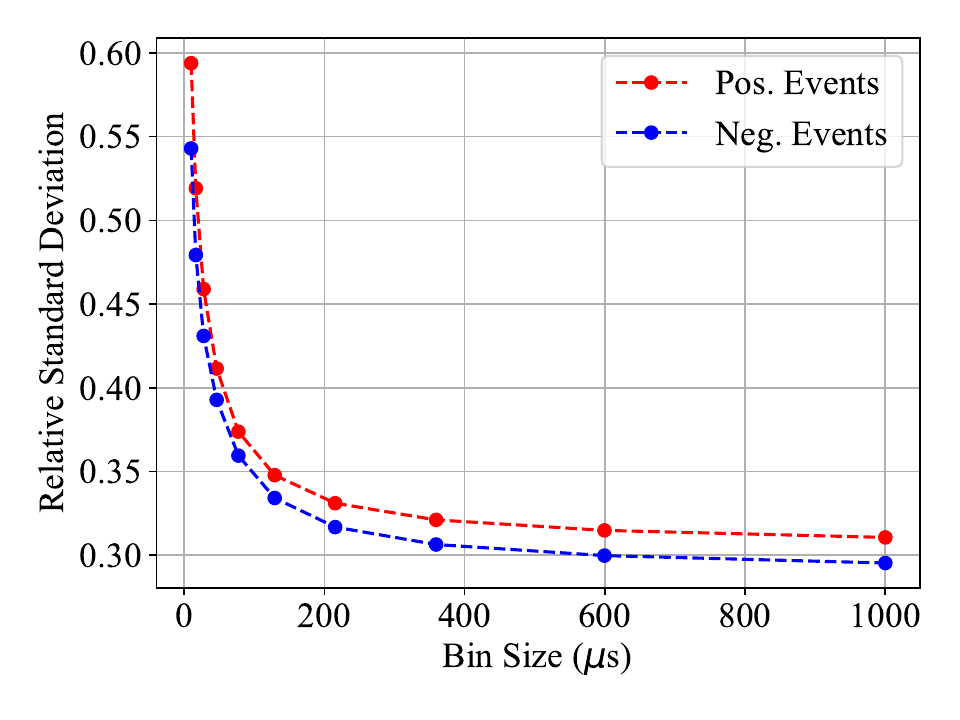}
    \caption{Relative standard deviations of $\hat{P}_\pm$ resulting from binning events over time into bins of different sizes, for recording at default biases near the peak of $\hat{P}_\pm$. The elbow point occurs at a bin size of $\approx100 \;\mu$s.}
    \label{fig:elbow method}
\end{figure}

\section{Comparison of Light Sources} \label{section: light sources}
See Fig.~\ref{fig:compar light sources}.

\begin{figure}[!ht]
    \centering
    \includegraphics[width=0.5\linewidth]{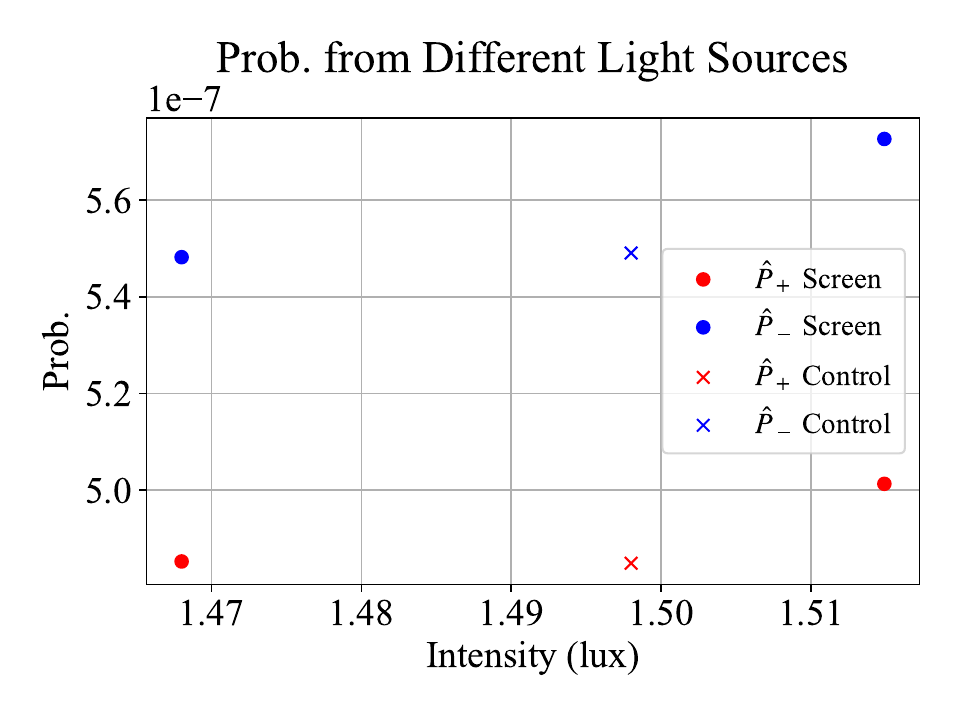}
    \caption{Noise event probabilities from the screen (LED monitor) and control (incandescent bulb) light sources. At similar intensities, the noise event probabilities are quite similar. This indicates that the spectral differences of the sources has a negligible impact of their noise event rates, and so results obtained from them are suitable for comparison with each other. }
    \label{fig:compar light sources}
\end{figure}

\section{Radiometric Cross-check of Lux-Photon Conversion Factor}\label{section: alpha crosscheck}
To cross-check our fitted value of $\alpha$, we can estimate it from basic radiometry. A single EC pixel has area
\begin{equation}
A_{\mathrm{px}}=(4.86~\mu\mathrm{m})^2
=\bigl(4.86\times 10^{-6}\ \mathrm{m}\bigr)^2
=2.36196\times 10^{-11}\ \mathrm{m}^2.
\end{equation}
If the scene illuminance is $I$ (in lux $=\mathrm{lm}\ \mathrm{m}^{-2}$), then the luminous flux passing through one pixel is
\begin{equation}
\Phi_v = I\,A_{\mathrm{px}}\quad [\mathrm{lm}].
\end{equation}
Approximating our incandescent source's (photopic) luminous efficacy by
\begin{equation}
K_v \approx 20\ \mathrm{lm}\ \mathrm{W}^{-1},
\end{equation}
which is in the upper range for modern incandescent bulbs \citep{Agrawal_Efficiency}, the corresponding radiant flux (radiant power) through the pixel is
\begin{equation}
\Phi_e = \frac{\Phi_v}{K_v}=\frac{I\,A_{\mathrm{px}}}{K_v}\quad [\mathrm{W}].
\end{equation}
Assuming an average wavelength $\lambda = 666~\mathrm{nm}$ for the incandescent source (based on the color temperature, as noted earlier), the mean photon rate incident on one pixel is
\begin{align}
\dot N
&=\frac{\Phi_e}{h c/\lambda} = \frac{I\,A_{\mathrm{px}}}{K_v}\,\frac{\lambda}{h c} \\
&= I\times\Biggl[\frac{(2.36196\times 10^{-11}\ \mathrm{m}^2)\,(666\times 10^{-9}\ \mathrm{m})}{(20\ \mathrm{lm}\ \mathrm{W}^{-1})\,(6.62607015\times 10^{-34}\ \mathrm{J\,s})\,(2.99792458\times 10^{8}\ \mathrm{m\,s^{-1}})}\Biggr] \\
&\approx (3.96\times 10^{6})\, I\quad \mathrm{s}^{-1}.
\end{align}
Therefore, for an exposure (timestep) of duration $\Delta t$,
\begin{equation}
N_{\mathrm{px}} \approx \dot N\,\Delta t \approx \bigl(3.96\, I\bigr)\,\frac{\Delta t}{\mu\mathrm{s}}
\quad\text{photons per pixel},
\end{equation}
so that for $\Delta t=1~\mu\mathrm{s}$ each pixel receives, on average, about $4I$ photons. This agrees well with our independently fitted value $\alpha\simeq 4.5$ (in the same units), lending additional confidence to that estimate.
\par
The conversion from luminous to radiant quantities via a single $K_v$ is an approximation that implicitly folds in the source's spectrum and photopic weighting; optical losses and detector quantum efficiency are also neglected, as we assume that these are implicit in the reported on-pixel $I$. These simplifications are adequate here because we only seek an order-of-magnitude cross-check of $\alpha$.  

\section{Greyscale to Intensity Mapping}
\label{section:greyscale_to_intensity_mapping}
To enable the generation of synthetic noise event images via the various noise event probability distributions, a mapping from the greyscale pixel values to the intensity observed by the EC had to be determined. A series of uniform greyscale images was displayed on the screen (at maximum brightness), with the observed intensity recorded, as reported by the EC's illuminometer. These were then used to determine the synthetic intensity at the $i$-th pixel, $I_i$, as given by the mapping
\begin{equation}
    I_i\approx(2.15\cdot10^{-5})(g_i)^{2.521}+0.15,
\end{equation}
where $g_i$ is the greyscale value (integers 0-255) at the $i$-th pixel.

\section{Outlier Pixels} \label{section: hot pixels}
In an ideal EC, all of the pixels in the sensor array have the same characteristics and exhibit the same behavior. Of course, in real ECs this is not the case, and there exist per-pixel variations in characteristics. It is then desirable to have an EC of sufficient quality that these variations are minimal enough to allow the pixel array to be treated as uniform for many cases. While great progress has been made in EC development in recent years, it is still necessary to ensure that all the pixels in the array are sufficiently well behaved. This involves the identification and filtering of pixels whose behavior differs greatly from the desired case, and whose inclusion in data may significantly skew results. 
\par
Here we take inspiration from and build upon the classification system presented by Wang \textit{et al.} for these outlier pixels  \citep{wang2020eventcameracalibrationperpixel}, and discuss strategies for identifying them. Broadly, outlier pixels can be grouped into four categories, not all of which necessarily require filtering:
\begin{itemize}
    \item \textbf{Hot Pixels} are pixels that report far more events than would be expected (often greater than $20\sigma$ above the mean event counts in our observations). This may be due to having a much lower $B$ value than intended, but can also be due to pixels exhibiting other forms of erroneous behavior, which we discuss further in the following text. These are often desirable to filter.
    \item \textbf{Warm Pixels} are pixels that report slightly more events than would be expected. This is more likely to be the result of pixels with a slightly lower $B$ value than intended. These may not warrant filtering.
    \item \textbf{Cool Pixels} are pixels that report slightly fewer events than would be expected. This is more likely to be the result of pixels with a slightly higher $B$ value than intended. These may not warrant filtering.
    \item \textbf{Cold Pixels} are pixels that report far fewer events than would be expected. Similar to hot pixels, this may be due to pixels having a much higher $B$ than intended, but may also be due to other erroneous behavior. Cold pixels that do not report any events are referred to as \textit{frozen pixels}. These may be desirable to filter.
\end{itemize}
A pixel may exhibit these categories for both positive and negative events, or just one or the other. So a pixel may report expected positive event counts, but could be considered a hot pixel for negative events.
\par
We also observed an additional type of outlier behavior, not necessarily related to any of the above categories, in which a pixel reports more than one event per timestep. As confirmed by our camera's manufacturer, the intended function of correctly operating pixels is that intensity changes should result in a single event at a given timestep  \citep{Prophesee_LargeIntensityResponse}, and so any reports of multiple events in one timestep by a single pixel represent strictly erroneous behavior. We refer to outliers that do not exhibit this behavior, and solely belong to one of the above categories, as \textbf{Type 1} outliers, and ones that do exhibit this behavior as \textbf{Type 2} outliers. 
\par
A hot pixel Type 2 would then be a pixel that reports excessive event counts \textit{and} more than one event per timestep, whereas a hot pixel Type 1 only reports excessive event counts. A Type 2 outlier need not belong to any of the above categories either. If a pixel reports multiple events in a single timestep, but otherwise normal event counts, then it would solely be a Type 2 outlier.
\par
Pixels may not remain within one outlier classification, or may not even always exhibit outlier behavior at all. We observed cases where, when taking a static recording to observe noise events, an otherwise well-behaved pixel would have a massive jump in its reported events, becoming a hot pixel for an extended period of time (typically on the order of seconds), before its reported events would plummet and it would regain normal behavior. 
\par
The shifting of a pixel between normal and outlier behavior, or between outlier classifications, appears to be influenced by the bias settings of the camera. Different combinations of `bias\_diff\_on/off', `bias\_fo', `bias\_hpf', and `bias\_refr' result in different pixels exhibiting outlier behaviors. Precise relationships between the bias settings and occurrences of pixels exhibiting outlier behaviors are outside the scope of this work; however, we were able to observe some general trends. Bias setting values that resulted in greater overall event counts were often associated with greater numbers of pixels becoming hot pixels (such as with `bias\_diff\_on/off': $-10$), whereas setting values that resulted in a divergence between positive and negative events were associated with greater occurrences of pixels exhibiting Type 2 behavior for whichever polarity was more numerous (such as with `bias\_fo': $+50$ and `bias\_hpf': $+120$). 
\par
In our experience, hot pixels are the most commonly encountered form of outliers that have an outsized impact on analyses of event data (though, as with the above outlier classification characteristics, this may differ between ECs, models, and manufacturers). As such, much of our efforts to detect and filter outlier pixels were dedicated to hot pixels. 
\par
Here we adopt the following notation. Let $\ell$ denote the frequency of consecutive runs of length $m$ of events of a given polarity at a given pixel. If a given pixel reports the following event polarities (0 representing no events at that timestep) over six timesteps: $+1, +1, 0, 0, -1, 0$; this would have a positive $m=2$ run of frequency $\ell=1$ and a negative run of $m=1, \ell=1$. If a pixel reported over six timesteps: $+1, +1, 0, 0, +1, +1$, this would have a positive run of $m=2, \ell=2$ and no negative event runs. The case of a pixel reporting multiple events in a single timestep contributes those events to the current run, so if a pixel reports, over four timesteps: $0, +1, +1\!+\!1, 0$, this would be counted as a positive run of $m=3, \ell=1$.
\par
The following are strategies that can be employed to test for the presence of and identify outlier pixels:
 
\begin{itemize}
\item \textbf{Multiple Events per Timestep.} \textit{Type 2 pixels.} Check if any pixels report any $m\ge2$ cases. These represent deviations from intended pixel behavior via two different phenomena: 1.\ After an event detection at a given timestep, a pixel should be blind for the refractory period; if the pixel reports an event at the next consecutive timestep, then it is not observing the refractory period and is acting erroneously. 2.\ If a pixel reports multiple events per timestep, this is erroneous behavior, as it would not be physically allowed under correct pixel operation. Label any pixels that report $m\ge2$ cases as outliers. 
 
\item\textbf{Excessive Counts.} \textit{Hot pixels.} Compare the frequency $\ell$ of $m=1$ length runs per pixel (effectively, the total event counts excluding cases of $m\ge2$), for positive and negative events separately. Constructing a histogram may be useful, such as in Fig.~\ref{fig:frequency histogram}. Label any pixels that report ``excessive'' frequencies as outliers. The threshold as to what is considered excessive is situationally dependent, but we found that a threshold of $20\sigma$ greater than the mean was often effective.
 
\begin{figure}[h]
    \centering
    \includegraphics[width=0.49\linewidth]{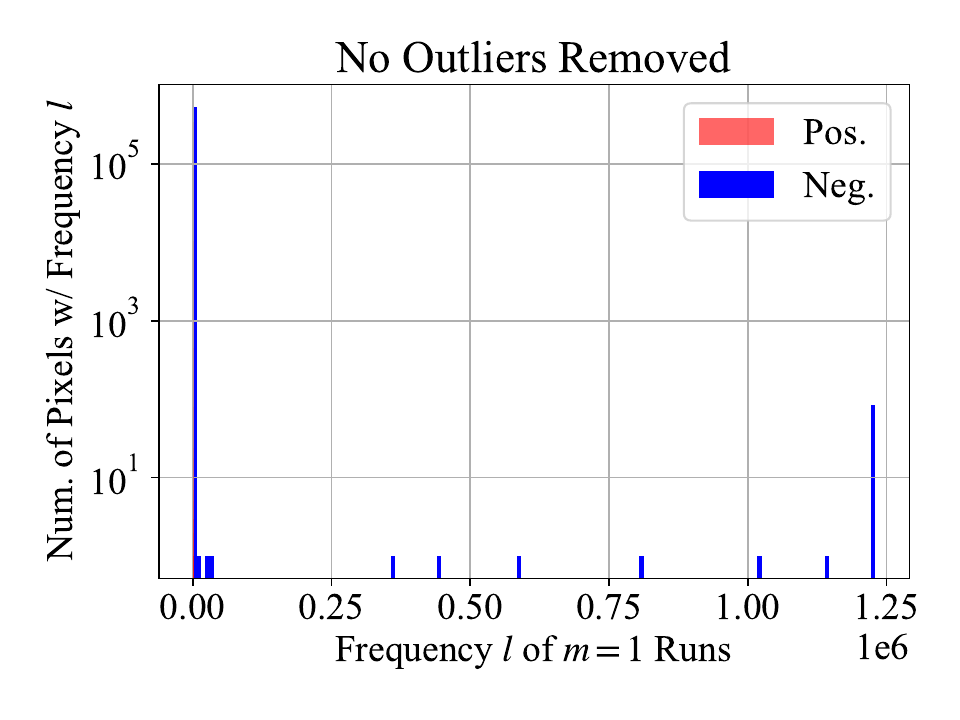}
    \includegraphics[width=0.49\linewidth]{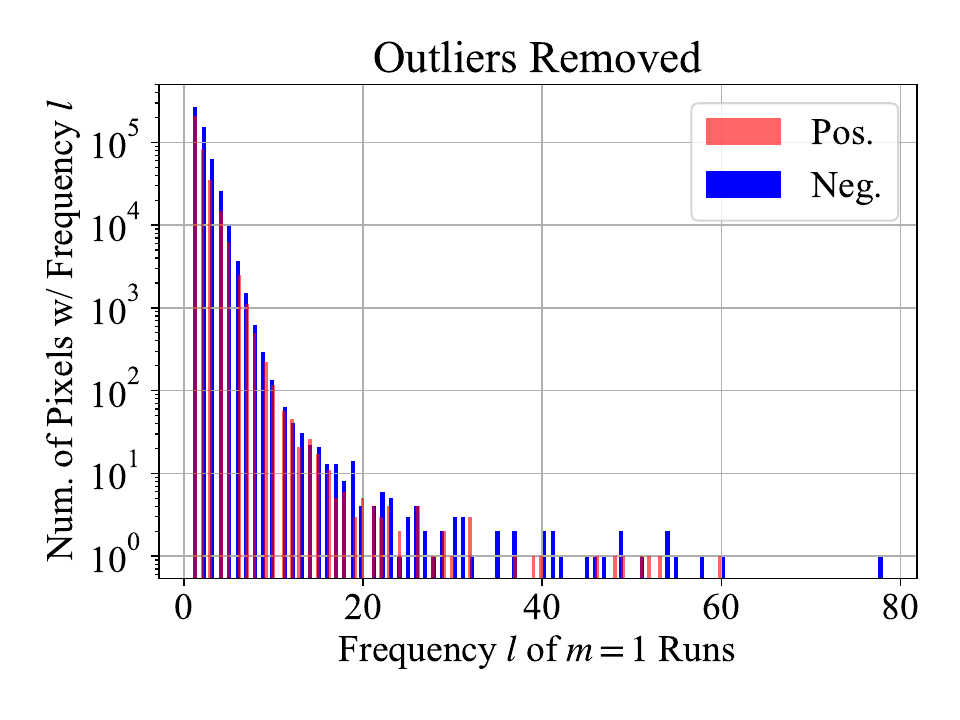}
    \caption{Histograms showing the number of pixels that have a given frequency (vertical axis, log scale) vs the frequency $\ell$ of $m=1$ runs (horizontal axis) for a particular recording. The left histogram is without any filtering of outliers. The large gaps in the horizontal density, at low numbers of pixels, are indicative of the presence of a small number of pixels that have $\ell$ values far greater than the majority of the array. On the right, such a histogram for the same recording, but with the outliers removed. The lack of excessively large horizontal gaps and a gradual descent of the pixel numbers vs frequency is indicative that the included pixels are generally well behaved and outliers reporting excessive counts have been removed.}
    \label{fig:frequency histogram}
\end{figure}
 
\item\textbf{Repeated Bright Spots.} \textit{Hot pixels.} Generate noise event accumulation images of uniform static scenes from multiple different recordings. If there are pixels that repeatedly result in bright spots, that is, a high contrast to the other pixels, as in Fig.~\ref{fig: noise accumulation bright spots}, label these pixels as hot pixels. 
 
\begin{figure}[h]
    \centering
    \newlength{\imgw}
    \setlength{\imgw}{0.475\textwidth} 
    
    \newcommand{\drawAnnotatedImage}[3]{
        \node[inner sep=0pt] (#1) at (#2)
            {\includegraphics[width=\imgw]{#3}};
        \draw[red, thick, dashed]    ($(#1.center)+(-.04\imgw, -.16\imgw)$) circle[radius=.08\imgw];
        \draw[red, thick, dashed]    ($(#1.center)+( .00\imgw,  .1\imgw)$) circle[radius=.08\imgw];
        \draw[red, thick, dashed]    ($(#1.center)+(-.07\imgw,  .30\imgw)$) circle[radius=.08\imgw];
        \draw[yellow, thick, dashed] ($(#1.center)+( .25\imgw,  .30\imgw)$) circle[radius=.08\imgw];
        \draw[green, thick, dashed]  ($(#1.center)+(-.34\imgw, -.01\imgw)$) circle[radius=.15\imgw];
    }
    
    \begin{tikzpicture}
        \drawAnnotatedImage{L}{0,0}{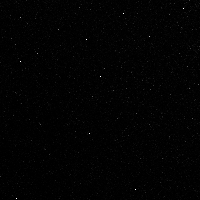}
        \drawAnnotatedImage{R}{1.1\imgw,0}{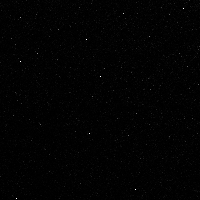}
    \end{tikzpicture}
    \caption{Cropped noise accumulation images for two separate recordings. The dashed red circles highlight some of the pixels that are ``bright'' and have maximal contrast with the majority of pixels in both recordings, indicating that they are hot pixels and warrant filtering. The dashed yellow circle highlights a pixel that has high contrast in the left recording, but is only somewhat bright in the right, indicating a pixel that is a hot pixel some of the time and may warrant filtering. The overall similarity of the noise pattern is a result of the spatial consistency of per-pixel differences in sensitivity. The dashed green circle highlights an area where some differences between the noise accumulation images can be seen, to make it clear that these are separate images despite the overall similarities.}
    \label{fig: noise accumulation bright spots}
\end{figure}
 
\item\textbf{Changes in Event Count Rates.} \textit{Hot and Cold Pixels.} Analyze the event counts over time of a recording of a uniform static scene (the same approach used to observe noise events). It may be useful to plot event counts vs time, such as in Fig.~\ref{fig:events over time}. The event counts should be fairly constant, within a given range. If there are periods where the event count level changes drastically, such as increasing or decreasing to a different level, this is indicative of pixels briefly shifting to hot or cold behavior, depending on the direction of the change. Label the pixels whose event rates changed as outliers of the appropriate category. 
 
\begin{figure}[h!]
    \centering
    \includegraphics[width=0.8\linewidth]{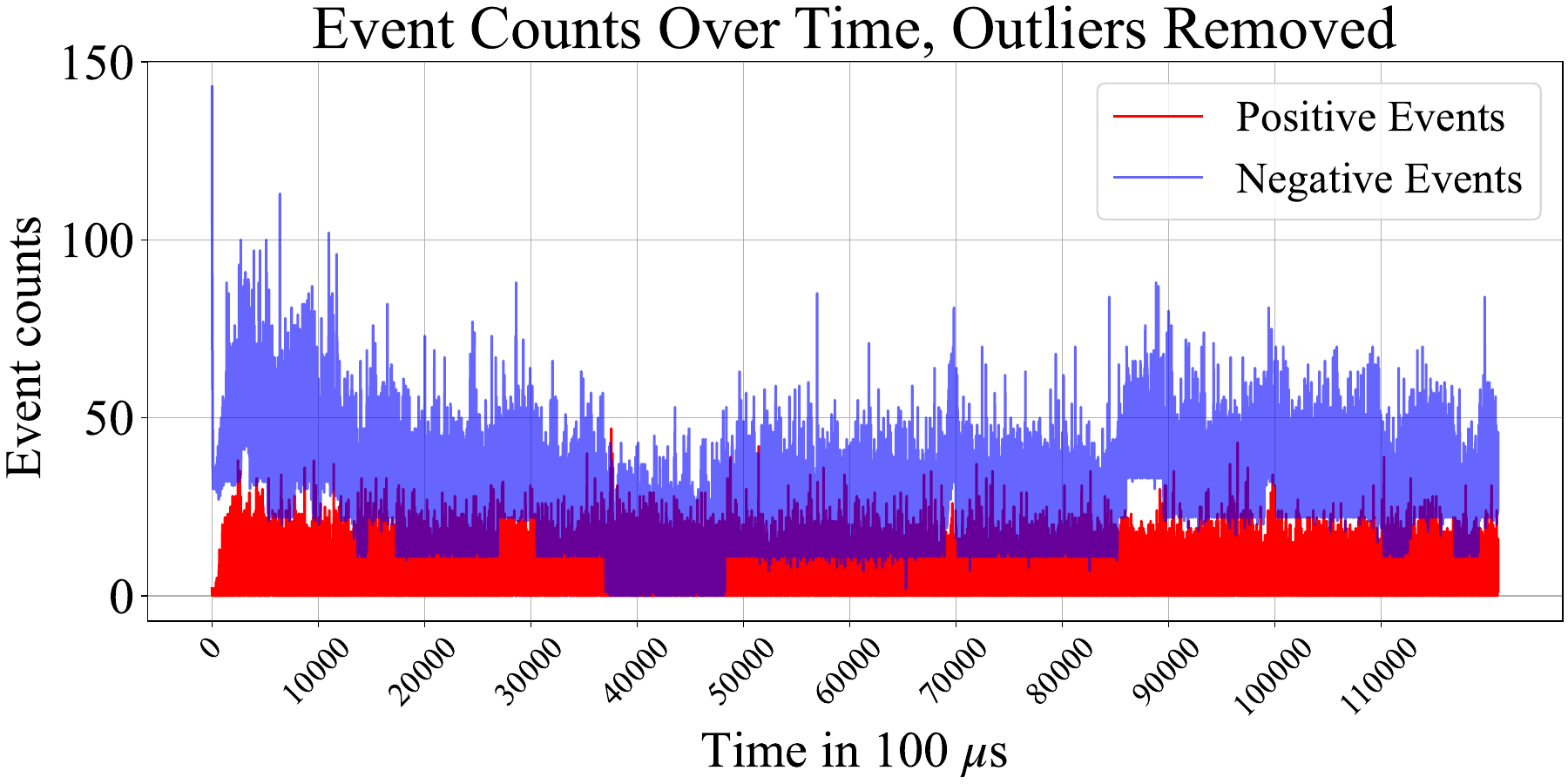}
    \includegraphics[width=0.8\linewidth]{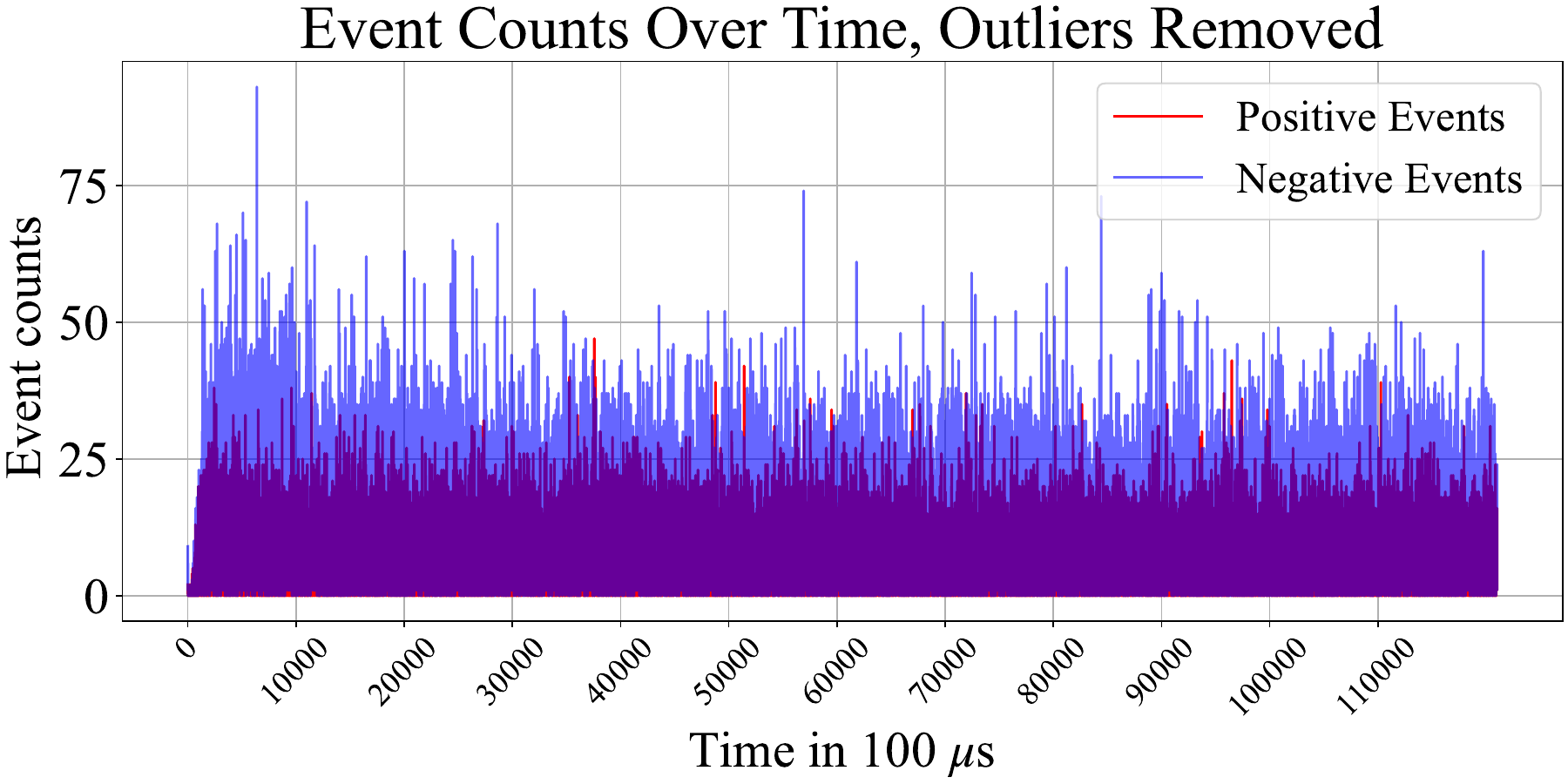}
    \caption{Event counts over time for a specific uniform static scene recording (all events are then noise events), with the events grouped into bins of 100 $\mu$s $=$ 0.0001 s (total recording length $\approx12.1$ s). The top plot shows events over time before all outlier pixels have been removed. The positive noise event counts are stable, whereas the negative noise event counts jump between different levels at different periods of time, indicating that there are pixels shifting in and out of being hot pixels. The bottom plot is for the same recording after the outliers have been removed, which has stabilized the negative noise event levels.}
    \label{fig:events over time}
\end{figure}
 
\item\textbf{Reduced Reaction to Stimuli.} \textit{Cold pixels.} Expose the camera to intensity changes of various significant amplitudes, such as a bright, strobing light. Correctly behaving pixels should respond to the intensity changes. Label pixels that only respond to the largest of intensity changes, well past where most pixels report events, as cold pixels. Label pixels that do not respond to the intensity changes at all and \textit{report no events} as frozen pixels. 
 
\item\textbf{Poisson Outliers.} \textit{Hot pixels.} Given that the underlying mechanism for noise events is a Poisson process, we can look for outliers in noise event data based on the resulting statistical properties. Construct the positive or negative frequency $\ell_{i\pm}$ of $m=1$ length runs for each $i$-th pixel in the sensor array. Let $r_{i\pm}$ denote the Poisson deviance residual, for positive or negative events, at the $i$-th pixel, 
\begin{equation}
    r_{i\pm}=\text{sign} \left(\ell_{i\pm}-\bar{\ell}_\pm \right)\sqrt{2 \left(\ell_{i\pm}\ln\left(\frac{\ell_{i\pm}}{\bar{\ell}_\pm}\right)-\left(\ell_{i\pm}-\bar{\ell}_\pm \right)\right)},
\end{equation}
where $\bar{\ell}_\pm$ is the mean of $\ell_{i\pm}$ across the pixel array, with the convention that $\ell_{i\pm}\ln(\ell_{i\pm}/\bar{\ell}_\pm)=0$ when $\ell_{i\pm}=0$. Under the null hypothesis of no outliers, the deviance residuals are roughly $\mathcal{N}(0,1)$. We use the deviance residual over others, such as the Pearson residual, as it handles small mean values better (as would be the case under very low or very high intensities). 
\par
From here, one can simply pick a residual threshold $\epsilon$, such as, if going with the ``$3\sigma$ rule,'' $\epsilon=3$, and so pixels with $|r_{i\pm}|>3$ are flagged as outliers. However, if we want to be careful about over-filtering (incorrectly flagging too many well-behaved pixels as outliers), we can use an adjusted approach, such as the Bonferroni-adjusted threshold. Let $\hat{a}$ denote the estimated false positive rate ($\hat{a}=0.05$ would denote 5 pixels incorrectly flagged as outliers for every 100 pixels). The Bonferroni-adjusted threshold $\epsilon_B$ is then 
\begin{equation}
    \epsilon_B=\Phi^{-1}\left(1-\frac{\hat{a}}{2M}\right)=S^{-1}\left(\frac{\hat{a}}{2M}\right),
\end{equation}
where $\Phi^{-1}$ is the inverse standard normal CDF, or the quantile function, $S^{-1}$ is the inverse survival function, and $M$ is the number of pixels in the array. 
\par
There is some ambiguity in choosing the correct $\hat{a}$ value, as access to a control group of known outliers and well-behaved pixels is not generally feasible. $\hat{a}$ can be estimated by starting with a ``very small'' value, filtering the flagged outliers, and increasing $\hat{a}$ until any evidence of outlier behavior has been suppressed. In our experience, $\hat{a}\approx0.01$ was often an acceptable rate.
\par
We note that this approach is not valid for cases where no noise events are recorded.
\end{itemize}
\par
It may not be necessary to utilize all of these strategies at the same time. In some cases, utilizing Excessive Counts was sufficient to capture all the outliers; in others, all strategies had to be employed.

\section{Reconstruction from Noise Images} 
\label{sec:recon_from_noise_images}

\subsection{Network Architecture}
\label{sec:supp_architecture}

The reconstruction network is a U-Net with multi-head self-attention, adapted from the architecture used by Cao\emph{~et~al.~ }\citep{cao2025noise2image} for Noise2Image, which itself derives from the denoising diffusion probabilistic model (DDPM) U-Net of Ho~et~al.~ \citep{ho2020denoising} via the open-source implementation of Wang~ \citep{lucidrains}.

The network accepts a two-channel input tensor representing spatially binned ($2\times2$) positive and negative noise event counts aggregated over an integration window of 5~s, and produces a single-channel grayscale intensity image.

The encoder--decoder structure follows a four-level hierarchy with channel dimension multipliers of $(1, 2, 4, 8)$ relative to a base dimension of $c = 64$, yielding feature maps of depth 64, 128, 256, and 512 at successive encoder stages. The initial layer is a $7 \times 7$ convolution mapping the two input channels to the base dimension. Table~\ref{tab:architecture} summarises the complete layer structure, and the total number of trainable parameters is approximately $35.7 \times 10^6$.

Each encoder and decoder level comprises two residual blocks followed by an attention layer and a spatial resampling operation. Each residual block consists of two $3 \times 3$ convolutions with Group Normalization~ \citep{wu2018group} (8 groups) and SiLU activations, together with a residual skip connection implemented as a $1\times1$ convolution when input and output channel dimensions differ. The residual blocks are conditioned on a scalar time input via Feature-wise Linear Modulation (FiLM)~ \citep{perez2018film}: a sinusoidal positional encoding~ \citep{vaswani2017attention} of the integration time is projected through a two-layer MLP (with GELU activation, hidden dimension 256) to produce per-channel scale and shift parameters. In the original DDPM architecture, this pathway encodes the diffusion timestep; in our application, it receives the fixed integration time (5~s for all training and inference), so the FiLM layers effectively learn a constant affine transformation of the feature maps. The conditioning pathway is retained to preserve compatibility with the pre-existing architecture.

Linear attention~ \citep{katharopoulos2020transformers} with 4 heads (per-head dimension 32) is applied at the first three encoder and decoder levels, while full-scaled dot-product attention is used at the deepest level and in the bottleneck. All attention modules include 4 learned memory key--value pairs and apply RMS normalisation before the query--key--value projections. Downsampling in the encoder is performed via a pixel-unshuffle rearrangement ($2\times2$ spatial patches mapped to the channel dimension) followed by a $1\times1$ convolution, except at the deepest encoder stage, where a $3\times3$ convolution is used without spatial reduction. Upsampling in the decoder uses nearest-neighbor interpolation followed by a $3\times3$ convolution. Skip connections between corresponding encoder and decoder levels are implemented by channel-wise concatenation, with two skip connections per level (one after the first residual block and one after the second residual block plus attention).

The bottleneck consists of a residual block, a full self-attention layer, and a second residual block, all operating at channel dimension 512. The final output head concatenates the decoder output with the initial convolutional features, processes them through a residual block, and applies a $1\times1$ convolution to produce the single-channel output.


\begin{table}[ht]
    \centering
    \caption{Architecture of the attention U-Net used for noise-to-image reconstruction. ``ResBlock $\times 2$'' denotes two sequential residual blocks with time conditioning (FiLM). ``LinAttn'' and ``FullAttn'' denote linear and full self-attention layers, respectively. Spatial dimensions are shown relative to the input spatial size $H \times W$. Skip connections (``$\oplus$'') denote channel-wise concatenation with features from the corresponding encoder level.}
    
    \label{tab:architecture}
    \begin{tabular}{l l c c l}
    \hline
    \textbf{Stage} & \textbf{Operation} & \textbf{Channels} & \textbf{Spatial Size} & \textbf{Notes} \\
    \hline
    \multicolumn{5}{l}{\textit{Encoder}} \\
    Input & Conv2d $7\times7$, stride 1 & 64 & $H \times W$ & From 2-ch input \\
    Enc-1 & ResBlock $\times 2$ + LinAttn & 64 & $H \times W$ & \\
          & Pixel-unshuffle + Conv $1\times1$ & 128 & $\frac{H}{2} \times \frac{W}{2}$ & Downsample \\
    Enc-2 & ResBlock $\times 2$ + LinAttn & 128 & $\frac{H}{2} \times \frac{W}{2}$ & \\
          & Pixel-unshuffle + Conv $1\times1$ & 256 & $\frac{H}{4} \times \frac{W}{4}$ & Downsample \\
    Enc-3 & ResBlock $\times 2$ + LinAttn & 256 & $\frac{H}{4} \times \frac{W}{4}$ & \\
          & Pixel-unshuffle + Conv $1\times1$ & 512 & $\frac{H}{8} \times \frac{W}{8}$ & Downsample \\
    Enc-4 & ResBlock $\times 2$ + FullAttn & 512 & $\frac{H}{8} \times \frac{W}{8}$ & \\
          & Conv2d $3\times3$ & 512 & $\frac{H}{8} \times \frac{W}{8}$ & No spatial change \\
    \hline
    \multicolumn{5}{l}{\textit{Bottleneck}} \\
          & ResBlock + FullAttn + ResBlock & 512 & $\frac{H}{8} \times \frac{W}{8}$ & \\
    \hline
    \multicolumn{5}{l}{\textit{Decoder}} \\
    Dec-4 & $\oplus$ skip + ResBlock $\times 2$ + FullAttn & 512 & $\frac{H}{8} \times \frac{W}{8}$ & \\
          & Upsample (NN) + Conv $3\times3$ & 256 & $\frac{H}{4} \times \frac{W}{4}$ & \\
    Dec-3 & $\oplus$ skip + ResBlock $\times 2$ + LinAttn & 256 & $\frac{H}{4} \times \frac{W}{4}$ & \\
          & Upsample (NN) + Conv $3\times3$ & 128 & $\frac{H}{2} \times \frac{W}{2}$ & \\
    Dec-2 & $\oplus$ skip + ResBlock $\times 2$ + LinAttn & 128 & $\frac{H}{2} \times \frac{W}{2}$ & \\
          & Upsample (NN) + Conv $3\times3$ & 64 & $H \times W$ & \\
    Dec-1 & $\oplus$ skip + ResBlock $\times 2$ + LinAttn & 64 & $H \times W$ & \\
          & Conv2d $3\times3$ & 64 & $H \times W$ & \\
    \hline
    \multicolumn{5}{l}{\textit{Output}} \\
          & $\oplus$ input features + ResBlock & 64 & $H \times W$ & \\
          & Conv2d $1\times1$ & 1 & $H \times W$ & \\
    \hline
    \end{tabular}
\end{table}

\subsection{Datasets}
\label{sec:supp_datasets}

The base images for training consisted of 1,013 high-resolution artistic human portraits from Unsplash. Noise images were generated either synthetically from these base images using the probability models or experimentally, as described in the main article (Sec.~\ref{m-sec: noise_event_images}). Specific discussion of the nuanced differences between our approach and that of Cao \textit{et al.} in the construction of noise images is discussed in subsection \ref{sec:si_synthetic_comparison} below. For synthetic datasets, each base image was used to generate 2 noise image variations, with the base image augmented by random vertical and horizontal flips and a random brightness multiplier uniformly sampled from $[0.7, 1.3]$, yielding 2,026 noise-image pairs per synthetic dataset. A 5~s accumulation time was used for both synthetic and experimental datasets so as to increase the amount of information available for the reconstruction, allowing the differences between the different synthetic probability models to be more apparent. The mixed training dataset combined the synthetic dataset generated using the saddle-point model (Sec.~\ref{m-subsubsec: saddle_point_approximation}) with the experimental dataset. No normalization of the event counts was applied; preliminary testing showed that unnormalized inputs yielded higher-quality reconstructions, though we make no claim as to why.

Validation during training and final evaluation after training were performed on separate, out-of-distribution datasets derived from the DIV2K image dataset~ \citep{Agustsson_2017_CVPR_Workshops}. The validation set comprised 30 experimental accumulated noise images and associated base images, and was used for checkpointing, learning rate scheduling, and early stopping. The test set comprised 70 experimental accumulated noise images and associated base images, and was used exclusively for final evaluation after training was complete. We note that the datasets used here are an adaptation of those used by Cao~et~al.~ \citep{cao2025noise2image}; their published data repository contains 1,013 base images, although their paper states 1,004. We cannot explain this discrepancy. While the validation and test sets are small in absolute terms, the out-of-distribution images encompass a substantially greater variety of subjects and textures compared to the portrait-only training set, and thus represent a more challenging reconstruction task. Because the base images vary in aspect ratio and are spatially registered to the event camera frame via an affine transformation, the resulting image pairs contain black borders of varying width at the frame edges.

\subsection{Training Procedure}
\label{sec:supp_training}

The network was trained using the Adam optimizer~ \citep{kingma2015adam} with $\beta_1 = 0.9$, $\beta_2 = 0.999$, and no weight decay. The initial learning rate was $2\times10^{-5}$ for the synthetically trained models and $5\times10^{-5}$ for the experimentally trained model, with the lower rate for the synthetic case chosen to mitigate overfitting to the synthetic data. The learning rate was reduced by a factor of 10 upon plateauing of the validation loss, with a patience of 5 epochs (\texttt{ReduceLROnPlateau}). Training proceeded for up to 100 epochs with early stopping: training was halted if all monitored validation metrics (loss, PSNR, and SSIM) simultaneously worsened relative to the previous epoch for 7 consecutive epochs. Training for the experimental and mixed datasets utilized the full 100 epochs, while early stopping was triggered for the synthetic datasets such that the best model was selected at approximately 10 epochs. The batch size was 2.

The loss function is the pixel-wise mean squared error (MSE) between the predicted and ground-truth intensity images:
\begin{equation}
    \mathcal{L} = \frac{1}{N} \sum_{i=1}^{N} \left\| \hat{I}_i - I_i \right\|_2^2 \,,
    \label{eq:mse_loss}
\end{equation}
where $\hat{I}_i$ and $I_i$ denote the predicted and ground-truth images for the $i$-th sample, respectively, and $N$ is the batch size.

During training, random horizontal and vertical flips were applied to the noise-image pairs as data augmentation. Input spatial dimensions were constrained to be divisible by 8 (the network's total downsampling factor) via center cropping. The model was compiled using \texttt{torch.compile} with the \texttt{max-autotune} setting and converted to channels-last memory format for improved GPU utilization on the NVIDIA Blackwell architecture. Training was performed on the full images, including black borders. For validation and test metric computation, the content region of each sample was identified by thresholding the ground-truth image at a pixel intensity of 2/255 (grey scale pixel values ranged from 0 to 255), and PSNR and SSIM were computed only within the resulting bounding box. This per-sample dynamic cropping prevents the black borders from inflating the reported metric values.

\subsection{Computational Environment}
\label{sec:supp_compute}

Training was performed on a single NVIDIA GeForce RTX~5080 GPU (16~GB GDDR7). The software environment consisted of Python~3.11, PyTorch~2.8.0, PyTorch Lightning~2.5.5, and CUDA~12.9. Mixed-precision (fp16) training was employed. Total training wall-clock time was between 30 minutes and 1 hour for the synthetic datasets (due to early stopping to avoid overfitting), approximately 12 hours for the experimental dataset, and approximately 15 hours for the mixed dataset. Inference time was approximately 60~ms per image at 640 by 360 resolution. 

\subsection{Comparison of Synthetic Noise Event Generation Approaches}
\label{sec:si_synthetic_comparison}

Because the quality of synthetically trained CNNs depends directly on the fidelity of the synthetic data, it is instructive to compare our approach to synthetic noise image generation with that of Cao~\emph{et~al.}\citep{cao2025noise2image}, whose code we examined in detail.\footnote{Available at \url{https://github.com/rmcao/noise2image}.}  The two approaches share the same high-level structure, mapping a greyscale image to per-pixel event probabilities and then sampling event counts, but differ substantially in their underlying models, the number and nature of their fitted parameters, and, consequently, in their generalizability. A summary of the key differences is provided in Table~\ref{tab:synthetic_comparison}, with each point elaborated below.

\subsubsection{Probability Model}

Cao~\emph{et~al.} employ exclusively the Gaussian approximation of Poisson-distributed photon arrivals, yielding their noise event probability (their Eq.~3), which, in their notation, is
\begin{equation}\label{eq:n2i_prob}
    p_e(\lambda) = \frac{1}{2} - \frac{1}{2}\,\mathrm{erf}\!\left(\frac{(\lambda + b_{\mathrm{pr}})(e^{\epsilon} - 1)}{\sqrt{2\lambda\,(1 + e^{2\epsilon})}}\right),
\end{equation}
where $\lambda$ is the mean photon count, $\epsilon$ is the log-contrast threshold, and $b_{\mathrm{pr}}$ is a photoreceptor bias term.  As discussed in the main text (Sec.~4), the Gaussian approximation is valid only for $\lambda \gtrsim 10$ and systematically underestimates event probabilities in
the low-intensity regime.
\par
Our approach, by contrast, builds upon this by introducing an exact Poisson and a saddle-point approximation model, which remain valid across all intensity regimes.  Rather than introducing $b_{\mathrm{pr}}$ as a correction later on, we introduce the leakage term as a fundamental and physically motivated parameter $\theta$ directly in the event triggering condition, Eq.~\ref{m-eq:3} and \ref{m-eq:4} in the main article.

\subsubsection{Parameter Structure}

The parameters employed by the two approaches are enumerated in Table~\ref{tab:synthetic_comparison}.  Several distinctions are worth highlighting.

\paragraph{Threshold symmetry.}
In our approach, the log-contrast threshold $B$ is constrained to be the same for positive and negative events, consistent with the hardware settings that a single threshold magnitude is applied symmetrically. Polarity asymmetries are captured through differing coefficients in $\theta_{\pm}(\lambda)$. Cao~\emph{et~al.} instead allow separate thresholds for positive and negative events ($\epsilon_+ = 0.904$, $\epsilon_- = 1.024$, as defined in their code), which permits model error to be absorbed into a quantity that, on physical grounds, should be symmetric.

\paragraph{Effective time parameter.}
Perhaps the most consequential difference is in the treatment of the number of temporal trials.  In our approach, each temporally resolved timestep (each microsecond) of the event camera's recording is treated as a Bernoulli trial (that is, the time stamp of an event is recorded to the microsecond resolution, and so each microsecond represents a trial in which an event may occur). As such, for a given integration time, the number of Bernoulli trials is set equal to the number of microseconds in the integration time (e.g., $10^6$ for a 1\,s integration period/recording time).  
\par
Cao~\emph{et~al.} instead introduce a fitted parameter $N_{\mathrm{eff}} \approx 22$, ('num\_time' in their code) such that the expected event count over a 1 s period is $N_{\mathrm{eff}} \cdot p_e$.  This quantity does not correspond to the physical number of independent trials, which should be on the order of $10^6$ for a 1\,s window; rather, it is an effective scaling factor that compensates for discrepancies between the model probability $p_e$ and the observed event rates.  Because $N_{\mathrm{eff}}$ is fitted jointly with the other parameters, it can absorb errors from the probability model, the intensity mapping, or both, making it difficult to disentangle these contributions.  Moreover,  it is not clear from the formulation how $N_{\text{eff}}$ should change in the case that time steps of a different temporal resolution are used (if an EC used time steps, say, in nanoseconds instead), since it was fitted for a specific experimental configuration.

\paragraph{Additive corrections.}
The Cao~\emph{et~al.} implementation includes an additive illuminance offset ($I_{\mathrm{off}} = 0.193$), which shifts all pixel illuminance values upward, and an additive Poisson noise floor applied exclusively to negative events ($\mu_{\mathrm{neg}} = 0.14$).  Neither of these terms has a direct physical derivation; they serve as empirical corrections to align the synthetic output with a specific camera's behavior.  In our framework, the analogous low-intensity behavior is governed by $\theta(\lambda)$, which is derived from the triggering condition and jointly fitted to both noise and
S-curve data.  We do include a small additive probability floor $c_V$ (of order $10^{-8}$ or smaller), but this is retained only when it improves the overall fit and is orders of magnitude smaller than the Cao~\emph{et~al.} corrections, indicating that the probability model itself accounts for most of the
observed behavior.

\subsubsection{Sampling Distribution and Variance Model}

Our synthetic noise images are generated by sampling from a binomial distribution $\mathrm{Bin}(T,\, P^{\mathrm{eff}}_{i\pm})$, where the effective probability $P^{\mathrm{eff}}_{i\pm}$ incorporates a dead-time correction (Eq.~\ref{m-eq: p_eff_i_pm}) and $T$ is the physical integration time in microseconds. Pixel-to-pixel variations in both $B$ and $\theta$ introduces overdispersion relative to a model in which all pixels share identical parameters, providing a physically motivated mechanism for the excess variance
observed in experimental data.

Cao~\emph{et~al.} provide two sampling modes: Poisson sampling with rate $N_{\mathrm{eff}} \cdot p_e(I)$, or sampling from a generalized negative binomial distribution with intensity-dependent dispersion parameters $r_+(I)$ and $r_-(I)$.  The latter are obtained by interpolation from an empirically calibrated lookup table (stored in a supplementary data file, 'synthetic\_param.npz' in their code), rather than derived from a physical model of the variance.  This approach effectively absorbs the pixel-to-pixel variability, which in large part physically arises from heterogeneity in threshold and leakage parameters, into the event sampling distribution itself, rather than modeling its source.

Additionally, while our approach explicitly models the refractory dead time $R$ and its effect on both the mean and variance of event counts, Cao~\emph{et~al.} make no explicit dead-time correction.  Any dead-time effects are implicitly absorbed into the fitted $N_{\mathrm{eff}}$. As such, how changes in pixel dead-time (such as via EC bias settings or differences between ECs) should be handled is obfuscated in their approach.

\subsubsection{Pixel-to-Pixel Variation}

In our synthetic generation, per-pixel thresholds are sampled as $B_i \sim \mathcal{N}_{[0,\infty)}(\mu_B,\, \sigma_B^2)$, with $\sigma_B / \mu_B \approx 4{-}6\%$, consistent with the order of manufacturer specifications for typical per-pixel variation.  Per-pixel leakage is perturbed multiplicatively, $\theta_{i\pm} = X_i \cdot \theta_{\pm}(\lambda_i)$ with $X_i \sim \mathcal{N}(1, \sigma_X^2)$.  Both sources of variation have clear physical origins.

In the Cao~\emph{et~al.} implementation, a uniform random scale factor $\in [0.8, 1.2]$ is applied to the threshold parameters, but this scale is drawn once per image rather than per pixel.  Per-pixel overdispersion is instead captured entirely by the empirically determined negative binomial dispersion parameters, as described above.

\subsubsection{Live vs Precomputed Construction}

In our approach, both empirical and synthetic training data sets of noise and base-image pairs, as described above, are computed and saved before training is initiated. During the training, these precomputed datasets are accessed and (other than augmentation with random horizontal and vertical flips) used as is. No additional computation is needed for the datasets during training. 
\par
In contrast, in their approach, Cao~\emph{et~al.} employed live sampling and noise image construction during training. Regarding synthetic data, during each epoch the base image is loaded and the full process to generate synthetic event counts for the noise image is performed (including computation of probabilities and the sampling of events, with the randomly scaled threshold parameter mentioned above), in addition to random horizontal and vertical flips and random brightness multiplier. For training on empirical data, during each epoch, a randomly selected clip of a recording, for the specified integration time, is used to construct the noise and base-image pairs. 
\par
While live construction has the advantage of exposing the CNN to greater variation over the course of training, as each epoch presents a distinct realization of the noise, it adds non-trivial computational overhead to an already intensive training procedure. We elected to use precomputed datasets for two reasons. First, precomputating eliminates the per-epoch cost of probability computation and event sampling, substantially reducing total training time. Second, and more importantly for the purposes of this work, using identical datasets across all training runs ensures that differences in reconstruction quality between the probability models are attributable solely to the models themselves, rather than to any stochastic variation in the training data. We note that the use of precomputed datasets does not preclude the generation of large and diverse training sets; the diversity is simply determined at the dataset construction stage rather than introduced on-the-fly.

\subsubsection{Summary}

The approach of Cao~\emph{et~al.} can produce synthetic noise images that closely match their specific experimental setup, as demonstrated by the quality of their synthetically trained CNN.  However, the reliance on a larger number of empirically fitted parameters---several of which lack direct physical motivation and serve primarily to compensate for limitations of the Gaussian-only probability model---limits the transferability of the approach to other sensors or operating conditions without recalibration of these parameters.

Our approach uses fewer free parameters, each of which has a clear physical interpretation and is determined through a joint fit to both noise-event probability curves and S-curve data.  The probability model itself, rather than auxiliary correction terms, accounts for the observed noise behavior across the full intensity range.  Because the parameters ($B$, $\alpha$, $\theta(\lambda)$) correspond to measurable physical quantities of the sensor, they can in principle be determined for any event camera using the procedure described in Section~\ref{m-subsection: Practical Guidance}, enabling the generation of synthetic data matched to a new device without ad hoc recalibration of effective scaling factors or empirical dispersion curves.

\begin{table}[ht]
\centering
\caption{Comparison of synthetic noise event generation approaches.}
\label{tab:synthetic_comparison}
\small
\begin{tabular}{p{3.4cm} p{4.7cm} p{4.7cm}}
\hline
\textbf{Aspect} & \textbf{This work} & \textbf{Cao~\emph{et~al.}~[4]} \\
\hline
Probability model &
Exact Poisson, saddle-point, and Gaussian; valid across all intensity regimes &
Gaussian approximation only; breaks down at low intensities \\[4pt]
Leakage / bias treatment &
Intensity-dependent $\theta(\lambda) = c_1 + c_2\sqrt{\lambda} + c_3\lambda$, entering at the triggering condition &
Scalar bias $b_{\mathrm{pr}}$, added to $\lambda$; intensity-independent \\[4pt]
Threshold ($B$ / $\epsilon$) &
Shared across polarities ($B = 0.15$); asymmetry captured by $\theta_\pm$ &
Separate per polarity ($\epsilon_+ = 0.904$, $\epsilon_- = 1.024$) \\[4pt]
Temporal trials &
Physical: number of $\mu$s in integration time &
Fitted effective parameter $N_{\mathrm{eff}} \approx 22$ \\[4pt]
Dead-time correction &
Explicit (Eq.~\ref{m-eq: p_eff_i_pm}), affecting mean and variance &
Not modeled; implicitly absorbed by $N_{\mathrm{eff}}$ \\[4pt]
Sampling distribution &
$\mathrm{Bin}(T, P^{\mathrm{eff}}_{i\pm})$; overdispersion from pixel parameter sampling &
Poisson or negative binomial with empirically calibrated, intensity-dependent $r(I)$ \\[4pt]
Pixel-to-pixel variation &
Per-pixel sampling of $B_i$ and $\theta_i$ from physically motivated distributions &
Per-image threshold scaling; per-pixel variance via empirical $r(I)$ \\[4pt]
Additive corrections &
Probability floor $c_V \sim 10^{-8}$ (when needed) &
Illuminance offset ($0.193$); additive Poisson noise floor for negative events ($0.14$) \\[4pt]
Parameter determination &
Joint fit to noise curves and S-curves &
Fitted to noise calibration data only \\[4pt]
\hline
\end{tabular}
\end{table}

%
%

\ifSubfilesClassLoaded{\bibliography{Bib}}{}

\AddToHook{enddocument/afteraux}{%
\immediate\write18{
cp output.aux supplemental_info.aux
}%
}
\end{document}